%% file: main.tex
\documentclass[acmtog,final]{acmart}
\setcopyright{none}
\renewcommand\footnotetextcopyrightpermission[1]{} 

\usepackage[table]{xcolor}
\usepackage{multirow}

\AtBeginDocument{%
  }

\input{preamble}

\begin{document}

\title{VeraRetouch: A Lightweight Fully Differentiable Framework for Multi-Task Reasoning Photo Retouching}

\author{Yihong Guo}
\email{guoyihong@zju.edu.cn}
\affiliation{%
  \institution{Zhejiang University}
  \city{Hangzhou}
  \state{Zhejiang}
  \country{China}
}

\author{Youwei Lyu}
\email{youweilv@gmail.com}
\affiliation{%
  \institution{vivo BlueImage Lab}
  \city{Shanghai}
  \country{China}
}

\author{Jiajun Tang}
\email{jiajun.t.cn@gmail.com}
\affiliation{%
  \institution{vivo BlueImage Lab}
  \city{Hangzhou}
  \state{Zhejiang}
  \country{China}
}

\author{Yizhuo Zhou}
\email{zhouyizhuo@zju.edu.cn}
\affiliation{%
  \institution{Zhejiang University}
  \city{Hangzhou}
  \state{Zhejiang}
  \country{China}
}

\author{Hongliang Wang}
\email{hl.wang.ucas@gmail.com}
\affiliation{%
  \institution{University of Chinese Academy of Sciences}
  \city{Hangzhou}
  \state{Zhejiang}
  \country{China}
}

\author{Jinwei Chen}
\email{chenjinwei_1987@126.com}
\affiliation{%
  \institution{vivo BlueImage Lab}
  \city{Hangzhou}
  \state{Zhejiang}
  \country{China}
}

\author{Changqing Zou}
\authornote{Corresponding author.}
\email{aaronzou1125@gmail.com}
\affiliation{%
  \institution{Zhejiang Lab}
  \city{Hangzhou}
  \state{Zhejiang}
  \country{China}
}

\affiliation{%
  \institution{Zhejiang University}
  \city{Hangzhou}
  \state{Zhejiang}
  \country{China}
}

\author{Qingnan Fan}
\email{fqnchina@gmail.com}
\affiliation{%
  \institution{vivo BlueImage Lab}
  \city{Hangzhou}
  \state{Zhejiang}
  \country{China}
}



\input{00_Abstract}


\input{figs/texs/teaser}


\maketitle
\input{01_Introduction}
\input{02_Related_Work}
\input{03_Method}
\input{04_Experiments}

\input{05_Conclusion}

\clearpage
\bibliographystyle{ACM-Reference-Format}
\bibliography{11_Reference}
\clearpage

\appendix

\input{12_Appendix}

\end{document}

%% file: preamble.tex
\usepackage{xspace}
\usepackage{tabularx} 
\usepackage{booktabs}
\usepackage[table]{xcolor} 
\usepackage{booktabs}      
\usepackage{graphicx}      
\usepackage{bm}            
\usepackage{amsmath}

\makeatletter
\DeclareRobustCommand\onedot{\futurelet\@let@token\@onedot}
\def\@onedot{\ifx\@let@token.\else.\null\fi\xspace}

\def\eg{\emph{e.g}\onedot}

\makeatother

\newcommand{\tref}[1]{Tab.~\ref{#1}}

\newcommand{\fref}[1]{Fig.~\ref{#1}}

\newcommand{\sref}[1]{Sec.~\ref{#1}}

\newcommand{\auto}{\textit{Auto-Retouch}}
\newcommand{\style}{\textit{Style-Retouch}}
\newcommand{\param}{\textit{Param-Retouch}}
\newcommand{\dataset}{\textit{AetherRetouch-1M+}}
\newcommand{\samantha}{VeraRetouch}
\newcommand{\veraretouch}{VeraRetouch}

\definecolor{brandeisblue}{rgb}{0.0, 0.44, 1.0}
\definecolor{myGreen}{rgb}{0, .8, .3}
\definecolor{myRed}{rgb}{0.8, .2, .2}
\definecolor{mygray}{RGB}{150, 150, 150}

%% file: 00_Abstract.tex
\begin{abstract}
Reasoning photo retouching has gained significant traction, requiring models to analyze image defects, give reasoning processes, and execute precise retouching enhancements. However, existing approaches often rely on non-differentiable external software, creating optimization barriers and suffering from high parameter redundancy and limited generalization. To address these challenges, we propose VeraRetouch, a lightweight and fully differentiable framework for multi-task photo retouching. 
We employ a 0.5B Vision-Language Model (VLM) as the central intelligence to formulate retouching plans based on instructions and scene semantics. Furthermore, we develop a fully differentiable Retouch Renderer that replaces external tools, enabling direct end-to-end pixel-level training through decoupled control latents for lighting, global color, and specific color adjustments. To overcome data scarcity, we introduce AetherRetouch-1M+, the first million-scale dataset for professional retouching, constructed via a new inverse degradation workflow. Furthermore, we propose DAPO-AE, a reinforcement learning post-training strategy that enhances autonomous aesthetic cognition. Extensive experiments demonstrate that VeraRetouch achieves state-of-the-art performance across multiple benchmarks while maintaining a significantly smaller footprint, enabling mobile deployment. Our code and models are publicly available at \url{https://github.com/OpenVeraTeam/VeraRetouch}.
\end{abstract}

%% file: figs/texs/teaser.tex
\begin{teaserfigure}
  \includegraphics[width=\textwidth]{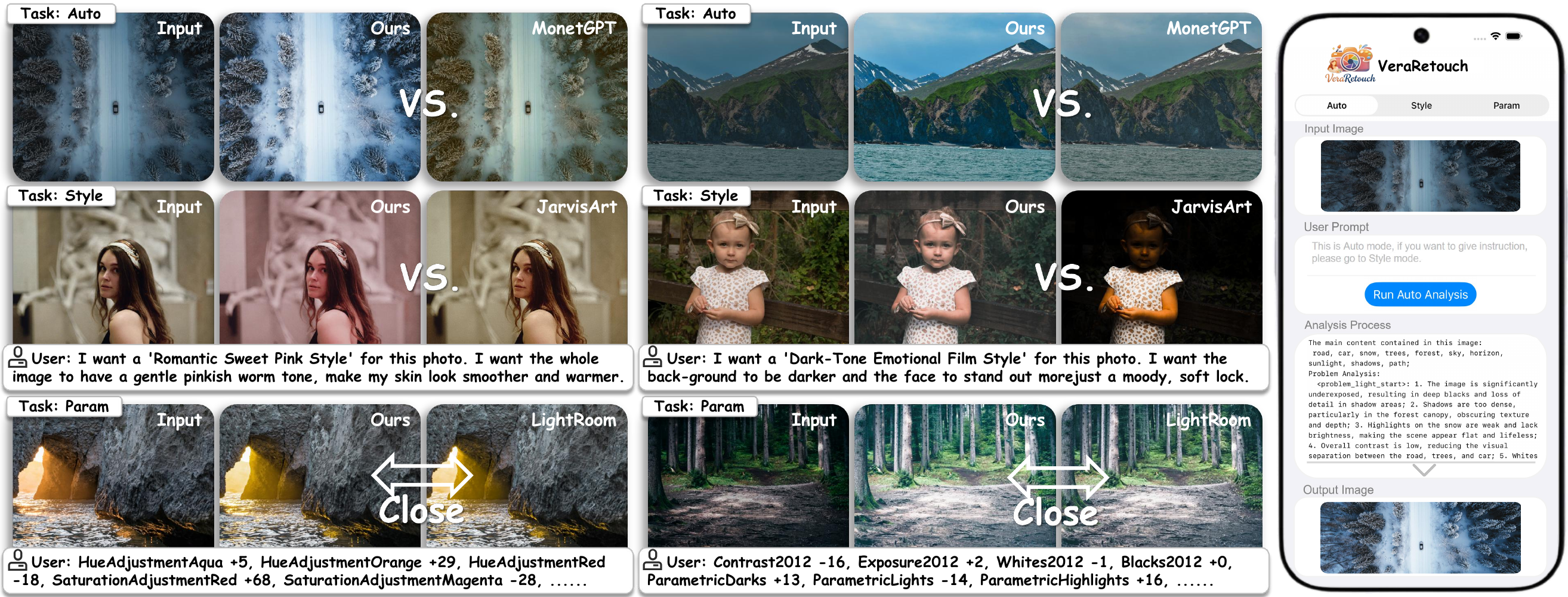}
  \caption{
  We present \textbf{VeraRetouch}, a lightweight, fully differentiable framework for reasoning photo retouching in multiple scenarios: 1) \textbf{\textit{Auto-Retouch}} (top left), with image input only; 2) \textbf{\textit{Style-Retouch}} (middle left), with stylistic prompt, and 3) \textbf{\textit{Param-Retouch}} (bottom left), parameter-driven;
  The mobile-oriented UI workflow (right) takes an input image with an optional user prompt and produces the retouched image with an interpretable analysis.
  }
  \label{fig:teaser}
\end{teaserfigure}

%% file: 01_Introduction.tex
\section{Introduction}
As a cornerstone of digital photography post-processing, photo retouching refines visual aesthetics through precise tone and color adjustments while preserving the original content and fine-grained details. In practice, users' retouching needs vary significantly across diverse scenarios and stylistic preferences, yet meeting these demands typically requires mastery of commercial software (e.g., LightRoom, Photoshop) and specialized color/tonal expertise, creating significant barriers for non-professionals. This necessitates a shift toward reasoning-aware and interactive automatic retouching systems, capable of translating vague user intentions into a logical sequence of professional visual enhancements.

Numerous attempts have been made to automate photo retouching. Early supervised and unsupervised methods~\cite{hu2018exposure, kosugi2020unpaired, yang2022adaint, ouyang2023rsfnet} aimed to learn expert retouching strategies from datasets like MIT-Adobe FiveK~\cite{fivek} and PPR10K~\cite{liang2021ppr10k}. These ``black box'' approaches lack an explicit reasoning process to understand the underlying scene semantics or aesthetic logic, while their generalization is limited due to the small-scale training datasets. With the advancement of diffusion models, diffusion-based approaches have achieved remarkable success in general image editing~\cite{wu2025qwen, labs2025flux, zhang2023magicbrush, brooks2023instructpix2pix}. However, they still struggle with inadequate instruction-following capabilities and insufficient preservation of fine-grained image textures in specialized retouching tasks. Recently, methods such as PhotoArtAgent~\cite{chen2025photoartagent} and JarvisArt~\cite{lin2025jarvisart} have integrated Multimodal Large Language Models (MLLMs) with professional retouching software or tools to enable instruction-driven and reasoning retouching. Nevertheless, these non-differentiable external tools create a fundamental optimization barrier that precludes direct pixel-level end-to-end training, ultimately compromising both retouching precision and generalization. Collectively, these limitations highlight four core challenges in the reasoning retouching field: (1) reliance on non-differentiable external retouching tools reduces the training precision of the model; (2) Suboptimal performance when simultaneously handling both automatic and instruction-based retouching within a single framework; (3) Extreme parameter redundancy and inefficient utilization of massive backbones for specialized color and tonal adjustments; and (4) Lack of large-scale training data limits generalization in complex, real-world scenarios.

To overcome these barriers, we propose VeraRetouch, a lightweight framework designed for multi-task and resolution-independent photo retouching. Centrally, a 0.5B Vision-Language Model (VLM) acts as the ``brain'' to analyze user instructions and scene semantics to formulate a retouching plan. To execute this plan without external software, we develop a fully differentiable Retouch Renderer. This module extracts three disentangled latents from the VLM’s features to independently control lighting, global color, and specific color adjustment. By replacing non-differentiable tools with our pixel-faithful Retouch Renderer, VeraRetouch enables direct end-to-end gradient backpropagation during training. Extensive experiments demonstrate that VeraRetouch achieves superior results even with a significantly smaller model size than existing approaches, while maintaining the capability for efficient mobile deployment.

To address diverse real-world retouching needs, we define three core retouching tasks: (1) \textit{\auto}, which enhances images autonomously without user prompts; (2) \textit{\style}, which applies retouching styles based on language instructions; and (3) \textit{\param}, which executes precise pixel adjustments via operational parameters. To resolve the generalization bottleneck caused by small-scale data, we construct \textbf{\textit{AetherRetouch-1M+}}, the first million-scale dataset covering all three workflows. Specifically, we employ an inverse strategy to ``degrade'' high-quality images for \textit{Auto-Retouch}, utilize over 5,000 style presets for \textit{Style-Retouch}, and map parameters directly to pixels for \textit{Param-Retouch}. By integrating VLM-generated reasoning chains, this dataset enables VeraRetouch to understand "why" behind each adjustment and generalize across complex scenes. In summary, our contributions can be summarized as follows:


(1) We propose VeraRetouch, the first fully differentiable framework to achieve multi-task reasoning photo retouching without any reliance on external retouching software or tools.

(2) We implement the VeraRetouch framework with a mere 0.5B VLM, outperforming existing SOTA methods in both quality and efficiency, while enabling mobile deployment.

(3) We develop an inverse ``degradation'' workflow to synthesize high-quality retouching pairs and construct AetherRetouch-1M+, the first million-scale dataset for multi-task professional retouching.

%% file: 02_Related_Work.tex
\vspace{-0.5cm}
\section{Related Work}
\label{sec:related}
\subsection{Traditional Photo Retouching}
Photo retouching represents an essential and widely adopted practice in the post-processing pipeline of digital photography. 
Early RL-based methods~\cite{hu2018exposure, park2018distort, yu2018deepexposure, kosugi2020unpaired} aimed to mimic human retouching by modeling it as a Markov process, however, they proved time-consuming due to multi-step iterations and consistently fell short in capturing artistic aesthetics. In contrast, another line of research formulates retouching as an end-to-end task, employing fully convolutional generators~\cite{chen2018deep, deng2018aesthetic, chai2020supervised, pan2021miegan, kneubuehler2020flexible, kim2020pienet} to directly output enhanced images or predict parameters for physical models~\cite{moran2020deeplpf, chai2020supervised, gharbi2017deep, kim2021representative, serrano2024namedcurves, zeng2020learning, yang2022adaint, yang2024taming} (e.g., 3D LUTs, tone curves). While efficient, these approaches inherently lack the capacity for deep user interaction and typically fail to produce diverse stylistic outputs. More recently, diffusion models~\cite{wu2025qwen, labs2025flux, zhang2023magicbrush, brooks2023instructpix2pix} have been introduced for image editing, yet their application to photo retouching still faces challenges in preserving content integrity and fine-grained image details. Additionally, most of these works are built on the training of MIT-Adobe FiveK~\cite{fivek} and PPR10K~\cite{liang2021ppr10k} datasets. However, due to limitations in data scale and category coverage, the generalization ability of most methods is restricted in practical scenarios.
\subsection{Reasoning Photo Retouching}
Reasoning photo retouching is a newly proposed and highly focused research task that requires models to understand user instructions, reason about image defects, generate targeted retouching strategies or parameters, and ultimately output a retouched image.
MonetGPT~\cite{dutt2025monetgpt} first introduced VLMs to reasoning retouching tasks. To address the non-differentiable nature of certain retouching operations, it employed a puzzle-based training strategy combined with LoRA fine-tuning, enabling the VLM to indirectly comprehend retouching operations and acquire the ability to generate operation parameters step by step. Subsequently, PhotoArtAgent~\cite{chen2025photoartagent} constructed a training-free agent system by leveraging multiple VLMs and the LightRoom API, incorporating multi-round reasoning and self-feedback mechanisms. Further advancing this direction, JarvisArt~\cite{lin2025jarvisart} implemented a single-inference retouching agent through direct prediction of LightRoom parameters and fine-tuning via GRPO reinforcement learning. However, these methods are constrained by either multi-round reasoning or large model sizes, leading to persistent challenges in inference speed. Moreover, their reliance on external tools introduces issues of version dependency and potential copyright concerns. Additionally, since the retouching operations performed by these external tools are non-differentiable, they do not permit direct pixel-level gradient backpropagation, limiting end-to-end optimization.


%% file: 03_Method.tex
\section{Method}
\label{sec:method}

\subsection{Retouch Encoder and Retouch Renderer}
\label{sec:retouch-model}
Existing Reasoning photo retouching methods~\cite{dutt2025monetgpt,lin2025jarvisart} rely on non-differentiable tools (e.g., LightRoom, Photoshop), creating optimization barriers for end-to-end pixel-level training. Drawing inspiration from the differentiable MLP retouching designs~\cite{lin2023adacm,muruts2025instantretouch}, we propose a fully differentiable dual-module framework (as shown in Fig. \ref{fig:rtouch-encoder-renderer}) comprising a Retouch Encoder \(E\) and a Retouch Renderer \(R\) that can replace professional retouching tools with precise, controllable retouching modeling.

The Retouch Encoder \(E\), built on the ResNet structure~\cite{resnet}, extracts disentangled control latents from pairs of input and target reference images \((I_{\rm ref}^{\rm in}, I_{\rm ref}^{\rm tar})\). Following the principle of independent adjustments in professional retouching workflows and prior studies~\cite{dutt2025monetgpt, chen2025photoartagent}, we decompose the retouching space into three core latent dimensions: 
\begin{equation}
(\mathbf{z}_{\rm l}, \mathbf{z}_{\rm gc}, \mathbf{z}_{\rm sc}) = E(I_{\rm ref}^{\rm in}, I_{\rm ref}^{\rm tar}) .
\end{equation}
Here, \(\mathbf{z}_{\rm l}\), \(\mathbf{z}_{\rm gc}\), and \(\mathbf{z}_{\rm sc}\) target lighting (\eg, exposure, shadows), global color (\eg, tint, temperature), and specific color adjustments (\eg, red luminance), respectively. To enforce disentanglement, we introduce binary masks \(M_{\rm l}, M_{\rm gc}, M_{\rm sc} \in \{0,1\}\) during training, which selectively activate individual latents to form the composite control latent:
\begin{equation}
\mathbf{z} = \texttt{Concat}(M_{\rm l} \cdot \mathbf{z}_{\rm l} ,\ M_{\rm gc} \cdot \mathbf{z}_{\rm gc},\ M_{\rm sc} \cdot \mathbf{z}_{\rm sc}).
\end{equation}

The Retouch Renderer \(R\), implemented as a lightweight pure MLP for per-pixel color mapping, aims at translating the composite latent $\mathbf{z}$ into pixel-level retouching effects. It synthesizes the output image $I^{\rm out} = R(I^{\rm in}; \mathbf{z})$ from an input image $I^{\rm in}$ by additively injecting the latent $\mathbf{z}$ into its hidden layers. Unlike diffusion-based generators, Retouch Renderer enables color and tone adjustments while strictly preserving input structure and high-frequency details. During training, we randomly apply the same set of operations to two different image pairs, where one pair is fed into \(R\) and the other pair serves as input and target for \(E\). 
Benefited from the modeling of \(E\) and \(R\), retouch attributes can be effectively extracted from a pair of reference images and accurately reproduced on the input image. More experimental details can be found in the Appendix.

This encoder-renderer framework serves as the technical foundation for two critical aspects of our work. First, for the end-to-end training of the VeraRetouch framework (\sref{sec:vlm}), the Retouch Renderer provides a differentiable bridge that allows the VLM to optimize parameters through direct image-based supervision. Second, the encoder-renderer framework enables the large-scale construction of the {\dataset} dataset (\sref{sec:dataset}). By utilizing the Retouch Encoder to extract control latents from existing expert-annotated pairs, we can synthesize realistic retouching pairs from any high-quality input with the Retouch Renderer, facilitating professional retouching data collection.
\input{figs/texs/retouchrenderer}
\input{figs/texs/datapipeline}
\subsection{{\dataset} Dataset}
\label{sec:dataset}
To address the generalization limitations of existing methods caused by small-scale, narrowly covered datasets (\eg, MIT-Adobe FiveK~\cite{fivek} and PPR10K~\cite{schuhmann2021laion}), we construct the {\dataset} dataset, with over 1 million retouching pairs tailored to three real-world user demands: Uninstructed Automatic Retouching ({\auto}), Instructed Style-based Retouching ({\style}), and Instructed Parameter-based Retouching ({\param}). \fref{fig:data-pipeline} presents our three data synthesis pipelines tailored for each retouching scenario.

\noindent\textbf{{\auto} Data Generation Pipeline.} This dataset is designed for scenarios where users only provide images without additional instructions. To circumvent the high cost of manual retouching, we adopt an inverse strategy: generating degraded "unretouched" versions from high-quality images. Specifically, we first curate an \textit{Expert Pair Database} by filtering FiveK~\cite{fivek} and PPR10K~\cite{liang2021ppr10k} with aesthetic score to retain only high-quality improvements. Then we select high-aesthetic images from a large-scale photography dataset as pseudo ``retouched'' images. For each pseudo ``retouched'' image, we retrieve data pairs with the most similar histogram features from the \textit{Expert Retouched Pair Database} as references. Feeding the pseudo ``retouched'' image and retrieved reference pairs into our Retouch Encoder and Retouch Renderer, the approach inverts expert retouching logic to generate a degraded ``unretouched'' image, which preserves the content structure of the original high-aesthetic input while embodying realistic flaws.

To further ensure input diversity and enhance generalization, we curate a supplementary dataset by extracting operator ranges and variances from FiveK and PPR10K. By applying randomly sampled operators to high-quality images, we generate disturbed "unretouched" images to expand our training distribution.

\noindent\textbf{{\style} Data Generation Pipeline.} For style-based instruction scenarios, we curate 5,030 online presets, categorized into 11 primary and 193 fine-grained subcategories. After sampling images from Unsplash dataset~\cite{unsplash_datasets}, Qwen2.5-VL~\cite{bai2025qwen2} classifies each image to match appropriate preset categories; one preset is randomly selected and applied via LightRoom API. Qwen3-VL~\cite{wu2025qwen} then generates \textbf{multiple variants of simulated user instructions} through semantic perturbations to expand the instruction diversity.

\noindent\textbf{{\param} Data Generation Pipeline.} For operation parameters instruction scenarios, we categorize retouching parameters into light, global color, and specific color adjustments (consistent with~\sref{sec:retouch-model}). Gaussian-random sampled operation parameter combinations are applied to randomly selected images via LightRoom.

\noindent\textbf{Generation of Reasoning Processes.} To provide reasoning processes for VLM training, we design a hierarchical structured reasoning process, featuring three key parts: (1) Key elements of image content; (2) A point-by-point issue analysis on the original image from three perspectives: light, global color, and specific color; (3) Detailed retouching plans point-by-point corresponding to above analysis. We feed the retouched image pairs and task requirements into Qwen3-VL~\cite{wu2025qwen} to simulate the reasoning process.

\subsection{{\samantha} Framework}
\label{sec:vlm}
\input{figs/texs/modelstructure}
As shown in~\fref{fig:model-structure}, {\samantha} consists of a FastViTHD Vision Encoder, Text Encoder, Multi-Modal LLM, MLP Retouch Adaptor, and Retouch Renderer. We build the framework on FastVLM-0.5B~\cite{vasu2025fastvlm} to reduce the model size and inference latency.
The FastViTHD Vision Encoder encodes the input image into visual tokens, while the Text Encoder converts user instructions into prompt tokens. For user instructions, we design three special tokens: $\langle|$Auto Retouch$|$$\rangle$, $\langle|$Style Retouch$|$$\rangle$, and $\langle|$Param Retouch$|$$\rangle$, which enable task selection (\sref{sec:dataset}) and reduce the model’s task discrimination training burden. Then visual tokens are concatenated with prompt tokens, fed into the Multi-Modal LLM, and autoregressively generate reasoning completions; to extract retouching control latents from completions, three additional special \textit{retouch tokens} for light, global color, and specific color adjustment (\sref{sec:retouch-model}) are designed, their last hidden layer features are fed into the MLP Retouch Adaptor for alignment to generate distangled control latents, which are then used by the Retouch Renderer to convert the input image into the final retouched image.
\input{figs/texs/domain}

\noindent\textbf{Domain Align Pretraining.} The retouch tokens generated by the Multi-Modal LLM exhibit a substantial distribution mismatch with the pre-trained control latents introduced in~\sref{sec:retouch-model}. As shown in \fref{fig:domain-align}, directly feeding control latents produced by Multi-Modal LLM into Retouch Renderer results in a severe degradation in the quality of the retouched images. To address this issue, we design a simple Retouch Adaptor (three-layer bottleneck MLP) for feature space transformation. Furthermore, starting from the pre-trained model, we freeze the Vision Encoder, and train the remaining components using the {\param} dataset. The aligned control latents are obtained by autoregressively generating special \textit{retouch tokens} from given parameters and rendering them into retouched images. For training losses, we adopt cross-entropy loss for token ids ($\mathcal{L}_{\text{CE}}^{\rm text}$) and L1 loss for image reconstruction ($\mathcal{L}_{1}^{\rm img}$), with the total loss computed as a weighted sum:
\begin{equation}
\mathcal{L}_{\text{total}} = \alpha \cdot \mathcal{L}_{\text{CE}}^{\rm text} + \mathcal{L}_{1}^{\rm img}.   
\end{equation}

\noindent\textbf{Reasoning Supervised Fine-tuning (RSFT).}
In this stage, we freeze all other modules and only train the Multi-Modal LLM, initialized from its pre-trained weights, which prevents the probability shift during the domain alignment training from affecting multi-task training. 
The loss function formulation remains consistent with that used in the domain alignment stage.
During training, we perform random sampling at an equal ratio across the {\auto}, {\style}, and {\param} datasets. This phase enables the model to: (1) output results in the predefined structured format; (2) learn the causal relationship between reasoning processes and control latents during autoregressive training; (3) acquire precise retouch adjustment capabilities through direct pixel-level supervision.

\noindent\textbf{DAPO-AE Post-Training.}
The RSFT stage has achieved robust instruction-following capabilities and high retouching quality.
To further boost the model's aesthetic perception and elevate the visual appeal of output images, we introduce a Reinforcement Post-Training (RPT) stage, leveraging decoupled clip and dynamic sampling policy optimization~\cite{yu2025dapo} for aesthetic enhancement (DAPO-AE) to inject refined aesthetic nuances. 
Unlike JarvisArt~\cite{lin2025jarvisart} employing numerous and complex rewards, our DAPO-AE consists of three simple rewards:
The format reward $R_f$ ensures adherence to the structured reasoning template and critical retouch tokens. The image similarity reward $R_s$ aligns with the target’s retouching trends. The aesthetic reward $R_a$, activated only for the \auto~task, enhances the visual aesthetic quality of the output. All individual rewards and their constituent scores are normalized to $[0,1]$ for balanced optimization. 

To adapt to diverse retouch tasks and avoid cross-task interference, we design task-specific reward configurations and training strategies: {\auto} uses all three rewards ($R_f, R_s, R_a$), while {\style} and {\param} only adopt $R_f$ and $R_s$. See appendix for more details.


%% file: figs/texs/retouchrenderer.tex
\begin{figure}
    \centering
  \includegraphics[width=0.48\textwidth]{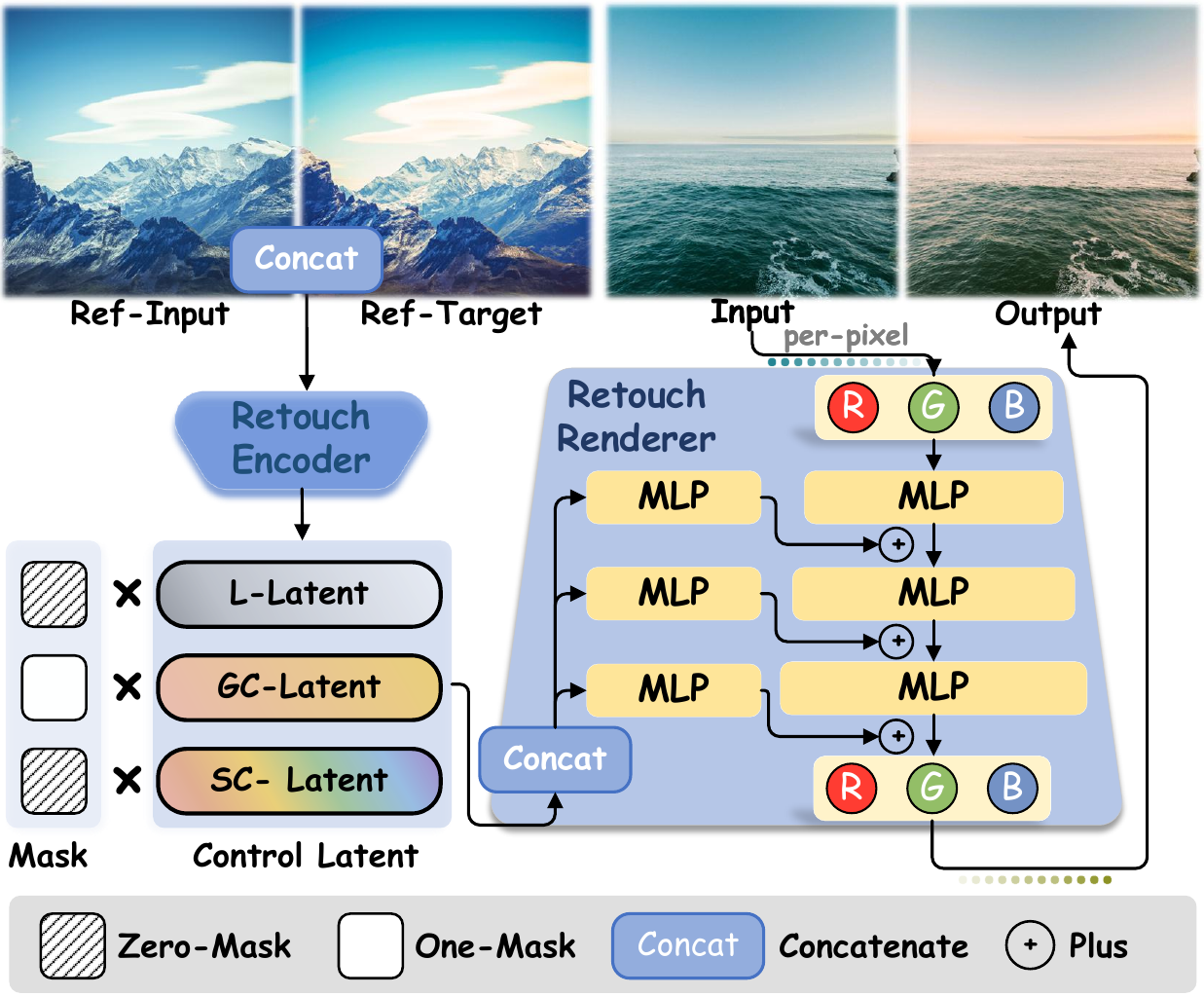}
  \caption{Retouch Encoder and Retouch Renderer Structure. A reference pair (\textit{Ref-Input}, \textit{Ref-Target}) is concatenated and encoded into three \textit{control latents}: \textbf{L-Latent} (light), \textbf{GC-Latent} (global color), and \textbf{SC-Latent} (specific color). A binary mask (zero/one) indicates which factors are activated. The control latents are then fed to the \textbf{Retouch Renderer}, implemented as stacked MLP-based RGB mappings, to produce the final retouched \textit{Output}.}
  \label{fig:rtouch-encoder-renderer}
\end{figure}

%% file: figs/texs/datapipeline.tex
\begin{figure*}
    \centering
  \includegraphics[width=\textwidth]{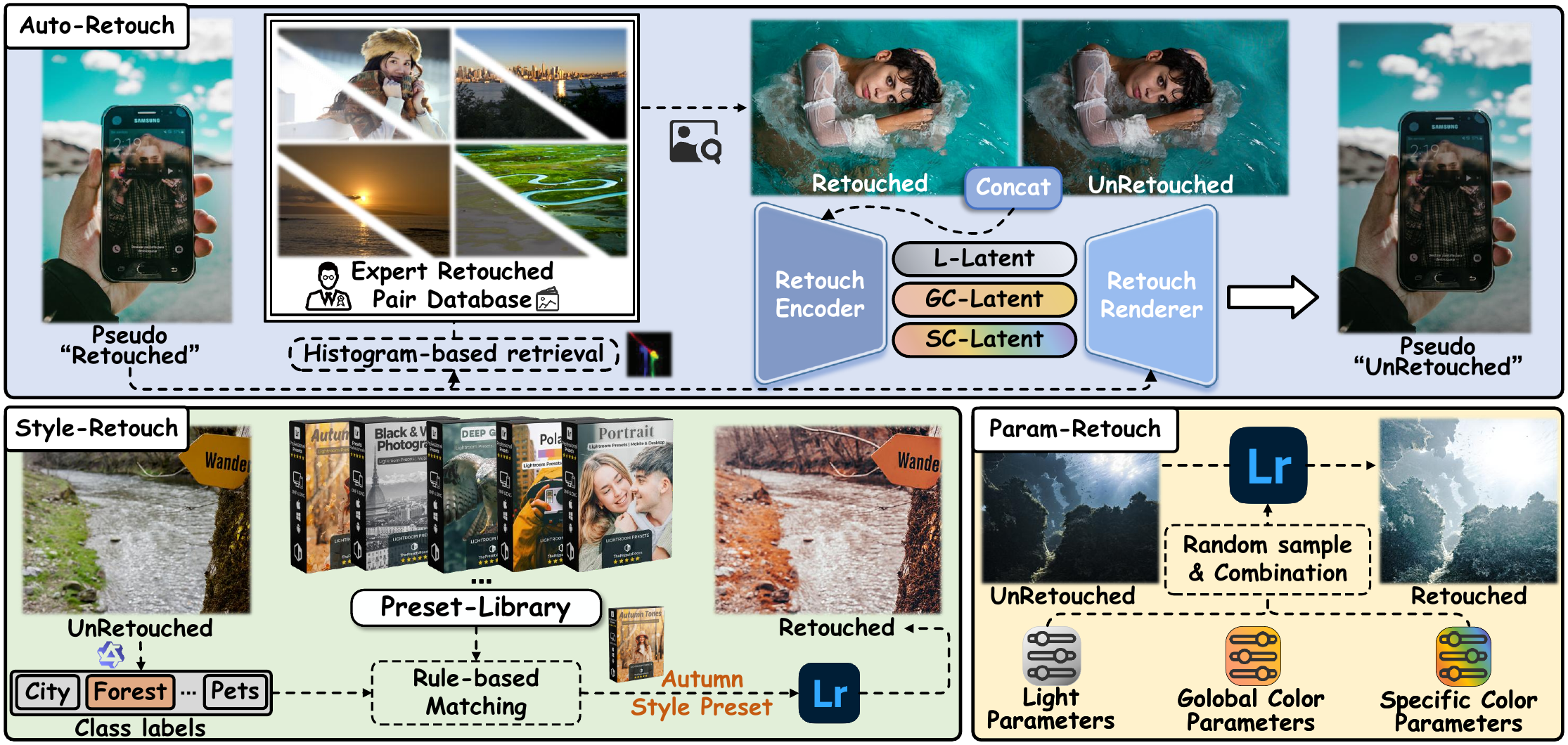}
    \caption{Data synthesis pipelines for \textbf{AetherRetouch-1M+}. Three workflows generate a million-scale multi-task retouching dataset: (1) \textbf{\textit{Auto-Retouch}}: inverting expert retouching to synthesize pseudo unretouched images from high-quality images; (2) \textbf{\textit{Style-Retouch}}:  applying LightRoom presets via rule-based matching; (3) \textbf{\textit{Param-Retouch}}: rendering images with randomly sampled LightRoom parameters.}

  \label{fig:data-pipeline}
\end{figure*}

%% file: figs/texs/modelstructure.tex
\begin{figure*}
    \centering
  \includegraphics[width=\textwidth]{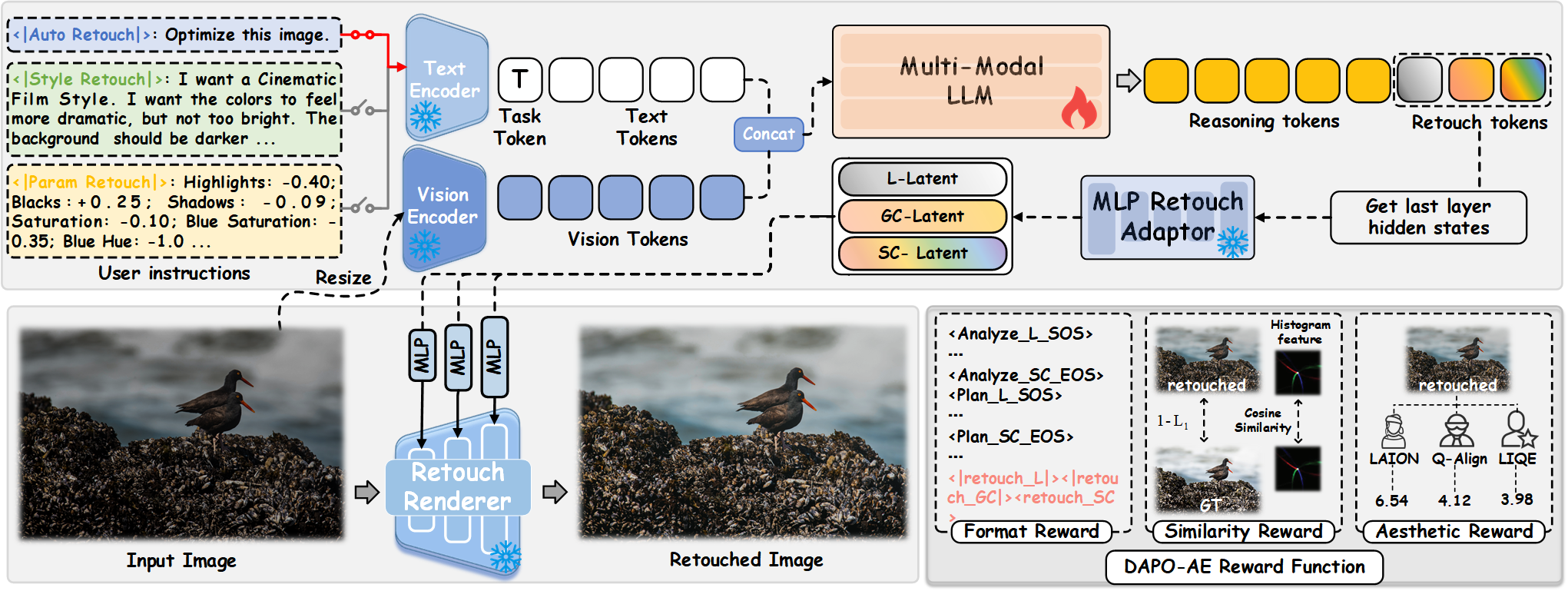}
  \caption{Overview of the \veraretouch~framework. Our framework processes an image and optional prompts through a compact VLM to generate structured reasoning and disentangled retouching latents. These latents drive a fully differentiable renderer to produce the final enhancement. The bottom-right panel illustrates the components of our DAPO-AE reward functions, designed to optimize the model for both logical consistency and high-level aesthetic appeal.}
  \label{fig:model-structure}
\end{figure*}

%% file: figs/texs/domain.tex
\begin{figure}
    \centering
  \includegraphics[width=0.47\textwidth]{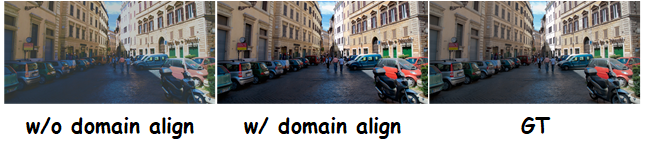}
  \caption{Directly training with pre-trained control latents leads to feature-space mismatch and degraded quality (w/o domain align). By introducing a lightweight adaptor and alignment pre-training, we successfully synchronize the latent space with LLM features, yielding results significantly closer to the ground truth (w/ domain align).}
  \label{fig:domain-align}
\end{figure}

%% file: 04_Experiments.tex
\section{Experiments}
\label{sec:experiment}

\input{tables/fivek-comprision}
\subsection{Experimental Settings}
\noindent\textbf{Datasets.} 
For \auto, we employ approximately 500 samples from the MIT-Adobe FiveK~\cite{fivek} dataset (\textbf{FiveK-Bench}) for real-world performance evaluation. Additionally, we construct a synthetic dataset of 250 images (\textbf{Aether-Bench (Auto)}), subjected to randomized operational perturbations, to assess the model's generalization under complex and varied inputs. For \style, we select presets that are outside the training distribution to generate 100 pairs of test samples (\textbf{Aether-Bench (Style)}). Finally, for \param, we curate a test set of 350 samples (\textbf{Aether-Bench (Param)}) by applying seven distinct retouching protocols to 50 unseen images, with the specific parameters for each protocol randomly sampled from Gaussian distributions during application.

\noindent\textbf{Metrics.} 
To evaluate the fidelity of predictions relative to the ground truth, we employ PSNR, SSIM, and LPIPS. We further utilize histogram intersections to measure the distributional consistency of contrast, luminance, and color saturation. To assess texture preservation and retouching consistency, we adopt DISTS~\cite{dists}, GMSD~\cite{xue2013gmsd}, and TD~\cite{dong2024movingcolor} metrics. Image aesthetics and perceptual quality are quantified using LAION~\cite{laionaesthetics}, Q-Align~\cite{wu2023qalign}, and LIQE~\cite{zhang2023liqe}. 
Following the evaluation protocol of MonetGPT~\cite{dutt2025monetgpt} for the FiveK-Bench, we report the maximum score across the five expert retouches for PSNR, SSIM, LPIPS, and DISTS, while histogram intersections are computed by considering the aggregate distribution across all experts.

\noindent\textbf{Baselines.} 
We compare our method against several state-of-the-art baselines, including RSFNet~\cite{ouyang2023rsfnet}, Nano-Banana~\cite{comanici2025gemini}, Flux.1 Kontext~\cite{labs2025flux}, Qwen-Image-2509~\cite{wu2025qwen}, MonetGPT~\cite{dutt2025monetgpt}, and JarvisArt~\cite{lin2025jarvisart}. Specifically, we retrain RSFNet on the {\auto} dataset. Furthermore, we implement LoRA fine-tuning for Flux.1 Kontext and Qwen-Image-2509 on our full AetherRetouch-1M+ dataset following the DiffSynth-Studio implementation.
\subsection{Comparison}
\noindent\textbf{Quantitative Comparison.} As shown in \tref{tab:fivek_results} and \ref{tab:aether_bench}, {\samantha} consistently achieves state-of-the-art performance across all benchmarks. For \textbf{\textit{\auto}}, it reaches a peak PSNR of 26.85 dB on FiveK-Bench, outperforming Flux.1 Kontext by 1.08 dB and securing top aesthetic scores in Q-Align and LIQE. In \textbf{\textit{\style}}, our model strikes a superior balance between visual enhancement and texture preservation, yielding the lowest Texture Distortion (TD: 0.526) and effectively suppressing generative artifacts. Furthermore, in \textbf{\textit{\param}}, our method achieves a remarkable PSNR of 30.18 dB, significantly surpassing the fine-tuned diffusion baseline.

\input{tables/aether_bench}

\noindent\textbf{Inference Time.} 
We evaluate the efficiency of {\samantha} by measuring the average inference time per image on a single NVIDIA H20 GPU with a batch size of 1 over a test set of 100  512p images. As shown in \tref{tab:efficiency_comparison}, our framework takes only 6.9s to process a single image, outperforming diffusion-based methods like Flux.1 Kontext ($\sim$16.8s) and large-scale agents such as JarvisArt ($\sim$14.3s) with a significant speedup of $\sim$2.5×. Furthermore, we extend our evaluation to edge devices, including a Macbook Air (M4) and an iPhone 16 Pro. As reported in the last two rows of \tref{tab:efficiency_comparison}, our model achieves satisfactory inference speeds of 7.4s and 13.5s respectively, demonstrating the exceptional efficiency and deployment potential of our framework on consumer-grade hardware.
\input{tables/infer-time}
\clearpage
\input{figs/texs/compare-all}
\clearpage

\noindent\textbf{Qualitative Comparison.} 
\fref{fig:comprision} demonstrates the visual superiority of {\samantha} across three retouching tasks. To further validate these results, we conducted a user study with 38 participants. We collected blind rankings of model outputs and converted them into scores on a scale of 1 to 5 (higher is better). We randomly selected 10 images each from the {\auto} and {\style} test sets; the former was evaluated on \textit{visual aesthetics} and \textit{texture consistency}, while the latter focused on \textit{instruction alignment}. As shown in \fref{fig:userstudy}, {\samantha} consistently receives the highest scores in Aesthetics, Prompt Fidelity, and Texture Consistency. These results confirm that our method aligns more closely with human preferences and intent while better preserving original image content.
\input{figs/texs/userstudy}

\subsection{Ablation Study}
\noindent\textbf{Latent-Prediction.} 
We evaluate the effectiveness of our control latent prediction against direct parameter prediction on the MIT-Adobe FiveK Expert-C dataset. While the former utilizes a Retouch Renderer to interpret continuous latents, the latter predicts discrete LightRoom parameters integrated via the LightRoom API.
As shown in \tref{tab:ablation_pred_method_brief}, the latent-prediction approach consistently outperforms the parameter-prediction baseline across all quantitative metrics. This advantage is primarily attributed to the direct gradient backpropagation enabled by the differentiable renderer, which allows the VLM to bypass the discretization gap of traditional APIs and learn more precise, pixel-level aesthetic adjustments.

\noindent\textbf{Data Scaling.} 
To assess the data scalability effect of {\samantha}, we evaluate {\samantha} on \textit{Auto-Retouch} with 5\%, 20\%, and 100\% training data. Quantitative metrics in \tref{tab:ablation_scaling} exhibit a consistent upward trend as the dataset expands. This improvement demonstrates that our model effectively leverages larger-scale data to refine its aesthetic reasoning and retouching precision. Additionally, our unified version (joint training on all three tasks without user instruction perturbations) achieves the best fidelity and consistency, verifying that improvements come from both larger data scale and multi-task supervision.
\input{tables/ablation_param_vs_latents_short}
\input{tables/ablation_data_scale}

\noindent\textbf{DAPO-AE.} 
Tables \ref{tab:fivek_results} and \ref{tab:aether_bench} compare the SFT baseline and the DAPO-AE training scheme. Despite marginal quantitative gains from the additional DAPO-AE stage, it plays a crucial role in refining model performance. We observe that DAPO-AE specifically benefits those challenging samples where the SFT model produces suboptimal reasoning and aesthetic results. \fref{fig:dapo} shows that DAPO-AE improves performance on challenging cases where the SFT model typically yields suboptimal reasoning and aesthetic results. 
The preference user study in \tref{tab:dapo-userstudy} also confirms DAPO-AE's superior aesthetic quality. With a 61.62\% user preference rate, our RL-based approach clearly enhances visual appeal in ways that standard numerical evaluations may fail to fully capture.
\input{figs/texs/dapo}

\input{tables/human-evaluation}
\input{figs/texs/decouple}

\noindent\textbf{Disentanglement Ability.} To verify the decoupling of operator categories(Light, Global Color and Specific Color), we performed an intervention study on 50 external images. We synthesized 350 Ground-Truth(GT) references by applying Gaussian-sampled parameters in various combinations. During inference, we provided all parameters while masking specific retouching latents to observe the model's isolation capability. As shown in \tref{tab:ablation-distangle}, our method maintains an average PSNR>28 with corresponding GTs across all mask scenarios, confirming that the latent space for each operator category is effectively disentangled and independent. These quantitative findings are further corroborated by qualitative results. Such quantitative results are consistent with qualitative observations in Fig. \ref{fig:decouple} on MIT-Adobe FiveK. Masking L-Latent only changes illumination and preserves original color attributes. Masking GC-Latent adjusts global color tones without affecting lighting, and SC-Latent masking selectively modulates local color components while maintaining overall color balance. 
\input{tables/distanglement}

%% file: tables/fivek-comprision.tex
\begin{table*}[!t]
\centering
\caption{Quantitative Comparison on \textbf{FiveK-Bench}. \textcolor{red}{Red} and \textcolor{blue}{blue} indicate the best and second-best results, respectively.}
\label{tab:fivek_results}
\resizebox{\textwidth}{!}{%
\begin{tabular}{l|ccc|cccc|ccc|ccc}
\toprule
\textbf{Method} & PSNR$\uparrow$ & SSIM$\uparrow$ & LPIPS$\downarrow$ & Hist-L$\uparrow$ & Hist-C$\uparrow$ & Hist-S$\uparrow$ & Hist-M$\uparrow$ & LAION$\uparrow$ & Q-Align$\uparrow$ & LIQE$\uparrow$ & DISTS$\downarrow$ & GMSD$\downarrow$ & TD$\downarrow$ \\
\midrule
RSFNet & 25.07 & 0.935 & 0.056 & 82.00\% & 72.08\% & 79.05\% & 77.71\% & 5.02$\pm$0.69 & 4.06$\pm$0.40 & 3.62$\pm$0.96 & 0.044 & \textcolor{red}{0.020} & \textcolor{red}{0.364} \\
Nano Banana & 20.30 & 0.616 & 0.137 & 82.38\% & 70.18\% & 63.80\% & 72.12\% & \textcolor{red}{5.18$\pm$0.75} & \textcolor{blue}{4.15$\pm$0.39} & 3.54$\pm$0.95 & 0.075 & 0.142 & 1.654 \\
Flux.1 Kontext & 25.77 & 0.896 & 0.079 & 88.42\% & \textcolor{red}{95.04\%} & \textcolor{blue}{92.79}\% & 92.09\% & 5.07$\pm$0.66 & 3.99$\pm$0.40 & 3.61$\pm$0.96 & 0.062 & 0.040 & 0.730 \\
Qwen-Image-2509 & 17.81 & 0.572 & 0.193 & 61.58\% & 66.77\% & 76.33\% & 68.23\% & 4.87$\pm$0.62 & 3.91$\pm$0.50 & 3.13$\pm$0.98 & 0.102 & 0.164 & 1.659 \\
MonetGPT & 22.91 & 0.914 & 0.064 & 79.30\% & 65.78\% & 78.01\% & 74.36\% & 4.86$\pm$0.64 & 4.01$\pm$0.40 & 3.54$\pm$0.96 & 0.057 & \textcolor{blue}{0.023} & \textcolor{blue}{0.480} \\
JarvisArt & 21.52 & 0.865 & 0.149 & 72.74\% & 60.23\% & 76.69\% & 69.89\% & \textcolor{blue}{5.14$\pm$0.61} & 4.05$\pm$0.45 & 3.02$\pm$0.95 & 0.108 & 0.039 & 0.771 \\
\midrule
\textbf{Ours-SFT} & \textcolor{blue}{26.04} & \textcolor{blue}{0.936} & \textcolor{blue}{0.053} & \textcolor{blue}{90.44\%} & 92.71\% & \textcolor{red}{95.33\%} & \textcolor{blue}{92.83\%} & 5.13$\pm$0.68 & \textcolor{red}{4.18$\pm$0.39} & \textcolor{red}{3.92$\pm$0.93} & \textcolor{blue}{0.040} & 0.061 & 0.694 \\
\textbf{Ours-DAPO-AE} & \textcolor{red}{26.85} & \textcolor{red}{0.939} & \textcolor{red}{0.049} & \textcolor{red}{96.35\%} & \textcolor{blue}{94.13\%} & 92.13\% & \textcolor{red}{94.20\%} & 5.10$\pm$0.68 & \textcolor{blue}{4.15$\pm$0.39} & \textcolor{blue}{3.88$\pm$0.94} & \textcolor{red}{0.039} & 0.045 & 0.607 \\
\bottomrule
\end{tabular}%
}
\end{table*}

%% file: tables/aether_bench.tex
\begin{table}[t]
\centering
\caption{Quantitative evaluation on \textbf{Aether-Bench}. Each sub-task (Auto, Style, and Param) features its unique set of evaluation metrics. \textbf{Rea.} denotes reasoning capability.}
\label{tab:aether_bench}
\resizebox{\columnwidth}{!}{%
\setlength{\tabcolsep}{2.5pt} 
\begin{tabular}{lc|ccccccc}
\toprule
\textbf{Method} & \textbf{Rea.} & \multicolumn{7}{c}{\textbf{Experimental Results}} \\
\midrule

\rowcolor{gray!15} \multicolumn{9}{c}{\textit{Aether-Bench (Auto)}} \\
\rowcolor{gray!5} & & \textbf{Hist-M$\uparrow$} & \textbf{LAION$\uparrow$} & \textbf{Q-Align$\uparrow$} & \textbf{LIQE$\uparrow$} & \textbf{DISTS$\downarrow$} & \textbf{GMSD$\downarrow$} & \textbf{TD$\downarrow$} \\
RSFNet          & $\times$     & 89.17\% & \textcolor{red}{6.85} & 4.23 & 3.22 & \textcolor{blue}{0.059} & \textcolor{blue}{0.030} & \textcolor{red}{0.343} \\
Nano Banana     & $\checkmark$ & 86.39\% & \textcolor{blue}{6.83} & \textcolor{blue}{4.26} & 3.26 & 0.088 & 0.143 & 1.547 \\
Flux.1 Kontext  & $\times$     & \textcolor{red}{89.81\%} & 6.76 & 4.25 & \textcolor{red}{3.31} & 0.063 & 0.039 & 0.665 \\
Qwen-Image-2509 & $\times$     & 79.65\% & 6.51 & 4.02 & 2.87 & 0.120 & 0.161 & 1.523 \\
MonetGPT        & $\checkmark$ & 85.03\% & 6.36 & 4.01 & 2.91 & 0.104 & 0.038 & 0.536 \\
JarvisArt       & $\checkmark$ & 81.14\% & 6.34 & 4.06 & 2.60 & 0.123 & 0.043 & 0.709 \\
\midrule
\textbf{Ours-SFT} & $\checkmark$ & 88.55\% & \textcolor{blue}{6.83} & \textcolor{red}{4.27} & \textcolor{blue}{3.30} & 0.061 & 0.035 & 0.435 \\
\textbf{Ours-DAPO-AE} & $\checkmark$ & \textcolor{blue}{89.59\%} & 6.82 & 4.25 & 3.28 & \textcolor{red}{0.055} & \textcolor{red}{0.026} & \textcolor{blue}{0.360} \\

\midrule
\rowcolor{gray!15} \multicolumn{9}{c}{\textit{Aether-Bench (Style)}} \\
\rowcolor{gray!5} & & \boldmath{$L_1 \downarrow$} & \textbf{PSNR$\uparrow$} & \textbf{SSIM$\uparrow$} & \textbf{LPIPS$\downarrow$} & \textbf{DISTS$\downarrow$} & \textbf{GMSD$\downarrow$} & \textbf{TD$\downarrow$} \\
Nano Banana     & $\checkmark$ & 0.125 & 16.66 & 0.596 & 0.242 & 0.138 & 0.151 & 1.579 \\
Flux.1 Kontext  & $\times$     & \textcolor{blue}{0.094} & 19.48 & 0.831 & 0.162 & 0.106 & 0.048 & 0.741 \\
Qwen-Image-2509 & $\times$     & 0.158 & 14.34 & 0.494 & 0.289 & 0.196 & 0.174 & 1.725 \\
JarvisArt       & $\checkmark$ & 0.147 & 15.72 & 0.677 & 0.288 & 0.170 & 0.100 & 1.235 \\
\midrule
\textbf{Ours-SFT} & $\checkmark$ & 0.097 & \textcolor{blue}{19.73} & \textcolor{blue}{0.839} & \textcolor{blue}{0.149} & \textcolor{blue}{0.100} & \textcolor{blue}{0.039} & \textcolor{blue}{0.592} \\
\textbf{Ours-DAPO-AE} & $\checkmark$ & \textcolor{red}{0.092} & \textcolor{red}{20.12} & \textcolor{red}{0.847} & \textcolor{red}{0.145} & \textcolor{red}{0.099} & \textcolor{red}{0.036} & \textcolor{red}{0.526} \\

\midrule
\rowcolor{gray!15} \multicolumn{9}{c}{\textit{Aether-Bench (Param)}} \\
\rowcolor{gray!5} & & \boldmath{$L_1 \downarrow$} & \textbf{PSNR$\uparrow$} & \textbf{SSIM$\uparrow$} & \textbf{LPIPS$\downarrow$} & \textbf{DISTS$\downarrow$} & \textbf{GMSD$\downarrow$} & \textbf{TD$\downarrow$} \\
Flux.1 Kontext  & $\times$     & 0.140 & 18.51 & 0.783 & 0.204 & 0.136 & \textcolor{red}{0.008} & \textcolor{red}{0.468} \\
Qwen-Image-2509 & $\times$     & 0.283 & 13.55 & 0.380 & 0.484 & 0.257 & 0.198 & 1.843 \\
\midrule
\textbf{Ours-SFT} & $\checkmark$ & \textcolor{red}{0.023} & \textcolor{red}{30.39} & \textcolor{blue}{0.946} & \textcolor{red}{0.039} & \textcolor{red}{0.040} & 0.071 & \textcolor{blue}{0.664} \\
\textbf{Ours-DAPO-AE} & $\checkmark$ & \textcolor{blue}{0.024} & \textcolor{blue}{30.18} & \textcolor{red}{0.947} & \textcolor{blue}{0.041} & \textcolor{blue}{0.042} & \textcolor{blue}{0.067} & \textcolor{blue}{0.644} \\

\bottomrule
\end{tabular}%
}
\end{table}

%% file: tables/infer-time.tex


\begin{table}[t]
\centering
\caption{Comparison on \textbf{model size} and \textbf{per-image latency}. Total Time denotes end-to-end latency, split into VLM Time (backbone) and Other Time (renderer/LightRoom/other tools, I/O).}
\label{tab:efficiency_comparison}
\resizebox{\columnwidth}{!}{%
\begin{tabular}{l|cc|c|ccc} 
\toprule
\textbf{Method} & \textbf{Task} & \textbf{Device} & \textbf{Params$\downarrow$} & \textbf{Total Time$\downarrow$} & \textbf{VLM Time} & \textbf{Other Time} \\
\midrule
Flux.1 Kontext   & \textit{Auto} & H20 & 16.87B & 16.78s & --- & --- \\
Qwen-Image-2509  & \textit{Auto} & H20 & 28.85B & 48.77s & --- & --- \\
MonetGPT         & \textit{Auto} & H20 & 8.29B  & 44.33s & 28.69s & 15.64s \\
JarvisArt        & \textit{Auto} & H20 & 8.29B  & 14.31s & 14.11s & 0.20s \\
\midrule
\textbf{Ours}    & \textit{Auto} & H20 & \textbf{0.63B} & \textbf{6.90s} & \textbf{6.86s} & \textbf{0.04s} \\
\textbf{Ours}   & \textit{Style} & H20 & \textbf{0.63B} & \textbf{3.83s} & \textbf{3.78s} & \textbf{0.05s} \\
\textbf{Ours}   & \textit{Param} & H20 & \textbf{0.63B} & \textbf{5.17s} & \textbf{5.14s} & \textbf{0.03s} \\
\midrule
\textbf{Ours}    & \textit{Auto} & Macbook Air(M4) & \textbf{0.63B} & \textbf{7.46s} & \textbf{6.69s} & \textbf{0.77s} \\
\textbf{Ours}    & \textit{Auto} & iPhone16 pro & \textbf{0.63B} & \textbf{13.56s} & \textbf{	11.58s} & \textbf{1.98s} \\
\bottomrule
\end{tabular}%
}
\end{table}

%% file: figs/texs/compare-all.tex
\begin{figure*}[ht]
    \centering
  \includegraphics[width=\textwidth]{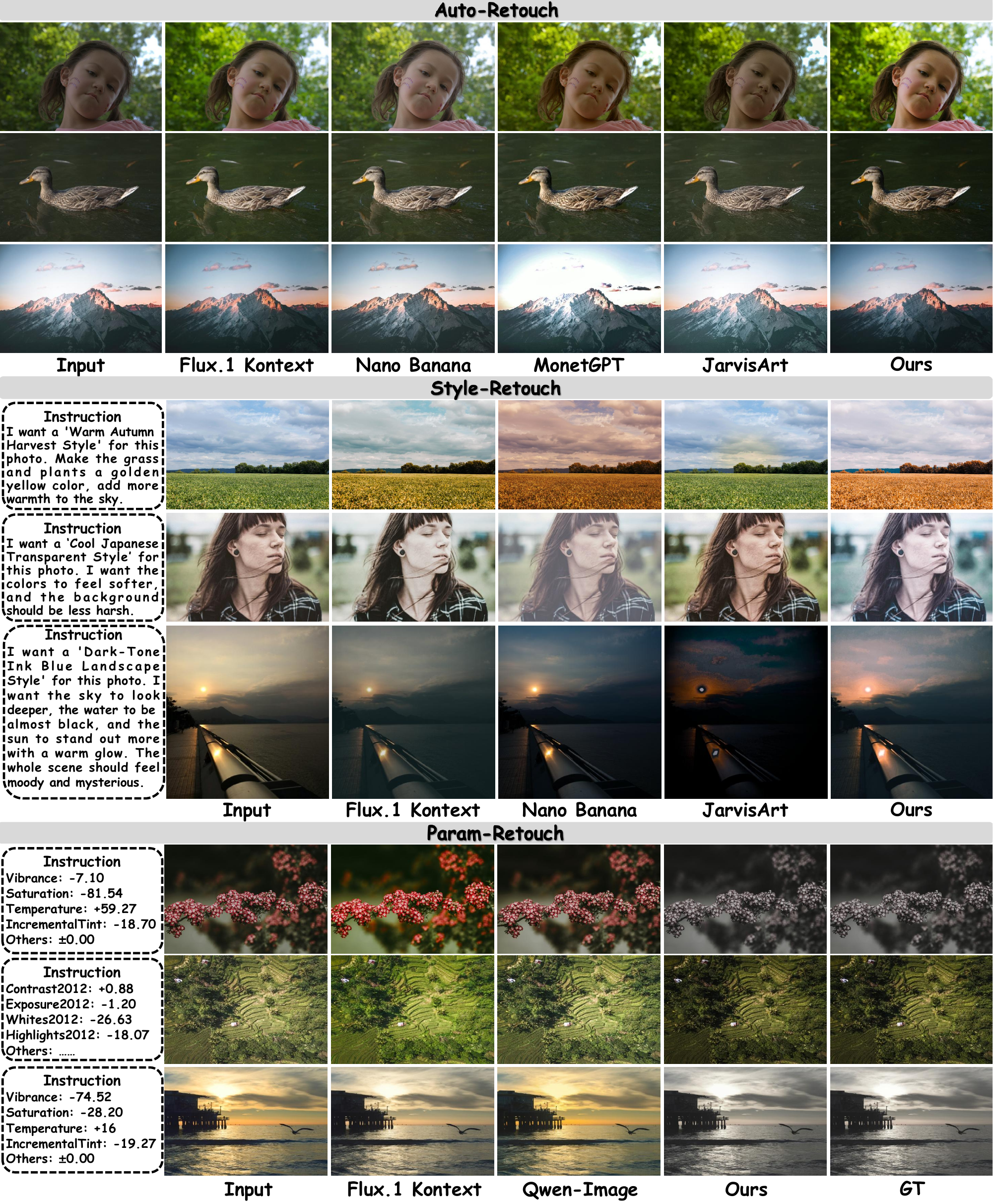}
  \caption{Visual comparison with baseline methods on \textbf{\textit{Auto-Retouch}} (image-only), \textbf{\textit{Style-Retouch}} (text-guided), and \textbf{\textit{Param-Retouch}} (parameter-driven).} 
  \label{fig:comprision}
\end{figure*}

%% file: figs/texs/userstudy.tex
\begin{figure}[th]
    \centering
    \includegraphics[width=0.48\textwidth]{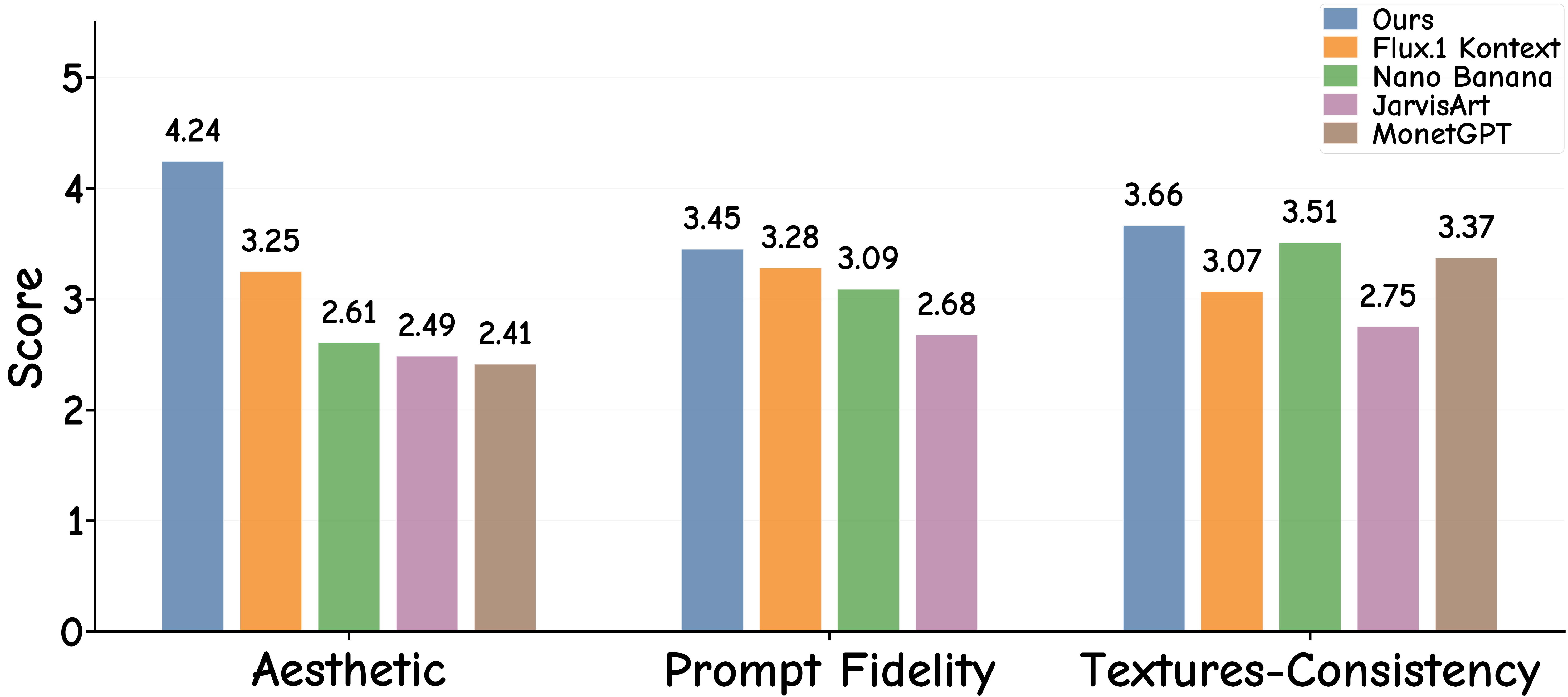}
    \caption{User study results on Aesthetics (visual appeal), Prompt Fidelity (instruction alignment), and Texture Consistency (detail preservation).}
    \label{fig:userstudy}
\end{figure}

%% file: tables/ablation_param_vs_latents_short.tex





\begin{table}[t]
\centering
\caption{Effectiveness of latent-prediction ablation study results, evaluated on the MIT Adobe-FiveK expert-C test dataset.}
\label{tab:ablation_pred_method_brief}
\small{
\begin{tabular*}{\columnwidth}{@{\extracolsep{\fill}}l|ccccc@{}}
\toprule
\textbf{Method} & $L_1 \downarrow$ & PSNR$\uparrow$ & SSIM$\uparrow$ & LPIPS$\downarrow$ & DISTS$\downarrow$ \\
\midrule
params-pred  & 0.125 & 18.07 & 0.800 & 0.155 & 0.086 \\
latents-pred & \textbf{0.061} & \textbf{24.11} & \textbf{0.905} & \textbf{0.057} & \textbf{0.042} \\
\bottomrule
\end{tabular*}
}
\end{table}

%% file: tables/ablation_data_scale.tex
\begin{table}[t]
\centering
\caption{Data-scaling ablation study, evaluated on \textbf{FiveK-Bench}.}
\label{tab:ablation_scaling}
\resizebox{\columnwidth}{!}{%
\begin{tabular}{l|ccc|c|ccc}
\toprule
\textbf{Scale} & PSNR$\uparrow$ & SSIM$\uparrow$ & LPIPS$\downarrow$ & Hist-M$\uparrow$ & LAION$\uparrow$ & Q-Align$\uparrow$ & LIQE$\uparrow$ \\
\midrule
5\%     & 25.39 & 0.931 & 0.058 & 87.81\% & 5.01$\pm$0.66 & 4.09$\pm$0.39 & 3.67$\pm$0.94 \\
20\%    & 26.06 & 0.935 & 0.052 & 94.49\% & 5.08$\pm$0.68 & 4.16$\pm$0.39 & 3.86$\pm$0.94 \\
100\%   & 26.57 & 0.935 & 0.052 & 94.46\% & 5.12$\pm$0.68 & 4.17$\pm$0.39 & 3.90$\pm$0.97 \\
\midrule
\textbf{Unified} & 26.81 & 0.939 & 0.050 & 94.54\% & 5.10$\pm$0.68 & 4.16$\pm$0.39 & 3.89$\pm$0.95 \\
\bottomrule
\end{tabular}%
}
\end{table}

%% file: figs/texs/dapo.tex
\begin{figure}[!h]
    \centering
  \includegraphics[width=0.48\textwidth]{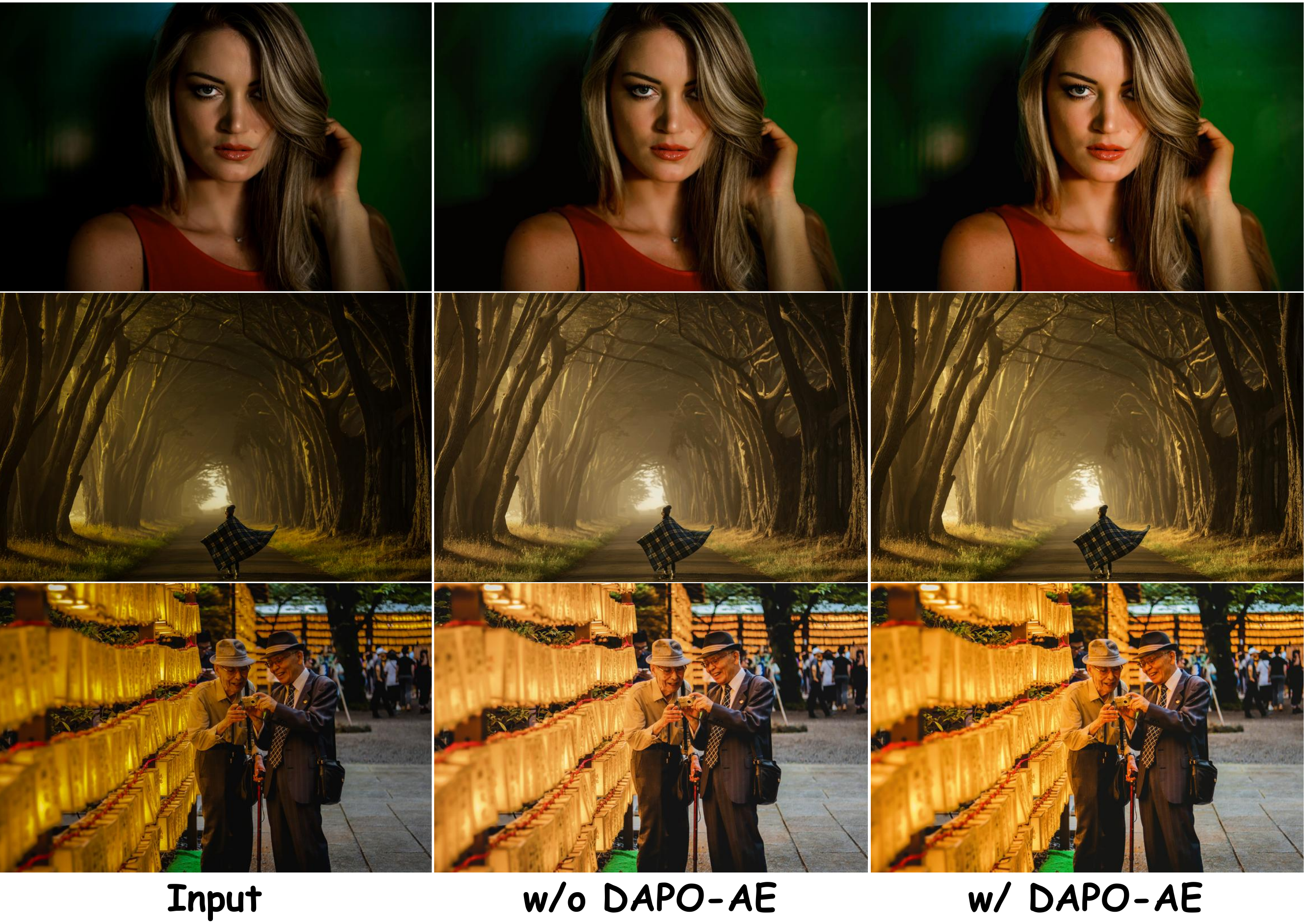}
  \caption{Qualitative comparison of image retouching results with and without our proposed DAPO-AE training.}
  \label{fig:dapo}
\end{figure}

%% file: tables/human-evaluation.tex


\begin{table}[ht]
\centering
\caption{Preference User Study Results of DAPO-AE.}
\label{tab:dapo-userstudy}
\small
\begin{tabular}{lcc} 
\toprule
\textbf{Method} & \textbf{w/ DAPO-AE} & \textbf{w/o DAPO-AE} \\
\midrule
Preference Rate & \textbf{61.62\%} & 38.38\% \\
\bottomrule
\end{tabular}
\end{table}

%% file: figs/texs/decouple.tex
\begin{figure*}[!h]
    \centering
  \includegraphics[width=\textwidth]{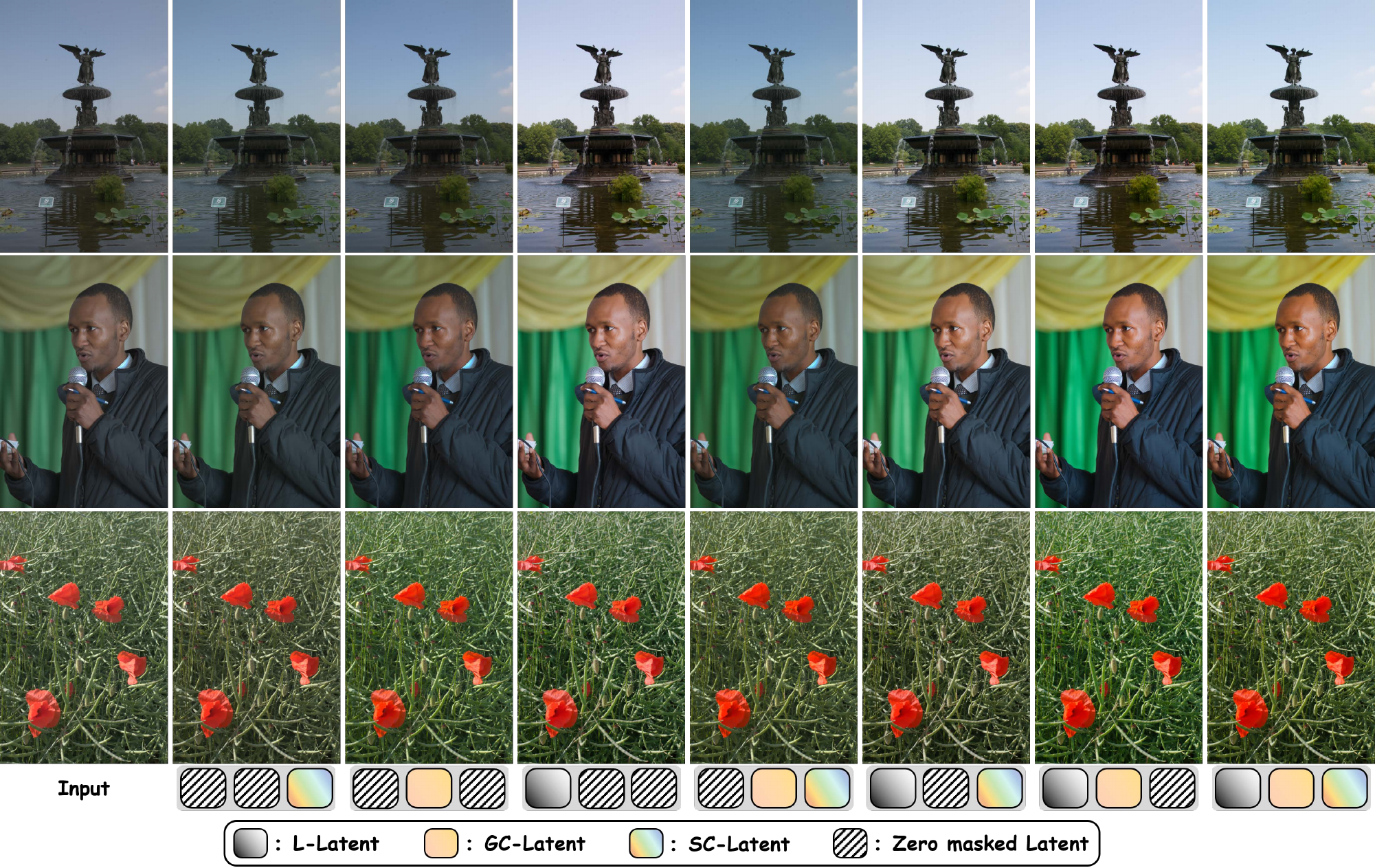}
  \caption{To demonstrate the disentangling capability of our retouch renderer, we apply zero masking to individual control latents during the Auto-Retouch process on MIT-Adobe FiveK samples. The resulting images illustrate the independent impact of each latent—Lighting (L), Global Color (GC), and Specific Color (SC)—demonstrating the effective decoupling of our differentiable renderer.}
  \label{fig:decouple}
\end{figure*}

%% file: tables/distanglement.tex
\begin{table}[ht]
\centering
\caption{Ablation study on the decoupling of operator categories. We report the performance under various mask interventions for $L$, $GC$, and $SC$ latents.}
\label{tab:ablation-distangle}
\begin{tabular*}{\columnwidth}{@{\extracolsep{\fill}} ccccccc @{}}
\toprule
mask-L & mask-GC & mask\_SC & L1$\downarrow$ & PSNR$\uparrow$ & SSIM$\uparrow$ & LPIPS$\downarrow$ \\
\midrule
$\times$ & $\checkmark$ & $\checkmark$ & 0.025 & 30.38 & 0.939 & 0.033 \\
$\checkmark$ & $\times$ & $\checkmark$ & 0.024 & 29.61 & 0.947 & 0.037 \\
$\checkmark$ & $\checkmark$ & $\times$ & 0.053 & 28.12 & 0.925 & 0.030 \\
$\times$ & $\times$ & $\checkmark$ & 0.024 & 30.67 & 0.947 & 0.034 \\
$\times$ & $\checkmark$ & $\times$ & 0.030 & 28.20 & 0.917 & 0.060 \\
$\checkmark$ & $\times$ & $\times$ & 0.026 & 29.20 & 0.942 & 0.045 \\
$\times$ & $\times$ & $\times$ & 0.027 & 29.19 & 0.928 & 0.053 \\
\bottomrule
\end{tabular*}
\end{table}

%% file: 05_Conclusion.tex
\vspace{-0.15cm}
\section{Conclusion}
We present {\samantha}, a new framework that integrates a 0.5B VLM with a fully differentiable Retouch Renderer for reasoning photo retouching. By formulating retouching as a structured autoregressive task, our method effectively bridges the gap between high-level retouching adjustment texts and low-level pixel adjustments. Supported by our million-scale AetherRetouch-1M+ dataset, extensive results demonstrate that even a lightweight model can achieve superior performance through meticulous data curation and model design, highlighting its potential for mobile deployment.

\noindent{\textbf{Limitations and Future Work.}}
The current model still exhibits constrained capabilities in local retouching. In future work, we plan to enhance the flexibility of localized editing by incorporating pixel-wise mask mechanisms into the framework, enabling more precise and region-specific image manipulation.

%% file: 12_Appendix.tex
\appendix
\section{DAPO-AE Reward Function Details}
Our DAPO-AE optimizes the model with only three simple yet effective rewards: format reward, image similarity reward, and image aesthetic reward.

\noindent\textbf{Format Reward ($R_f$)}: Ensures compliance with the structured reasoning template (\sref{sec:dataset}) 
and presence of three critical retouch tokens ($\langle|$Auto Retouch$|$$\rangle$, $\langle|$Style Retouch$|$$\rangle$, $\langle|$Param Retouch$|$$\rangle$). It is computed as:
\begin{equation}
     R_f = \frac{N_{\text{detected}}}{N_{\text{required}}} + \text{Penalty},
\end{equation}

where $N_{\text{detected}}$ is the number of detected structural tags (e.g., $\langle$problem light start$\rangle$), $N_{\text{required}}$ is the total number of required tags ($R_f \in [0,1]$), and Penalty = $-5$ if any critical retouch token is missing (0 otherwise).

\noindent\textbf{Similarity Reward ($R_s$)}: Aligning output image $I_{\text{out}}$ with the target’s retouching trends by reducing visual discrepancy via histogram similarity and $L_1$ loss. Computed as:
\begin{equation}
R_s = \gamma \cdot \text{HistSim}(I_{\text{out}}, I_{\text{tar}}) + (1-\gamma) \cdot \left( 1 - L_1(I_{\text{out}},I_{\text{tar}})\right),
\end{equation}
where $\text{HistSim}(\cdot)$ denotes cosine similarity of RGB histogram features~\cite{afifi2021histogan}, and the $L_1$ term is normalized to $[0,1]$.

\noindent\textbf{Aesthetic Reward ($R_a$)}: Explicitly optimizes aesthetic quality, activated only for the \textit{Auto-Retouch} task. Integrates LAION-V2.5~\cite{laionaesthetics} aesthetic score $S_{\rm LAION}$, Q-Align~\cite{wu2023qalign}) aesthetic score $S_{\rm Q-Align}$ and LIQE~\cite{zhang2023liqe} image quality score $S_{\rm LIQE}$:
\begin{equation}
\begin{split}
    R_a &= \alpha \cdot S_{\rm LAION}(I_{\text{out}}) + \beta \cdot S_{\rm Q\text{-}Align}(I_{\text{out}}) \\ 
    &\quad + (1-\alpha-\beta) \cdot S_{\rm LIQE}(I_{\text{out}}),
\end{split}
\end{equation}
where $\alpha$, $\beta$ are weighting coefficients, and all scores are normalized to $[0,1]$.

Specifically, we employ random alternating training where each step contains single-task samples exclusively to mitigate cross-task reward interference in multi-task training.

\section{More Details about \textit{AetherRetouch-1M+} Dataset}
\input{figs/texs/appendix_data}
\fref{fig:appendix_data} presents partial samples from each subset of the AetherRetouch-1M+ dataset. In the following, we provide more details on the construction process of each subset.
\subsection{More Details about \auto~Subset}
The \auto~sub-dataset consists of two complementary parts to ensure diversity and realism. The first part is synthesized with professional retouched image pairs as references, leveraging the synthetic framework detailed in Sec. \ref{sec:dataset} to generate high-quality, expert-aligned retouching samples. The second part is created through random operation perturbation-based degradation. To avoid irreversible visual artifacts caused by excessive perturbations, we first statistically analyze the variance of each retouching operation from a large-scale human retouching dataset. Based on these statistical results, we assign a specific standard deviation to each operation during the random perturbation process, which constrains the perturbation intensity within a reasonable range consistent with real-world retouching practices. The specific retouching operations adopted in this part and their corresponding standard deviations are summarized in Tab. \ref{tab:operations}.

\input{tables/operations}

\subsection{More Details about \style~Subset}
In constructing the \style~subset, we generated approximately 100 image pairs for training and testing for each of the 193 preset subcategories, ensuring balanced coverage across all stylistic directions. The specific workflow is as follows: For each subcategory among the 193 preset categories, we first randomly sample images from the Unsplash dataset to serve as unretouched inputs. These input images are then fed into Qwen3-VL for automatic category labeling, where the model assigns a semantic category tag (e.g., ``city'', ``forest'', ``pets'') based on image content. We predefine an allowed category label library for each preset subcategory, which specifies the image content types compatible with that style. If the automatically assigned category tag of an input image is included in the allowed category label library of the target preset subcategory, we randomly select one preset from this subcategory’s collection and apply it to the input image via the LightRoom API, generating the corresponding retouched output. This process ensures that each style preset is paired with semantically matching images, enhancing the relevance and effectiveness of the style retouching training data.

\subsection{More Details about \param~Subset}
As described in Sec. \ref{sec:dataset}, when constructing the \param~subset, we apply combinations of randomly perturbed retouching operations to unretouched images and generate corresponding target images via LightRoom. The specific operations adopted are summarized in Tab. \ref{tab:operations}. To cover a broader parameter domain, we employ a more aggressive sampling strategy: the standard deviation for Gaussian random sampling is set to either 25/100 or 35/100, ensuring sufficient diversity in parameter values. Additionally, we categorize all operations into three groups—Light (L) adjustment, Global Color (GC) adjustment, and Specific Color (SC) adjustment—consistent with the classification in Tab. \ref{tab:operations}. These three categories yield 7 distinct operation combinations: L, GC, SC, L+GC, L+SC, GC+SC, and L+GC+SC. During dataset construction, we assign corresponding binary masks to each combination, which enables the model to learn the disentanglement capability of the three core retouching dimensions during training.

\section{Reference-based Retouching Experiment Result of Retouch Encoder and Retouch Renderer}
To evaluate the performance of our Encoder-Renderer architecture, we constructed a test set of 700 samples, consisting of 100 samples for each of the 7 adjustment combinations derived from Lighting (L), Global Color (GC), and Specific Color (SC) adjustments (L, GC, SC, L+GC, L+SC, GC+SC, L+GC+SC). Each sample contains two image pairs that have undergone the same set of randomly sampled retouching operations.

Quantitative results are presented in Tab. \ref{tab:renderer-performance}. On single-adjustment test subsets, the model achieves high reconstruction accuracy: the GC setting delivers the best performance with an L1 distance of 0.0095 and a PSNR of 40.58 dB, while L and SC also yield strong results (PSNR $\geqslant$ 31 dB, SSIM $\geqslant$ 0.965). Performance decreases moderately on two-adjustment combinations, with PSNR values ranging from 27.42 dB to 32.05 dB. On the three-adjustment (L+GC+SC) subset, the PSNR reaches 27.48 dB. Visualization results are shown in \fref{fig:en_re_un}.

In practical use within the \auto~dataset construction pipeline, reference and input images often exhibit similar color histogram distributions. To assess performance under this scenario, we built an additional test set of 100 samples (all using L+GC+SC adjustments) where reference and input images have matched color distributions. As shown in Tab. \ref{tab:renderer-performance}, this test set shows significant improvements across all metrics, with PSNR increasing to 29.55 dB. This test set shows significant improvements across all metrics, with PSNR increasing to 29.55 dB. This enhancement is primarily driven by the improved performance of the Specific Color (SC) adjustment. In the original test set, the color distributions of reference and input images were mismatched, which meant the Encoder could not extract SC adjustments for colors present in the input but absent from the reference pair. By contrast, the matched color distributions in this additional test set largely mitigate this limitation, allowing the model to capture and transfer SC adjustments more accurately, thus leading to the observed performance gains. \fref{fig:en_re} presents the visualization results for this test set. Although a PSNR of 29.55 dB indicates the model does not perfectly transfer retouching effects from reference pairs to inputs in every case, this level of accuracy is sufficient for synthetic data generation. The inherent minor errors also act as random perturbations, enhancing the robustness of subsequent model training.
\input{tables/En-Re}
\input{figs/texs/en-re-un}
\input{figs/texs/en-re}

\section{Quantitative Comparison on Aether-Bench (Auto-Syn)}
\input{tables/auto-forward}
In addition to the FiveK-Bench designed for evaluating real-world scenario performance and the Aether-Bench (Auto) dedicated to testing generalization and robustness, we further construct the Aether-Bench (Auto-Syn) using a subset of synthetic data for the \auto~task. This benchmark consists of 250 retouching image pairs from different categories, which are sampled from the \auto~subset of the AetherRetouch-1M+ dataset. It is specifically used to evaluate the in-domain capability. Tab. \ref{tab:aether_bench_auto_forward_v2} presents the quantitative comparison results of our VeraRetouch against other state-of-the-art baselines on this benchmark. The experimental data clearly demonstrate that our method outperforms all competing approaches across core evaluation metrics, including histogram consistency, perceptual aesthetics, and texture preservation. This superiority fully validates the excellent in-domain retouching performance of the VeraRetouch framework.

\section{Quantitative Comparison on PPR10K-Bench}
\input{tables/ppr10k-bench}
To more comprehensively evaluate the Out-of-Distribution (OOD) performance of our algorithm, we constructed PPR10K-Bench (Auto) by randomly sampling 325 images from the PPR10K dataset. As shown in \tref{tab:ppr10k-bench}, our method leads in distribution-based and aesthetic metrics on PPR10K-Bench (Auto). The slightly lower reference-based scores, compared to our SOTA results on FiveK-Bench (Auto) in Tab.1, are primarily due to PPR10K's limited ground-truth diversity(3 vs. 5 in FiveK). In \auto~task with multiple plausible solutions, reference-based metrics are highly sensitive to reference diversity. Therefore, our superior performance in subjective and no-reference metrics better reflects our model’s true perceptual effectiveness.


\section{Human-Real Parameter-Retouch Evaluation}
\input{tables/human-expert-bench}
To further evaluate the robustness of our framework in professional workflows, we conducted an evaluation using expert-level metadata from the PPR10K dataset~\cite{liang2021ppr10k}. Specifically, we randomly selected 200 input images and collected their corresponding retouching metadata from three distinct expert styles (PPR10K-A, B, and C). From this metadata, we extracted 34 retouching parameters compatible with our renderer and applied them to the inputs to generate 600 Ground-Truth (GT) pairs. We then fed these expert-defined parameters into our model to predict the corresponding retouching effects.
As shown in \tref{tab:human-expert_bench}, we observe a performance degradation compared to the results on Aether-Bench (Param). Through further analysis, we identify a distribution gap between the two benchmarks. While Aether-Bench (Param) relies on independent Gaussian distribution modeling for each parameter, the parameter distributions in actual expert retouching are significantly more complex. In professional scenarios, parameters are often highly correlated. Constructing datasets with a more extensive distribution of retouching parameters remains a key priority to ensure the model can effectively capture and master a wider array of expert-level styles.

\section{Implementation Details.} 
The Retouch Encoder and Renderer are trained for $200\text{k}$ steps with a batch size of $16$ and a learning rate of $10^{-4}$. The domain-alignment pre-training and subsequent SFT stages are both conducted using the AdamW optimizer with a learning rate of $5 \times 10^{-5}$ for $600\text{k}$ and $500\text{k}$ steps, respectively, employing a balanced sampling strategy across datasets. For the DAPO-AE phase, we set the per-device batch size to $16$, with $8$ generations and $2$ steps per generation, training for a total of $13\text{k}$ steps. All experiments are implemented on two NVIDIA H20 GPUs.

\section{Ablation Study of User Instruction Perturbation}
\input{tables/instruction-purturbtion}
To enhance the model's generalization across diverse linguistic expressions, we employ user instruction perturbation in the \style~dataset to enrich the variety of user prompts. We conduct an ablation study to verify the effectiveness of this strategy. As reported in \tref{tab:ablation_instruction}, while instruction perturbation has a marginal impact on the performance of \auto~and \param~tasks, it significantly boosts the generalization capability in \style.

This improvement is particularly evident in the Aether-Bench (Style-InDistribution) test set, which consists of 250 pairs of unseen images applied with presets encountered during training. For this specific benchmark, the inclusion of instruction perturbation leads to a substantial performance leap, with the PSNR increasing from 22.82 to 29.11. These results demonstrate that diversifying user instructions effectively prevents the model from overfitting to specific linguistic patterns, thereby enabling it to more robustly generalize the learned stylistic transformations to unseen samples.

\section{Exploration of Multi-Round Retouching.}
Derived from our model’s structured chain-of-thought design (``image content analysis → problem analysis → retouching planning''), it inherently supports multi-round retouching for the Auto-Retouch task, where each iteration takes the output image from the previous round as its new input.

\fref{fig:multi-round} presents the multi-round inference results of our model on real-world camera-captured scenes. Take the sunset scene in the top row as an example: the initial input image features a muted, dim orange tone with the boat silhouette barely visible against the horizon. Through each successive round, the model iteratively boosts the sky’s contrast and saturation, with Round 3 revealing rich, layered gradients of warm orange and red in the clouds, making the sunset appear far more dramatic and the boat’s outline sharp and distinct. Similarly, the mountain landscape in the second row starts with underexposed shadows and a cool, flat blue palette; across rounds, the model progressively lifts the shadows to expose finer textures in the snow and rock faces, while enhancing the vibrancy of the sky’s blue, resulting in a scene with greater depth and visual impact.
\input{figs/texs/multi-round}

These results demonstrate that our structured reasoning process enables the model to iteratively refine its understanding of the image and its retouching strategy, leading to cumulative improvements in visual quality. The ability to perform incremental, multi-stage adjustments aligns with the iterative nature of professional retouching workflows, highlighting the practical value of our framework in real-world applications.

\section{Exploring the Potential of Video Retouching}
In this section, we further explore the potential of \textit{VeraRetouch} for automatic video retouching. We selected 5–15 second video clips, randomly sampled one frame from each clip as a reference key frame, and used \textit{VeraRetouch} to infer retouching effects and extract corresponding retouching latents. These latents were then applied to all frames of the input video via our Retouch Renderer. Derived from the lightweight design of the Retouch Renderer, this entire process can be completed efficiently.
\input{figs/texs/video}

Frame-by-frame comparisons of the input and output videos are visualized in \fref{fig:video}. The results demonstrate excellent temporal consistency and fidelity, with no flickering or frame-to-frame artifacts—common issues in existing video editing models. This robustness stems from the per-pixel color mapping design of our Retouch Renderer, which applies consistent adjustments while preserving structural details across all frames, ensuring a smooth and coherent visual experience throughout the video.

\section{Validation of Retouching Capability on Ultra-High-Resolution Images}
To verify the effectiveness of VeraRetouch on ultra-high-resolution (UHR) images, we tested the model on two 6000×3376 pixel (over 4K resolution) photographs: a coastal lighthouse scene and a low-light sunset landscape. Both input images were captured using a standard camera in real-world conditions, ensuring the test scenarios reflect practical photography workflows.

In the lighthouse example, the model enhanced the muted color palette to produce vivid cerulean water and a high-contrast lighthouse, while preserving details like the stone embankment and distant mountain texture. For the sunset scene, it lifted dark, murky tones to reveal vibrant gradients of blue and orange in the sky, enhancing the sense of luminous depth, as well as richer golden and green hues in the water, all while maintaining the delicate structure of the sunbeams. These results confirm that \textit{VeraRetouch} can deliver professional retouching on UHR images across diverse scenarios, preserving fine details while significantly improving visual appeal.
\input{figs/texs/4k-1}
\input{figs/texs/4k-2}

\section{Full Retouching Process Examples}
In this section, we present complete input-to-output examples of three retouching tasks: \auto, \style~and \param.

\noindent\textbf{\auto~(Figs. \ref{fig:auto_r1}-\ref{fig:auto_r7})}: In this task, the user provides only an input image without any additional instructions. The model autonomously generates a structured reasoning process, including a \textit{Content Overview}, \textit{Problem Analysis}, \textit{Retouch Plans}, and \textit{Retouch Tokens}, alongside the final retouched image.

\noindent\textbf{\style~(Figs. \ref{fig:style_r1}-\ref{fig:style_r7})}: Here, the user provides both an input image and a text prompt specifying the desired retouching style. The model produces a structured reasoning process, including a \textit{Content Overview}, \textit{Retouch Plans}, and \textit{Retouch Tokens}, followed by the stylistically enhanced output image.

\noindent\textbf{\param~(Figs. \ref{fig:param_r1}-\ref{fig:param_r7})}: For this task, the user provides an input image and explicit retouching operations' parameters. The model outputs a reasoning process consisting of \textit{Retouch Plans} and \textit{Retouch Tokens}, along with the adjusted final image.

\clearpage
\input{figs/texs/auto_result-s1}
\input{figs/texs/auto_result-s2}
\input{figs/texs/auto_result-s3}
\input{figs/texs/auto_result-s4}
\input{figs/texs/auto_result-s5}
\input{figs/texs/auto_result-s6}
\input{figs/texs/auto_result-s7}

\input{figs/texs/style_result-s1}
\input{figs/texs/style_result-s2}
\input{figs/texs/style_result-s3}
\input{figs/texs/style_result-s4}
\input{figs/texs/style_result-s5}
\input{figs/texs/style_result-s6}
\input{figs/texs/style_result-s7}

\input{figs/texs/professional_result-s1}
\input{figs/texs/professional_result-s2}
\input{figs/texs/professional_result-s3}
\input{figs/texs/professional_result-s4}
\input{figs/texs/professional_result-s5}
\input{figs/texs/professional_result-s6}
\input{figs/texs/professional_result-s7}

%% file: figs/texs/appendix_data.tex
\begin{figure*}
    \centering
  \includegraphics[width=\textwidth]{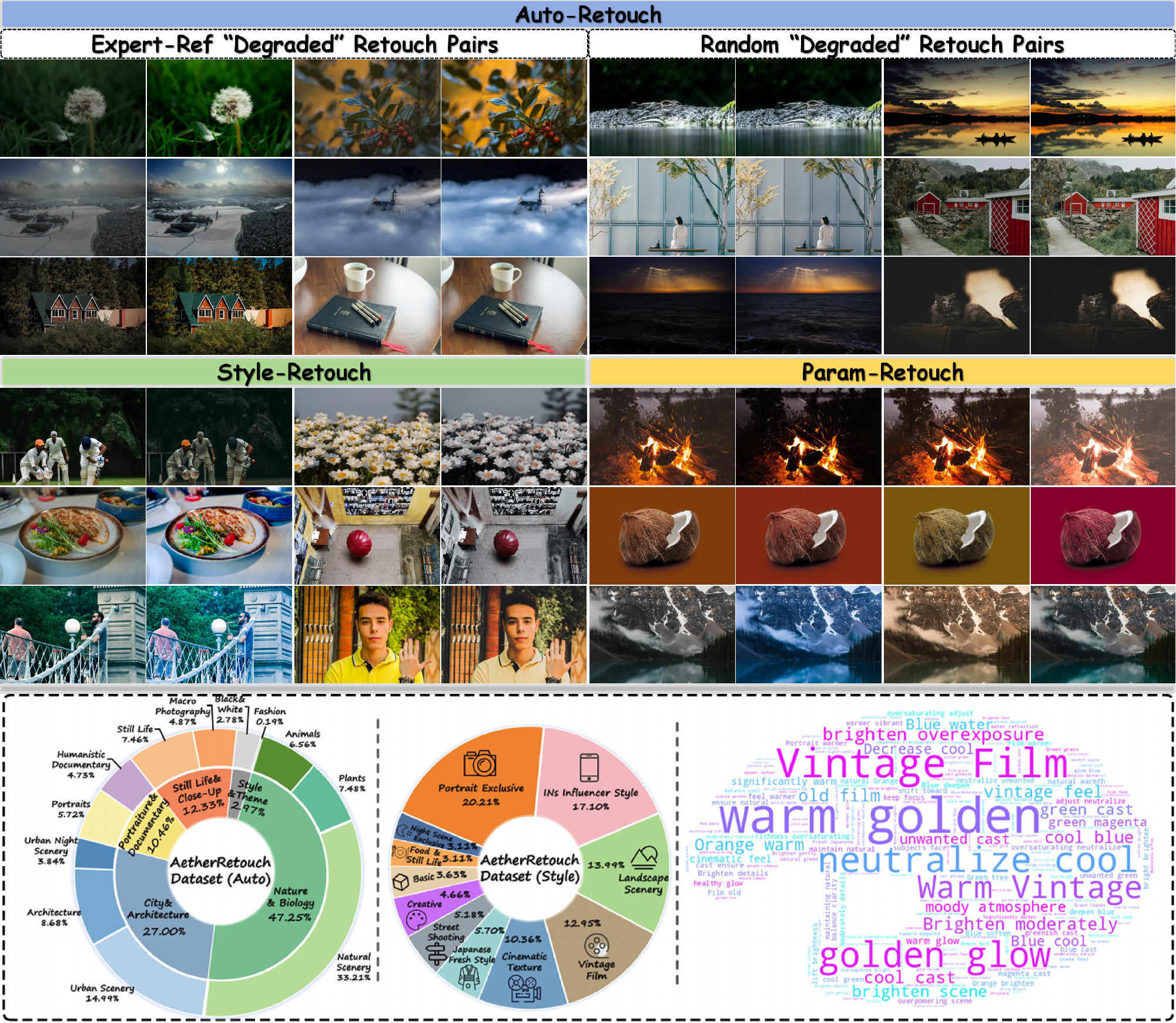}
  \caption{Visualization of the \textit{AetherRetouch-1M+} dataset. The upper part presents some retouching pairs for each dataset, covering diverse scenes and retouching requirements. The bottom-left subfigure is a donut chart showing the category distribution of the dataset and preset used in the \textit{Style-Retouch} subdataset. The bottom-right subfigure is a word cloud visualization of high-frequency terms in the retouching instructions, intuitively displaying core semantic elements that guide the retouching process.}
  \label{fig:appendix_data}
\end{figure*}

%% file: tables/operations.tex
\begin{table}[htbp]
    \centering
    \caption{Detailed LightRoom adjustment operations' name, range, and standard deviation we use.}
    \label{tab:operations}
    \footnotesize 
    \resizebox{0.95\linewidth}{!}{%
    \begin{tabular}{l c c}
        \toprule
        \textbf{Parameter Name} & \textbf{Range} & \textbf{Standard Deviation} \\
        \midrule
        
        \multicolumn{3}{c}{\cellcolor[gray]{0.9}\textbf{Light Adjustment}} \\
        Exposure2012 & [-5.0, 5.0] & $\pm$ 0.6543 \\
        Contrast2012 & [-100, 100] & $\pm$ 12.6789 \\
        Highlights2012 & [-100, 100] & $\pm$ 21.5888 \\
        Shadows2012 & [-100, 100] & $\pm$ 16.2265 \\
        Whites2012 & [-100, 100] & $\pm$ 16.4355 \\
        Blacks2012 & [-100, 100] & $\pm$ 15.5995 \\
        ParametricShadows & [-100, 100] & $\pm$ 7.2495 \\
        ParametricDarks & [-100, 100] & $\pm$ 15.8214 \\
        ParametricLights & [-100, 100] & $\pm$ 7.6688 \\
        ParametricHighlights & [-100, 100] & $\pm$ 9.1287 \\
        \midrule
        
        \multicolumn{3}{c}{\cellcolor[gray]{0.9}\textbf{Global Color Adjustment}} \\
        IncrementalTemperature & [-100, 100] & $\pm$ 15.0000 \\
        IncrementalTint & [-100, 100] & $\pm$ 15.0000 \\
        Vibrance & [-100, 100] & $\pm$ 7.8137 \\
        Saturation & [-100, 100] & $\pm$ 7.4315 \\
        \midrule
        
        \multicolumn{3}{c}{\cellcolor[gray]{0.9}\textbf{Specific Color Adjustment}} \\
        HueAdjustmentRed & [-100, 100] & $\pm$ 5.8140 \\
        HueAdjustmentOrange & [-100, 100] & $\pm$ 8.3549 \\
        HueAdjustmentYellow & [-100, 100] & $\pm$ 15.1914 \\
        HueAdjustmentGreen & [-100, 100] & $\pm$ 8.4875 \\
        HueAdjustmentAqua & [-100, 100] & $\pm$ 19.8922 \\
        HueAdjustmentBlue & [-100, 100] & $\pm$ 11.8419 \\
        HueAdjustmentPurple & [-100, 100] & $\pm$ 10.1451 \\
        HueAdjustmentMagenta & [-100, 100] & $\pm$ 19.0949 \\
        \addlinespace[2pt]
        SaturationAdjustmentRed & [-100, 100] & $\pm$ 19.8318 \\
        SaturationAdjustmentOrange & [-100, 100] & $\pm$ 9.6656 \\
        SaturationAdjustmentYellow & [-100, 100] & $\pm$ 18.2479 \\
        SaturationAdjustmentGreen & [-100, 100] & $\pm$ 17.7113 \\
        SaturationAdjustmentAqua & [-100, 100] & $\pm$ 7.4975 \\
        SaturationAdjustmentBlue & [-100, 100] & $\pm$ 15.6967 \\
        SaturationAdjustmentPurple & [-100, 100] & $\pm$ 21.7025 \\
        SaturationAdjustmentMagenta & [-100, 100] & $\pm$ 27.8002 \\
        \addlinespace[2pt]
        LuminanceAdjustmentRed & [-100, 100] & $\pm$ 10.0289 \\
        LuminanceAdjustmentOrange & [-100, 100] & $\pm$ 13.4234 \\
        LuminanceAdjustmentYellow & [-100, 100] & $\pm$ 16.2116 \\
        LuminanceAdjustmentGreen & [-100, 100] & $\pm$ 28.3202 \\
        LuminanceAdjustmentAqua & [-100, 100] & $\pm$ 17.1250 \\
        LuminanceAdjustmentBlue & [-100, 100] & $\pm$ 22.4162 \\
        LuminanceAdjustmentPurple & [-100, 100] & $\pm$ 18.2913 \\
        LuminanceAdjustmentMagenta & [-100, 100] & $\pm$ 25.4936 \\
        
        \bottomrule
    \end{tabular}
    }
\end{table}

%% file: tables/En-Re.tex
\begin{table}[t]
\centering
\caption{Quantitative evaluation of the Retouch Renderer under different adjustment settings. The metrics represent the reconstruction accuracy between the rendered output and the ground truth.}
\label{tab:renderer-performance}
\small
\begin{tabular}{@{}cccc@{}}
\toprule
\textbf{Adjustment Setting} & \textbf{L1 Distance} $\downarrow$ & \textbf{PSNR (dB)} $\uparrow$ & \textbf{SSIM} $\uparrow$ \\ \midrule
L                           & 0.0208                 & 33.41              & 0.965         \\
GC                          & 0.0095                 & 40.58              & 0.985         \\
SC                          & 0.0245                 & 31.29              & 0.966         \\ \midrule
L + GC                      & 0.0263                 & 30.75              & 0.947         \\
L + SC                      & 0.0406                 & 27.42              & 0.931         \\
GC + SC                     & 0.0237                 & 32.05              & 0.965         \\ \midrule
L + GC + SC                 & 0.0413                 & 27.48              & 0.929         \\
L + GC + SC (Similar Ref)   & 0.0278                 & 29.55              & 0.944         \\ \bottomrule
\addlinespace[1ex]
\multicolumn{4}{l}{\footnotesize L: Lighting, GC: Global Color, SC: Specific Color.}
\end{tabular}
\end{table}

%% file: figs/texs/en-re-un.tex
\begin{figure}
    \centering
  \includegraphics[width=0.48\textwidth]{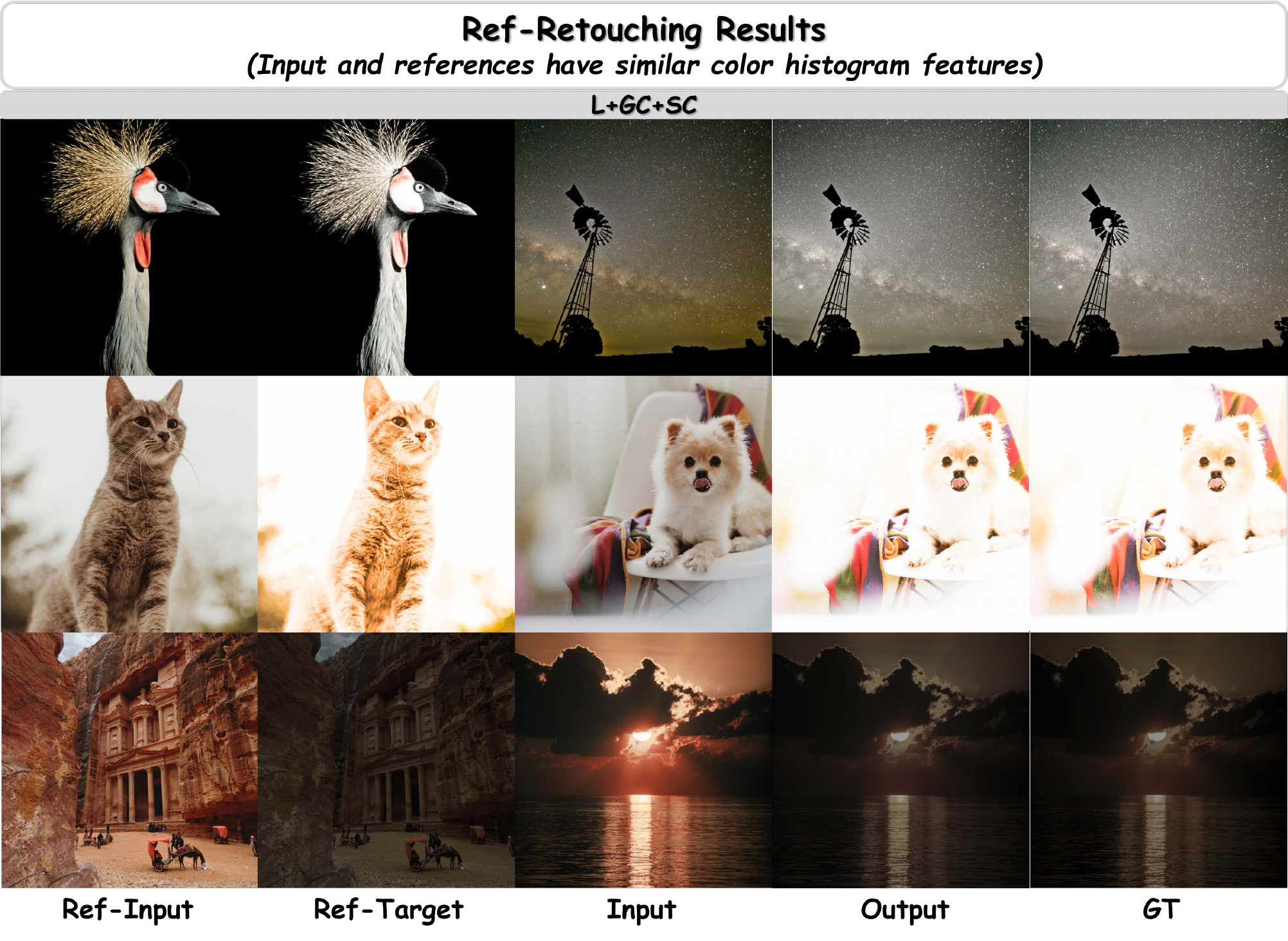}
  \caption{Visualization of reference-based retouching results(Input-GT pair and reference pair are randomly selected but share the same operation parameters).}
  \label{fig:en_re_un}
\end{figure}

%% file: figs/texs/en-re.tex
\begin{figure}
    \centering
  \includegraphics[width=0.48\textwidth]{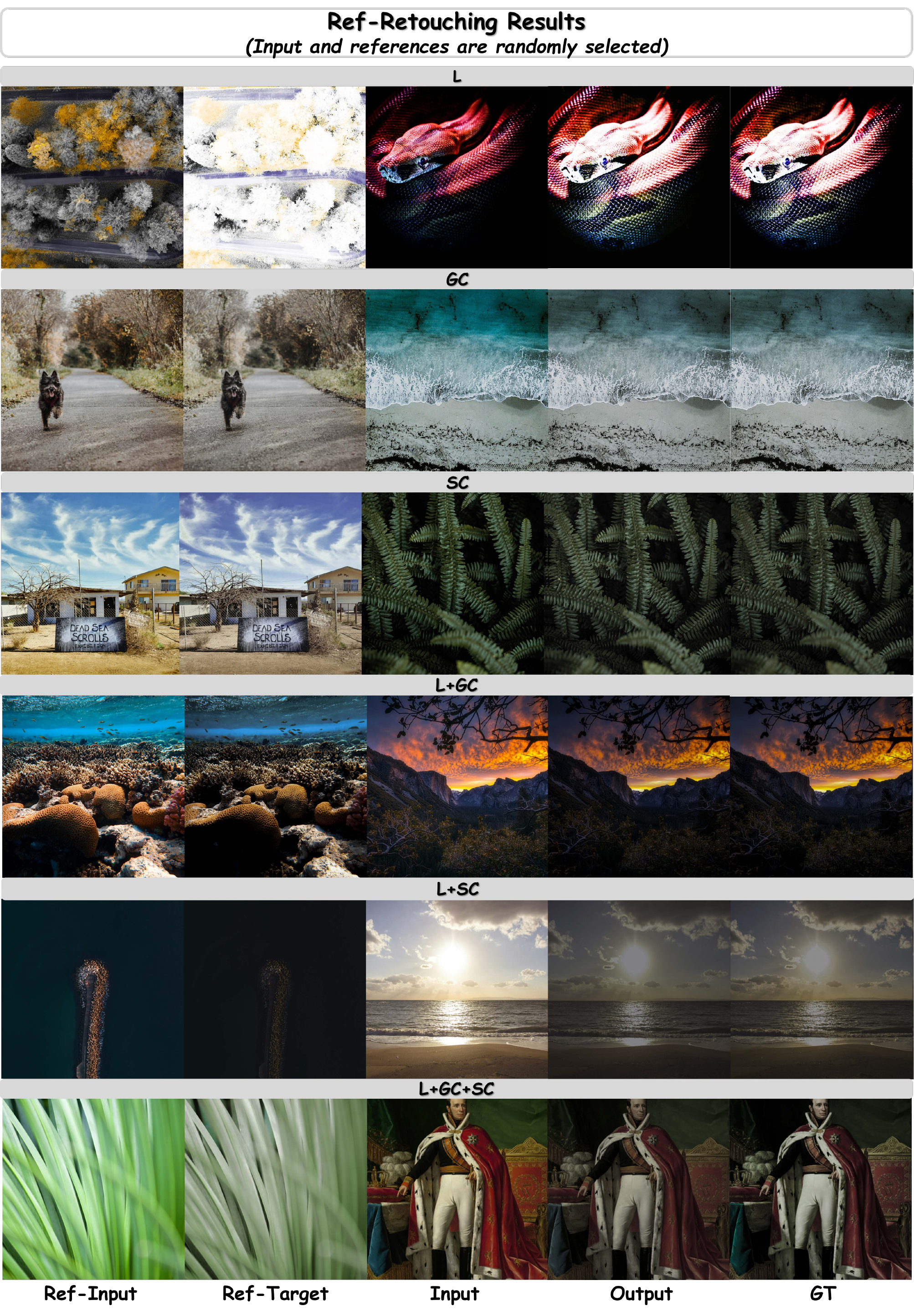}
  \caption{Visualization of reference-based retouching results(Input-GT pair and reference pair have similar color histogram features and share the same operation parameters).}
  \label{fig:en_re}
\end{figure}

%% file: tables/auto-forward.tex
\begin{table*}[t]
\centering
\caption{\textbf{Quantitative Comparison on Aether-bench (Auto-Syn).} We report histogram consistency, aesthetic quality, and texture preservation metrics. \textcolor{red}{Red} and \textcolor{blue}{Blue} indicate the best and second-best results, respectively. Values after $\pm$ denote the standard deviation.}
\label{tab:aether_bench_auto_forward_v2}

\footnotesize
\resizebox{\textwidth}{!}{%
\begin{tabular}{l|cccc|ccc|ccc}
\toprule
\multirow{2}{*}{\textbf{Method}} & \multicolumn{4}{c|}{\textbf{Histogram Consistency}} & \multicolumn{3}{c|}{\textbf{Aesthetic Quality}} & \multicolumn{3}{c}{\textbf{Texture Preservation}} \\
 & Hist-L $\uparrow$ & Hist-C $\uparrow$ & Hist-S $\uparrow$ & Hist-M $\uparrow$ & LAION $\uparrow$ & Q-Align $\uparrow$ & LIQE $\uparrow$ & DISTS $\downarrow$ & GMSD $\downarrow$ & TD $\downarrow$ \\
\midrule
\rowcolor{gray!10} \multicolumn{11}{c}{\textit{Aether-bench (Auto-Syn)}} \\
RSFNet & \textcolor{red}{90.95\%} & 87.52\% & \textcolor{blue}{88.28\%} & 88.92\% & 6.37{\scriptsize$\pm$0.49} & \textcolor{blue}{4.05{\scriptsize$\pm$0.57}} & 3.27{\scriptsize$\pm$1.06} & \textcolor{red}{0.046} & 0.044 & \textcolor{red}{0.412} \\
Nano Banana & 84.96\% & 74.70\% & 64.33\% & 74.67\% & 6.36{\scriptsize$\pm$0.57} & \textcolor{red}{4.07{\scriptsize$\pm$0.55}} & 3.13{\scriptsize$\pm$1.02} & 0.115 & 0.137 & 1.372 \\
Flux.1 Kontext & 88.88\% & \textcolor{blue}{91.72\%} & 86.78\% & \textcolor{blue}{89.12\%} & 6.32{\scriptsize$\pm$0.43} & 3.99{\scriptsize$\pm$0.56} & \textcolor{blue}{3.34{\scriptsize$\pm$1.08}} & 0.071 & 0.055 & 0.604 \\
Qwen-Image-2509 & 64.98\% & 55.95\% & 65.74\% & 62.22\% & 5.75{\scriptsize$\pm$0.50} & 3.66{\scriptsize$\pm$0.61} & 2.57{\scriptsize$\pm$0.93} & 0.153 & 0.095 & 1.068 \\
MonetGPT & 81.70\% & 70.35\% & 78.97\% & 77.01\% & 5.76{\scriptsize$\pm$0.55} & 3.75{\scriptsize$\pm$0.62} & 2.68{\scriptsize$\pm$1.04} & 0.116 & \textcolor{red}{0.036} & 0.507 \\
JarvisArt & 71.35\% & 57.38\% & 71.53\% & 66.75\% & 5.70{\scriptsize$\pm$0.45} & 3.61{\scriptsize$\pm$0.56} & 2.28{\scriptsize$\pm$0.82} & 0.155 & \textcolor{blue}{0.041} & 0.615 \\
\midrule
\textbf{Ours-SFT} & \textcolor{blue}{89.22\%} & \textcolor{red}{92.77\%} & \textcolor{red}{88.57\%} & \textcolor{red}{90.19\%} & \textcolor{red}{6.47{\scriptsize$\pm$0.42}} & \textcolor{red}{4.07{\scriptsize$\pm$0.56}} & \textcolor{red}{3.46{\scriptsize$\pm$1.03}} & \textcolor{blue}{0.048} & 0.066 & 0.537 \\
\textbf{Ours-DAPO-AE} & 88.95\% & 88.40\% & 82.08\% & 86.48\% & \textcolor{blue}{6.41{\scriptsize$\pm$0.42}} & 4.01{\scriptsize$\pm$0.56} & 3.29{\scriptsize$\pm$1.02} & 0.053 & 0.044 & \textcolor{blue}{0.435} \\
\bottomrule
\end{tabular}%
}
\end{table*}

%% file: tables/ppr10k-bench.tex
\begin{table*}[t]
\centering
\caption{\textbf{Quantitative Comparison on PPR10K-Bench.} We report histogram consistency, aesthetic quality, and texture preservation metrics. \textcolor{red}{Red} and \textcolor{blue}{Blue} indicate the best and second-best results, respectively. Values after $\pm$ denote the standard deviation.}
\label{tab:ppr10k-bench}
\resizebox{\textwidth}{!}{%
\begin{tabular}{l|ccc|cccc|ccc|ccc}
\toprule
\textbf{Method} & PSNR$\uparrow$ & SSIM$\uparrow$ & LPIPS$\downarrow$ & Hist-L$\uparrow$ & Hist-C$\uparrow$ & Hist-S$\uparrow$ & Hist-M$\uparrow$ & LAION$\uparrow$ & Q-Align$\uparrow$ & LIQE$\uparrow$ & DISTS$\downarrow$ & GMSD$\downarrow$ & TD$\downarrow$ \\
\midrule
AdaInt & \textcolor{blue}{24.75} & \textcolor{red}{0.939} & \textcolor{red}{0.046} & 88.79\% & 87.09\% & 91.44\% & 89.11\% & 6.56$\pm$0.46 & 4.11$\pm$0.38 & \textcolor{red}{4.28$\pm$0.74} & \textcolor{red}{0.041} & 0.021 & 0.271 \\
RSFNet & 23.08 & 0.921 & 0.059 & 83.85\% & 87.83\% & \textcolor{red}{92.99\%} & 88.22\% & \textcolor{red}{6.60$\pm$0.44} & \textcolor{red}{4.48$\pm$0.31} & 4.19$\pm$0.77 & 0.050 & \textcolor{blue}{0.012} & \textcolor{red}{0.219} \\
NamedCurves & 22.85 & 0.919 & 0.071 & 82.16\% & 87.56\% & 91.13\% & 86.95\% & 6.46$\pm$0.44 & \textcolor{red}{4.48$\pm$0.32} & 4.18$\pm$0.80 & 0.050 & \textcolor{red}{0.009} & \textcolor{blue}{0.265} \\
Nano Banana & 19.69 & 0.683 & 0.106 & 87.68\% & \textcolor{blue}{90.62\%} & 91.13\% & 89.81\% & 6.43$\pm$0.43 & \textcolor{blue}{4.46$\pm$0.32} & 4.20$\pm$0.82 & 0.061 & 0.127 & 1.248 \\
Flux.1 Kontext & \textcolor{red}{25.34} & \textcolor{blue}{0.936} & 0.062 & 88.63\% & \textcolor{red}{95.69\%} & 87.29\% & \textcolor{blue}{90.53\%} & 6.31$\pm$0.44 & 4.15$\pm$0.29 & 4.19$\pm$0.71 & 0.056 & 0.019 & 0.414 \\
Qwen-Image-2509 & 17.05 & 0.520 & 0.159 & 86.90\% & 88.72\% & 86.47\% & 87.37\% & 6.35$\pm$0.45 & 4.29$\pm$0.36 & 4.08$\pm$0.87 & 0.084 & 0.203 & 1.658 \\
MonetGPT & 21.58 & 0.873 & 0.095 & 91.05\% & 89.82\% & 83.74\% & 88.21\% & 6.23$\pm$0.52 & 4.30$\pm$0.39 & 3.97$\pm$0.95 & 0.074 & 0.046 & 0.539 \\
JarvisArt & 21.79 & 0.872 & 0.109 & 89.37\% & 84.99\% & 87.61\% & 87.32\% & 6.01$\pm$0.46 & 4.45$\pm$0.33 & 3.79$\pm$0.86 & 0.086 & 0.046 & 0.623 \\
\midrule
\textbf{Ours-SFT} & 23.88 & 0.922 & 0.065 & \textcolor{blue}{91.69\%} & 90.58\% & 81.70\% & 87.99\% & 6.53$\pm$0.46 & \textcolor{blue}{4.46$\pm$0.30} & 4.19$\pm$0.77 & 0.054 & 0.034 & 0.443 \\
\textbf{Ours-DAPO-AE} & 24.43 & 0.930 & \textcolor{blue}{0.055} & \textcolor{red}{94.22\%} & 88.46\% & \textcolor{blue}{92.79\%} & \textcolor{red}{91.82\%} & \textcolor{blue}{6.59$\pm$0.44} & \textcolor{red}{4.48$\pm$0.29} & \textcolor{blue}{4.21$\pm$0.77} & \textcolor{blue}{0.047} & 0.022 & 0.331 \\
\bottomrule
\end{tabular}%
}
\end{table*}

%% file: tables/human-expert-bench.tex
\begin{table}[t]
\centering
\caption{Human-Real Parameter-Retouch Quantitative Evaluation}
\label{tab:human-expert_bench}
\resizebox{\columnwidth}{!}{%
\setlength{\tabcolsep}{2.5pt} 
\begin{tabular}{lc|ccccccc}
\toprule
\textbf{Method} & \textbf{Human Expert} & \multicolumn{7}{c}{\textbf{Experimental Results}} \\

\midrule
\rowcolor{gray!5} & & \boldmath{$L_1 \downarrow$} & \textbf{PSNR$\uparrow$} & \textbf{SSIM$\uparrow$} & \textbf{LPIPS$\downarrow$} & \textbf{DISTS$\downarrow$} & \textbf{GMSD$\downarrow$} & \textbf{TD$\downarrow$} \\
\multirow{3}{*}{\textbf{Ours-SFT}}  
& A & 0.054 & 23.401 & 0.900 & 0.110 & 0.091 & 0.030 & 0.404 \\
& B & 0.053 & 23.632 & 0.897 & 0.109 & 0.089 & 0.032 & 0.400 \\
& C & 0.062 & 22.402 & 0.887 & 0.114 & 0.088 & 0.024 & 0.361 \\
\midrule
\multirow{3}{*}{\textbf{Ours-DAPO}} 
& A & 0.052 & 23.684 & 0.905 & 0.098 & 0.081 & 0.027 & 0.384 \\
& B & 0.057 & 23.115 & 0.893 & 0.109 & 0.086 & 0.026 & 0.369 \\
& C & 0.062 & 22.426 & 0.889 & 0.108 & 0.082 & 0.021 & 0.335 \\

\bottomrule
\end{tabular}%
}
\end{table}

%% file: tables/instruction-purturbtion.tex
\begin{table}[t]
\centering
\caption{Ablation Study esults of User Instruction Perturbation.}
\label{tab:ablation_instruction}
\resizebox{\columnwidth}{!}{%
\setlength{\tabcolsep}{3pt} 
\begin{tabular}{c|ccccccc} 
\toprule
\textbf{Instruction Perturbation} & \multicolumn{7}{c}{\textbf{Experimental Results}} \\
\midrule

\rowcolor{gray!15} \multicolumn{8}{c}{\textit{FiveK-Bench (Auto-Retouching)}} \\
\rowcolor{gray!5} & \textbf{Hist-M$\bm{\uparrow}$} & \textbf{LAION$\bm{\uparrow}$} & \textbf{Q-Align$\bm{\uparrow}$} & \textbf{LIQE$\bm{\uparrow}$} & \textbf{DISTS$\bm{\downarrow}$} & \textbf{GMSD$\bm{\downarrow}$} & \textbf{TD$\bm{\downarrow}$} \\
\textbf{w/}  & 92.83\% & 5.13 & 4.18 & 3.92 & 0.040 & 0.061 & 0.694 \\
\textbf{w/o} & 94.54\% & 5.10 & 4.16 & 3.89 & 0.039 & 0.046 & 0.602 \\

\midrule
\rowcolor{gray!15} \multicolumn{8}{c}{\textit{Aether-Bench (Style-Retouching)}} \\
\rowcolor{gray!5} & \boldmath{$L_1 \bm{\downarrow}$} & \textbf{PSNR$\bm{\uparrow}$} & \textbf{SSIM$\bm{\uparrow}$} & \textbf{LPIPS$\bm{\downarrow}$} & \textbf{DISTS$\bm{\downarrow}$} & \textbf{GMSD$\bm{\downarrow}$} & \textbf{TD$\bm{\downarrow}$} \\
\textbf{w/}  & 0.0970 & 19.73 & 0.839 & 0.149 & 0.100 & 0.039 & 0.592 \\
\textbf{w/o} & 0.0923 & 19.55 & 0.821 & 0.155 & 0.113 & 0.040 & 0.598 \\

\midrule
\rowcolor{gray!15} \multicolumn{8}{c}{\textit{Aether-Bench (Style-InDistribution)}} \\
\rowcolor{gray!5} & \boldmath{$L_1 \bm{\downarrow}$} & \textbf{PSNR$\bm{\uparrow}$} & \textbf{SSIM$\bm{\uparrow}$} & \textbf{LPIPS$\bm{\downarrow}$} & \textbf{DISTS$\bm{\downarrow}$} & \textbf{GMSD$\bm{\downarrow}$} & \textbf{TD$\bm{\downarrow}$} \\
\textbf{w/}  & 0.034 & 29.11 & 0.939 & 0.072 & 0.072 & 0.043 & 0.551 \\
\textbf{w/o} & 0.066 & 22.82 & 0.902 & 0.107 & 0.078 & 0.040 & 0.544 \\

\midrule
\rowcolor{gray!15} \multicolumn{8}{c}{\textit{Aether-Bench (Param-Retouching)}} \\
\rowcolor{gray!5} & \boldmath{$L_1 \bm{\downarrow}$} & \textbf{PSNR$\bm{\uparrow}$} & \textbf{SSIM$\bm{\uparrow}$} & \textbf{LPIPS$\bm{\downarrow}$} & \textbf{DISTS$\bm{\downarrow}$} & \textbf{GMSD$\bm{\downarrow}$} & \textbf{TD$\bm{\downarrow}$} \\
\textbf{w/}  & 0.023 & 30.39 & 0.946 & 0.039 & 0.040 & 0.071 & 0.664 \\
\textbf{w/o} & 0.025 & 29.94 & 0.944 & 0.044 & 0.045 & 0.067 & 0.628 \\

\bottomrule
\end{tabular}%
}
\end{table}

%% file: figs/texs/multi-round.tex
\begin{figure*}
    \centering
  \includegraphics[width=\textwidth]{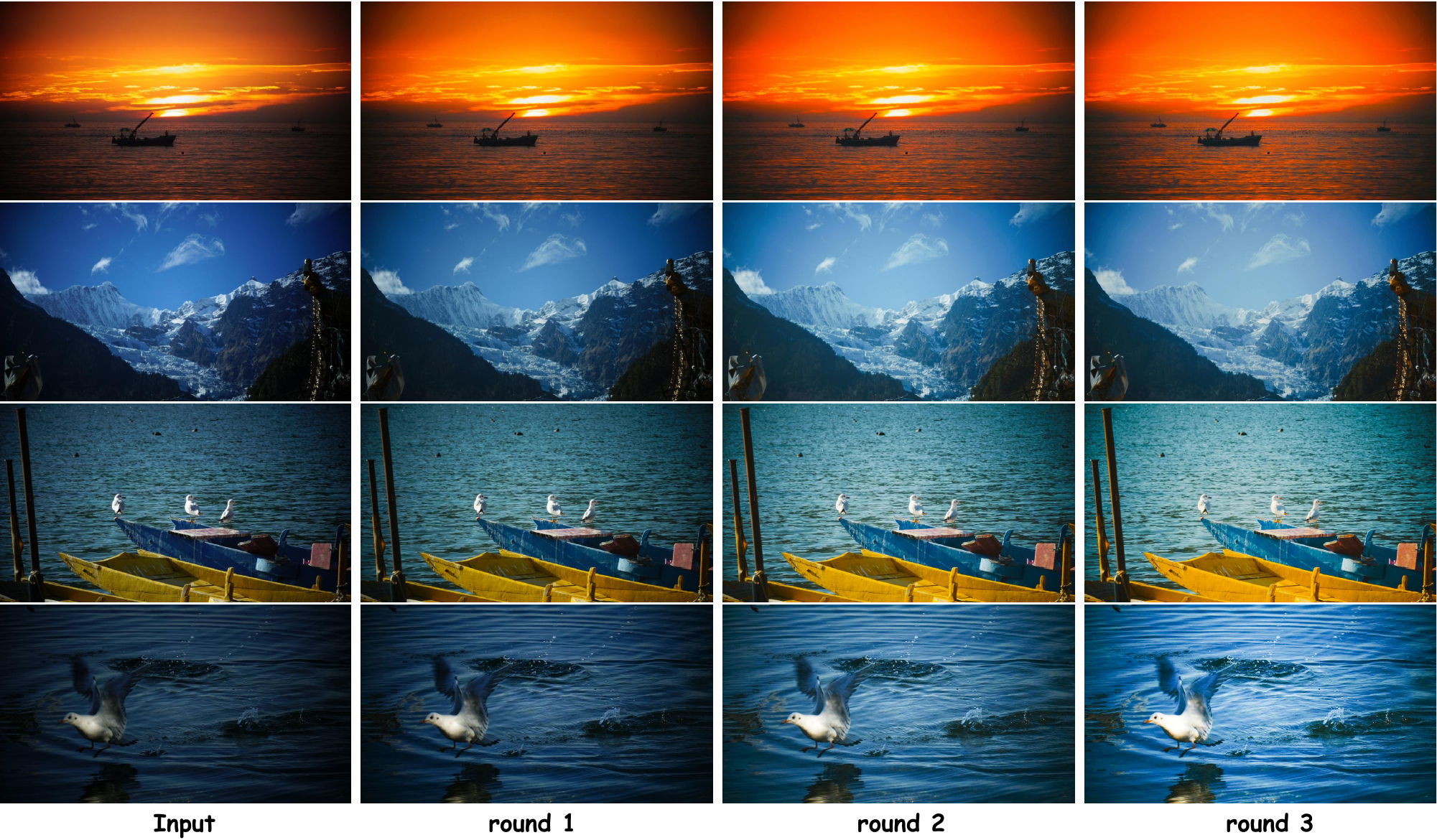}
  \caption{Visual results of multi-round inference. In each round, \textbf{VeraRetouch} takes the output image from the previous round as its new input.}
  \label{fig:multi-round}
\end{figure*}

%% file: figs/texs/video.tex
\begin{figure*}
  \includegraphics[width=\textwidth]{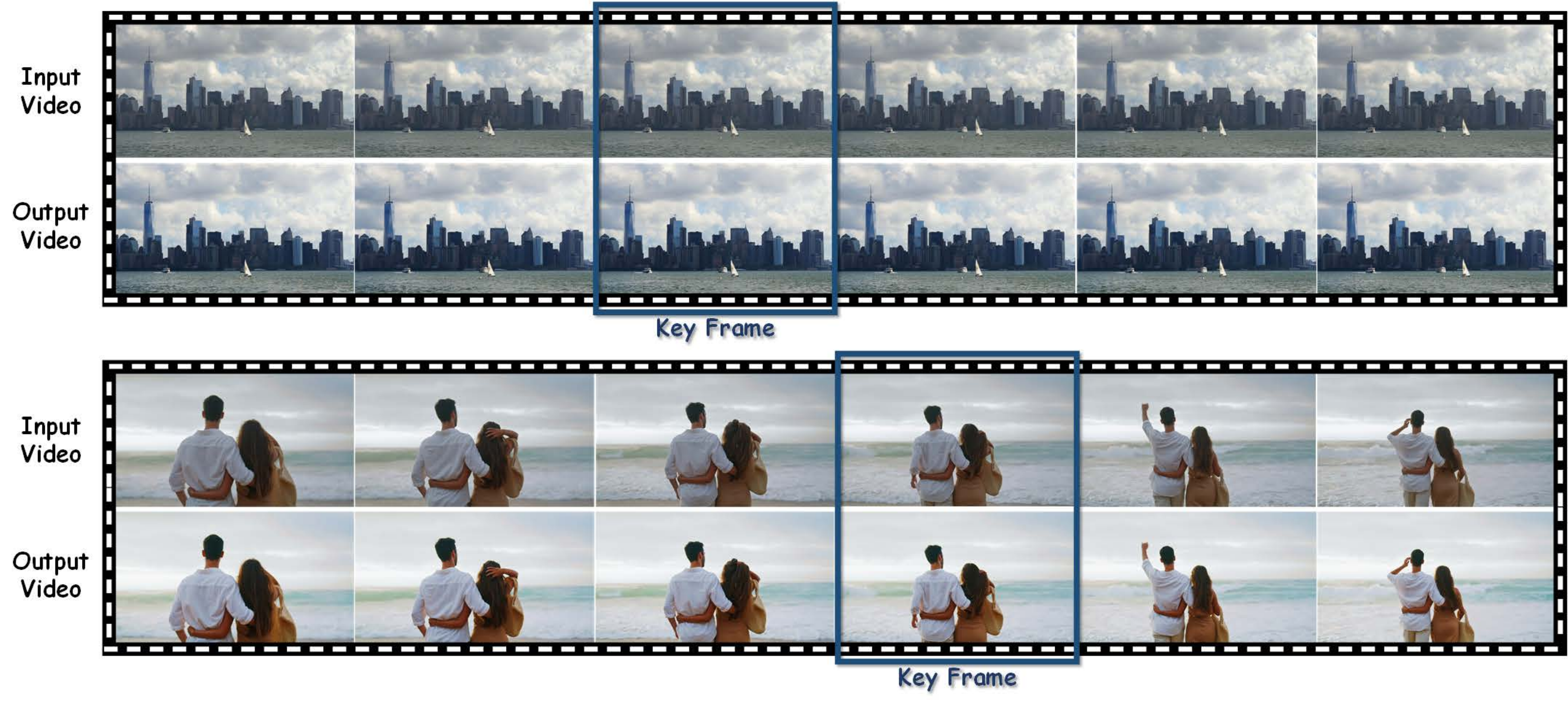}
  \caption{
   Video retouching results. The key frame (highlighted) is automatically retouched by \textbf{VeraRetouch} to generate retouching latents, which are then applied to all frames. The output videos exhibit consistent, artifact-free enhancements with no temporal flickering.
  }
  \label{fig:video}
\end{figure*}

%% file: figs/texs/4k-1.tex
\begin{figure*}
    \centering
  \includegraphics[width=\textwidth]{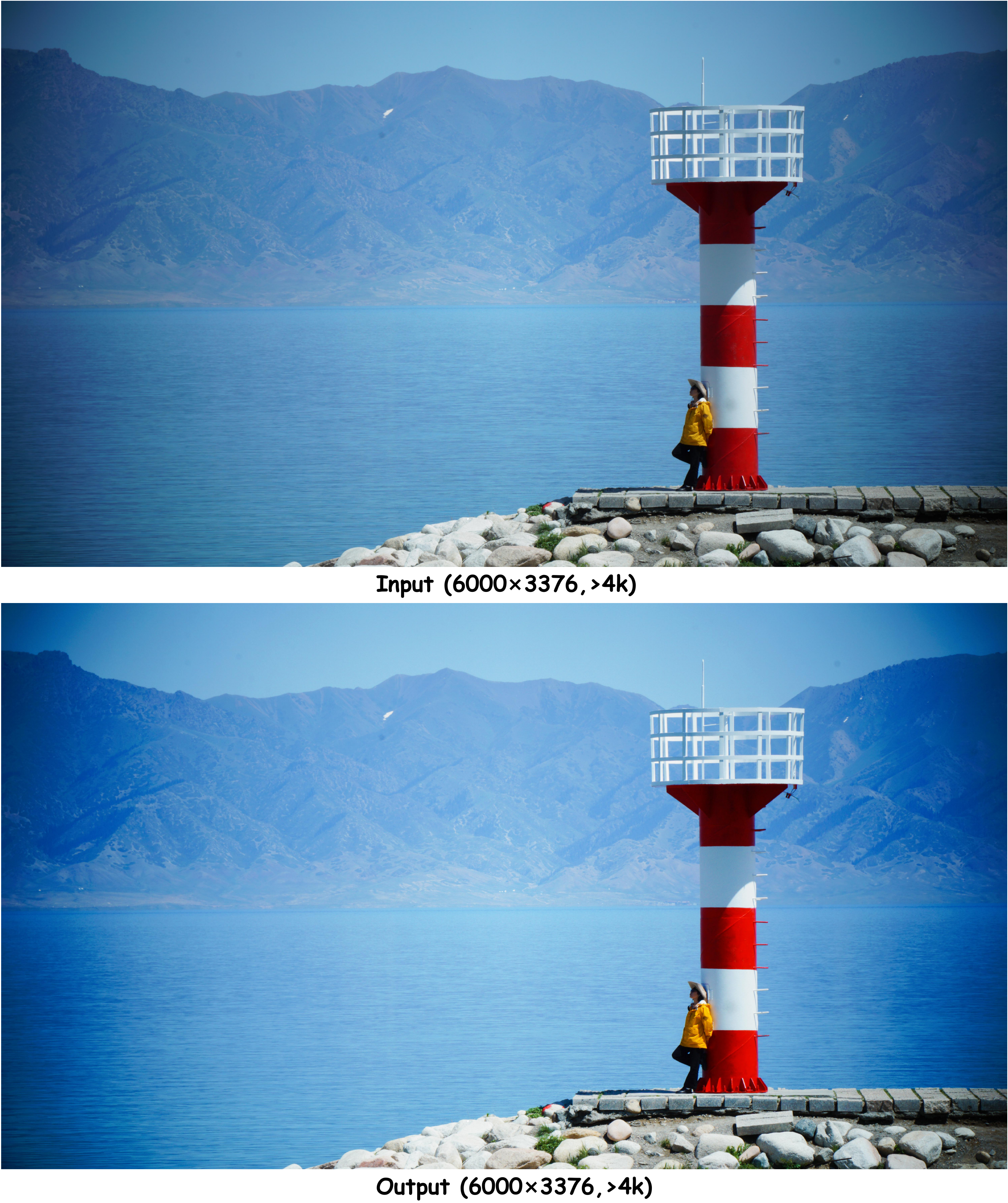}
  \caption{Retouching result on a 6000×3376 (over 4K) ultra-high-resolution image.}
  \label{fig:4k-1}
\end{figure*}

%% file: figs/texs/4k-2.tex
\begin{figure*}
    \centering
  \includegraphics[width=\textwidth]{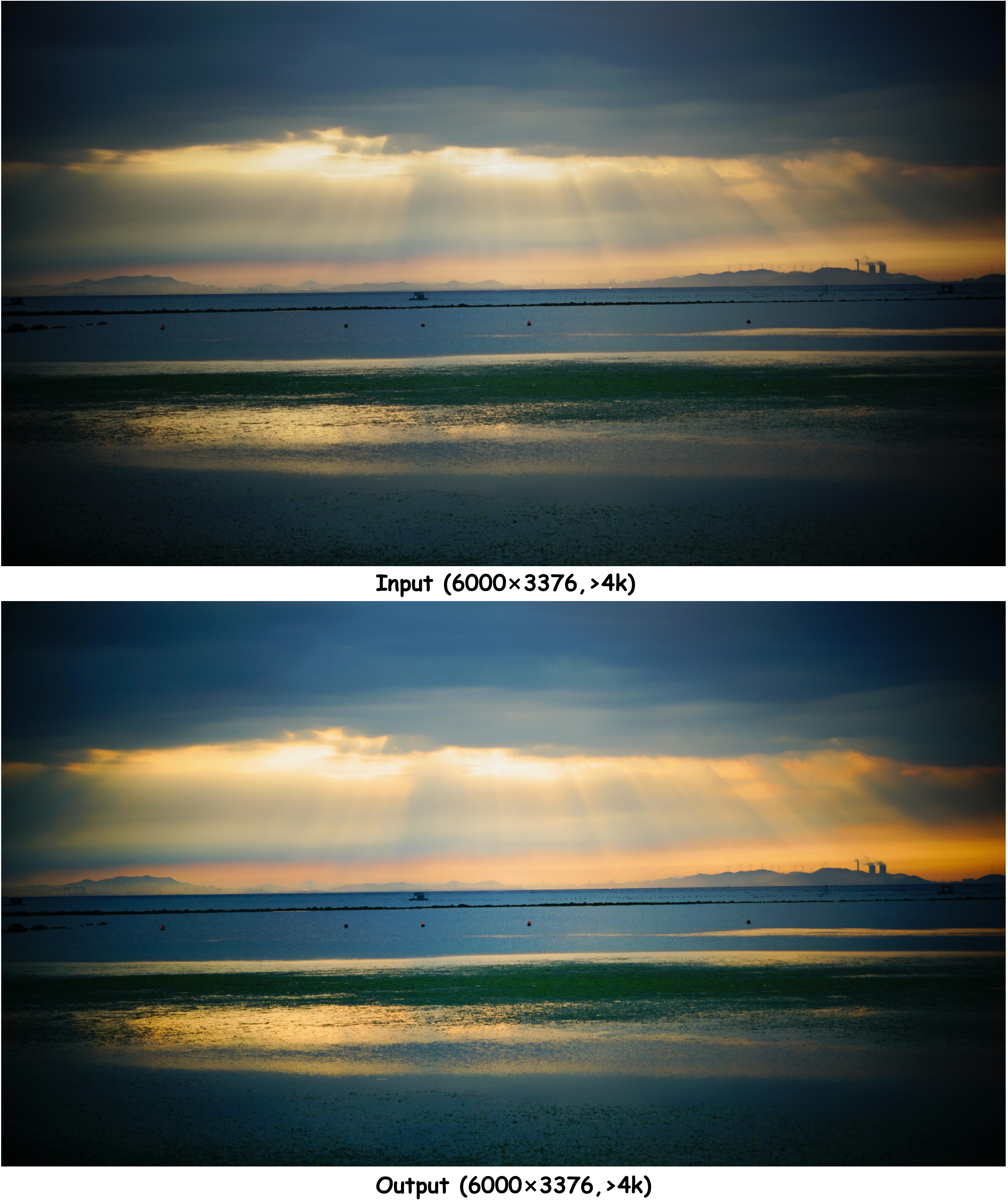}
  \caption{Retouching result on a 6000×3376 (over 4K) ultra-high-resolution image.}
  \label{fig:4k-2}
\end{figure*}

%% file: figs/texs/auto_result-s1.tex
\begin{figure*}
    \centering
  \includegraphics[width=\textwidth]{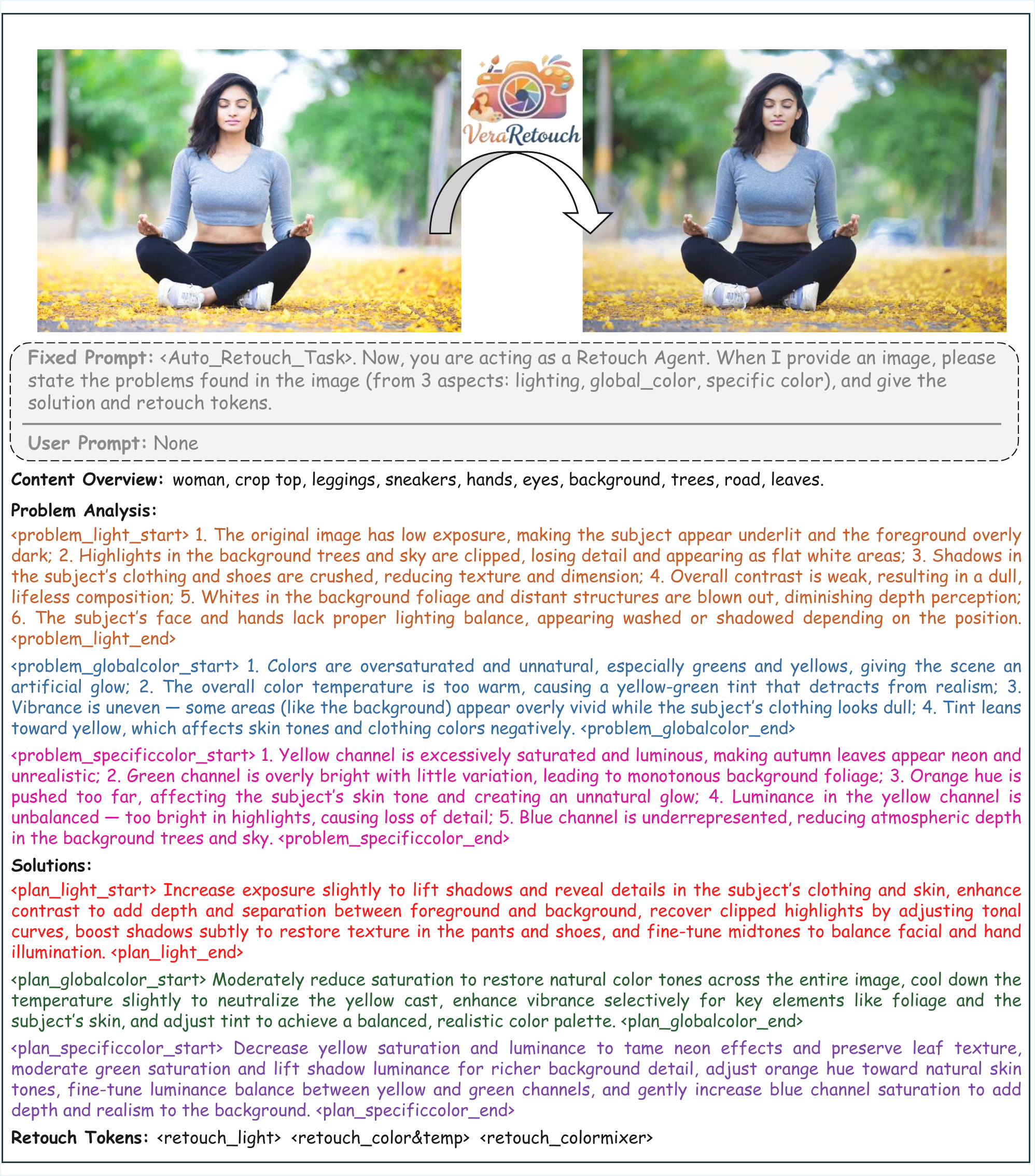}
  \caption{Complete input-output example of VeraRetouch on the \auto~task.}

  \label{fig:auto_r1}
\end{figure*}

%% file: figs/texs/auto_result-s2.tex
\begin{figure*}
    \centering
  \includegraphics[width=\textwidth]{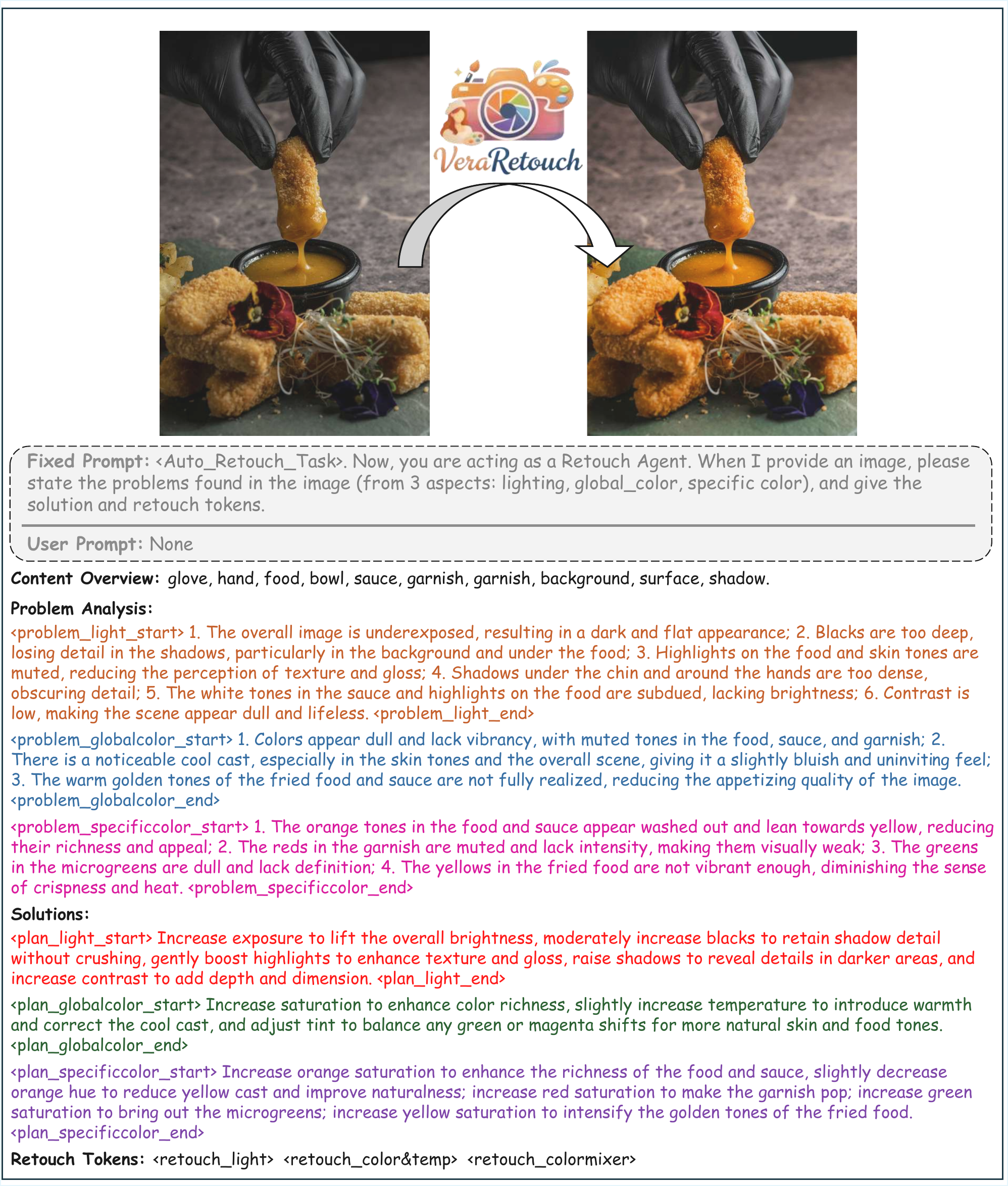}
  \caption{Complete input-output example of VeraRetouch on the \auto~task.}

  \label{fig:auto_r2}
\end{figure*}

%% file: figs/texs/auto_result-s3.tex
\begin{figure*}
    \centering
  \includegraphics[width=\textwidth]{figs/auto_result-s3.pdf}
  \caption{Complete input-output example of VeraRetouch on the \auto~task.}
  \label{fig:auto_r3}
\end{figure*}

%% file: figs/texs/auto_result-s4.tex
\begin{figure*}
    \centering
  \includegraphics[width=\textwidth]{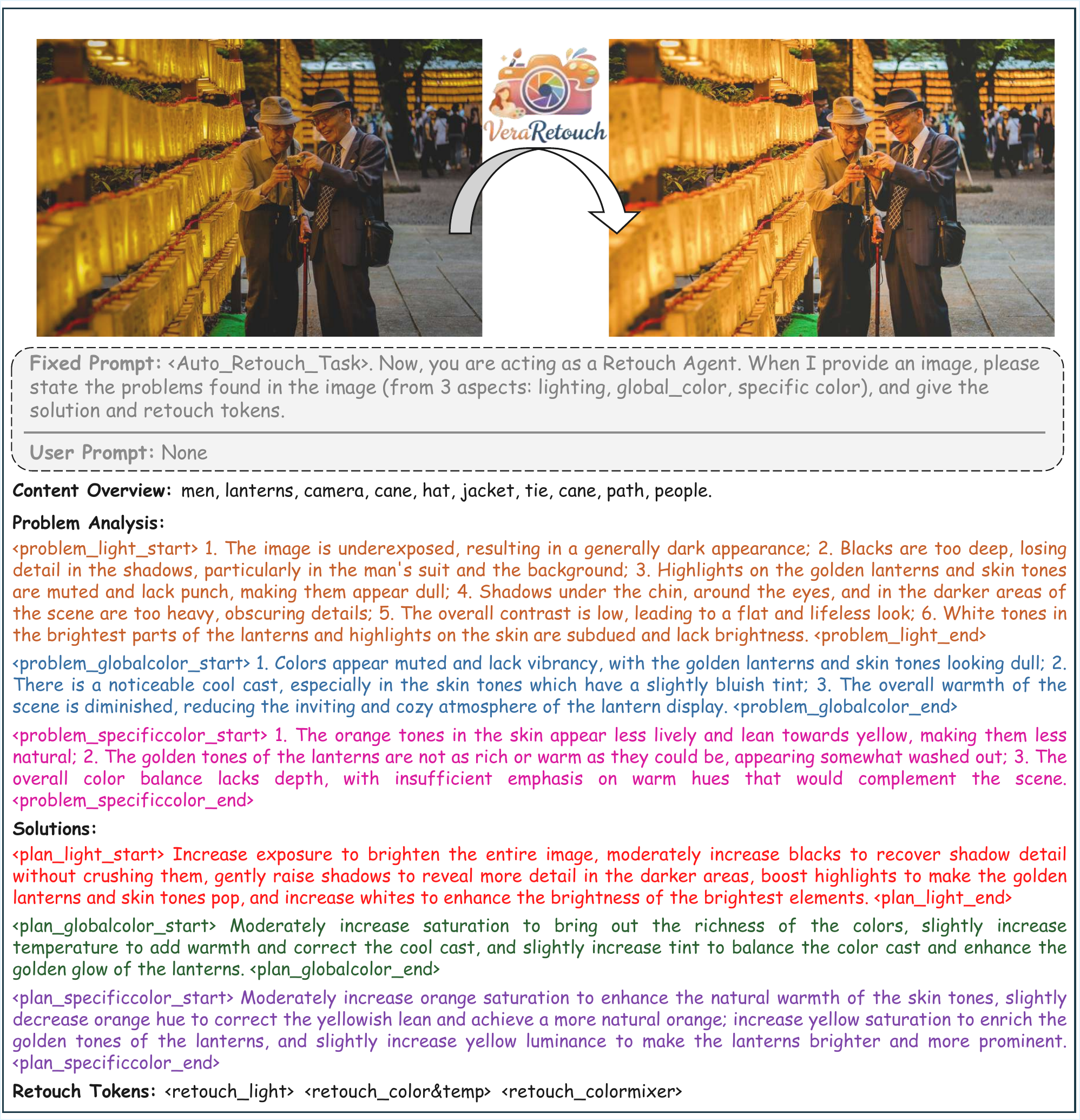}
  \caption{Complete input-output example of VeraRetouch on the \auto~task.}

  \label{fig:auto_r4}
\end{figure*}

%% file: figs/texs/auto_result-s5.tex
\begin{figure*}
    \centering
  \includegraphics[width=\textwidth]{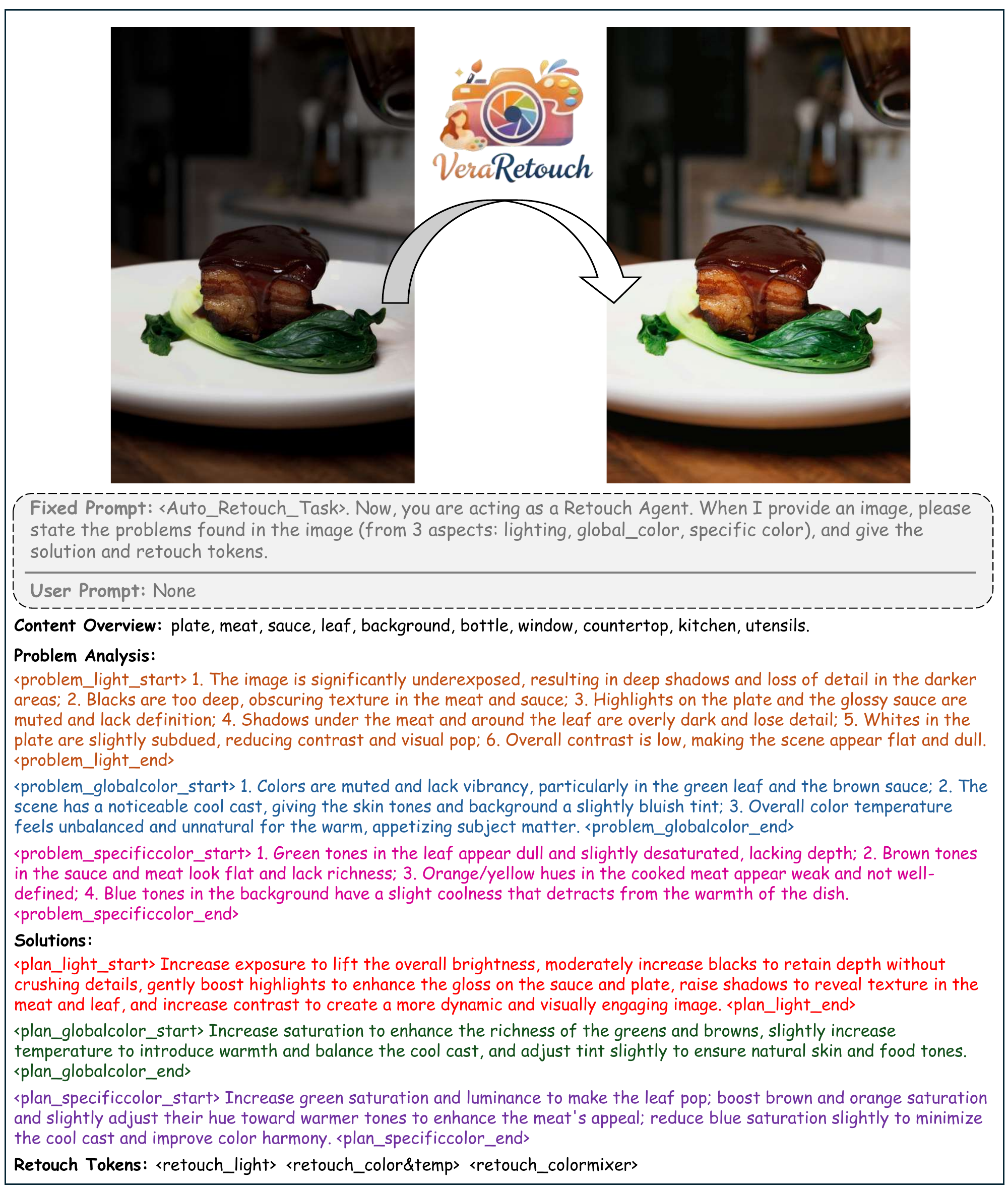}
  \caption{Complete input-output example of VeraRetouch on the \auto~task.}

  \label{fig:auto_r5}
\end{figure*}

%% file: figs/texs/auto_result-s6.tex
\begin{figure*}
    \centering
  \includegraphics[width=\textwidth]{figs/auto_result-s6.pdf}
  \caption{Complete input-output example of VeraRetouch on the \auto~task.}

  \label{fig:auto_r6}
\end{figure*}

%% file: figs/texs/auto_result-s7.tex
\begin{figure*}
    \centering
  \includegraphics[width=\textwidth]{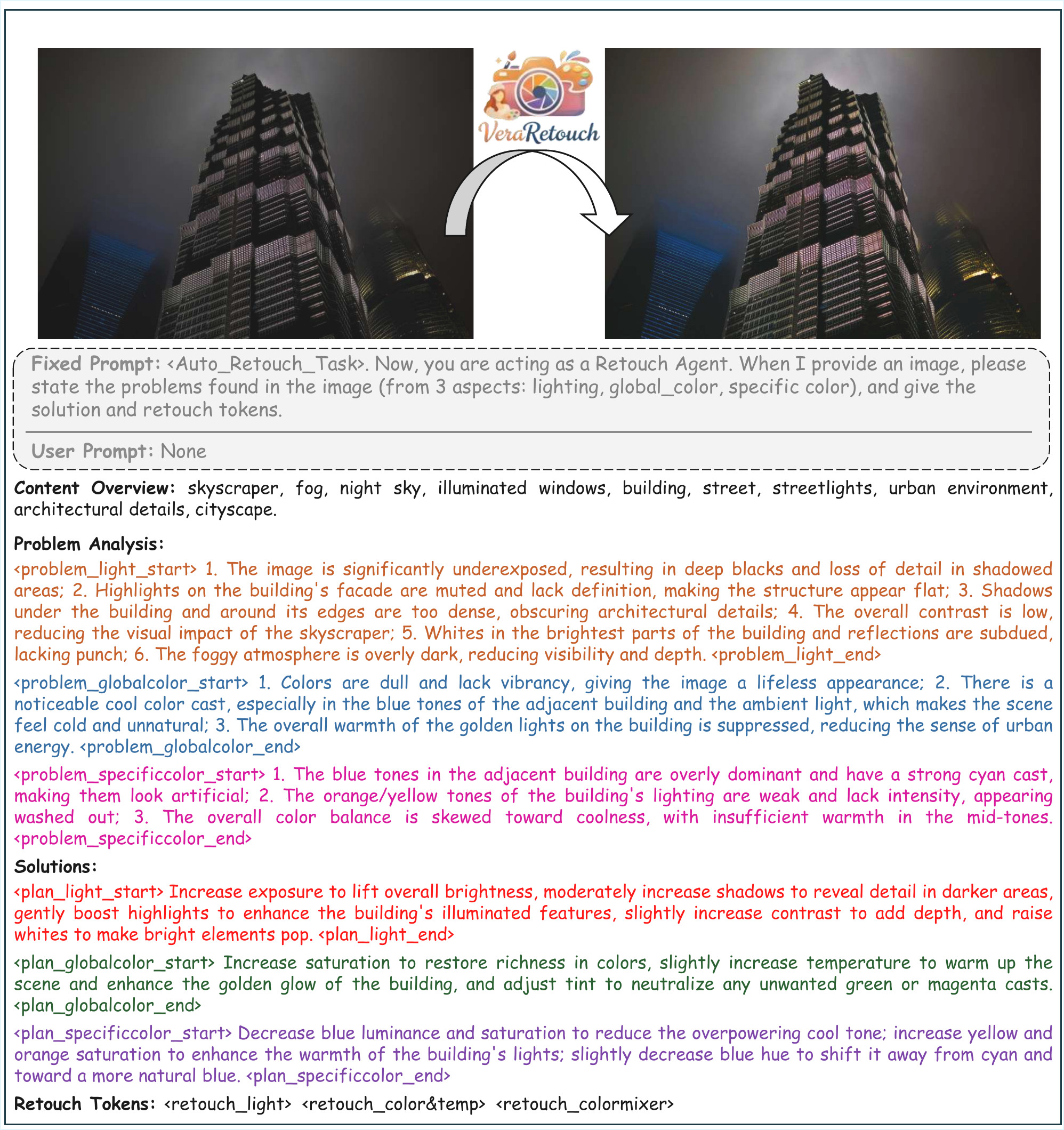}
  \caption{Complete input-output example of VeraRetouch on the \auto~task.}

  \label{fig:auto_r7}
\end{figure*}

%% file: figs/texs/style_result-s1.tex
\begin{figure*}
    \centering
  \includegraphics[width=\textwidth]{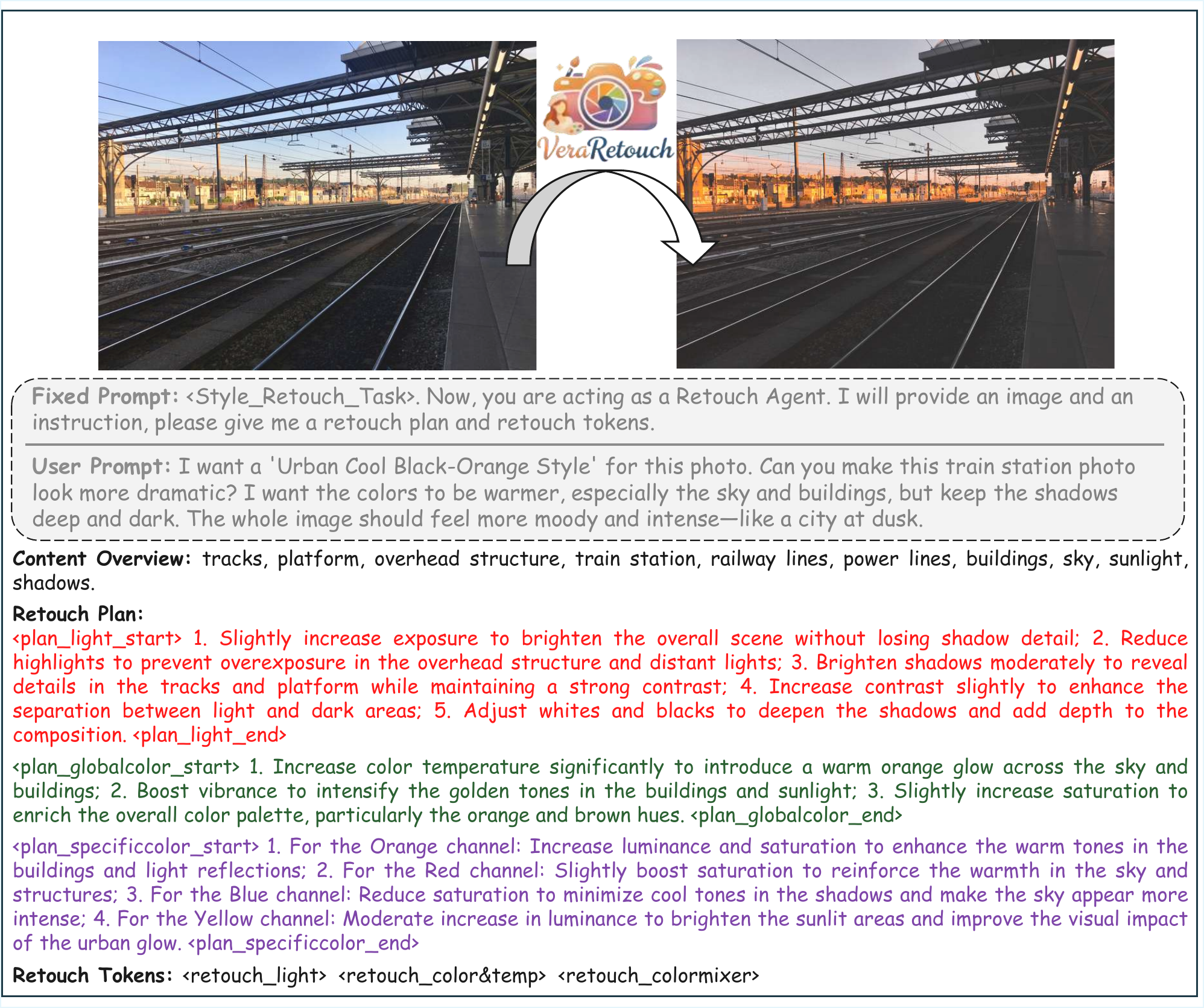}
  \caption{Complete input-output example of VeraRetouch on the \style~task.}

  \label{fig:style_r1}
\end{figure*}

%% file: figs/texs/style_result-s2.tex
\begin{figure*}
    \centering
  \includegraphics[width=\textwidth]{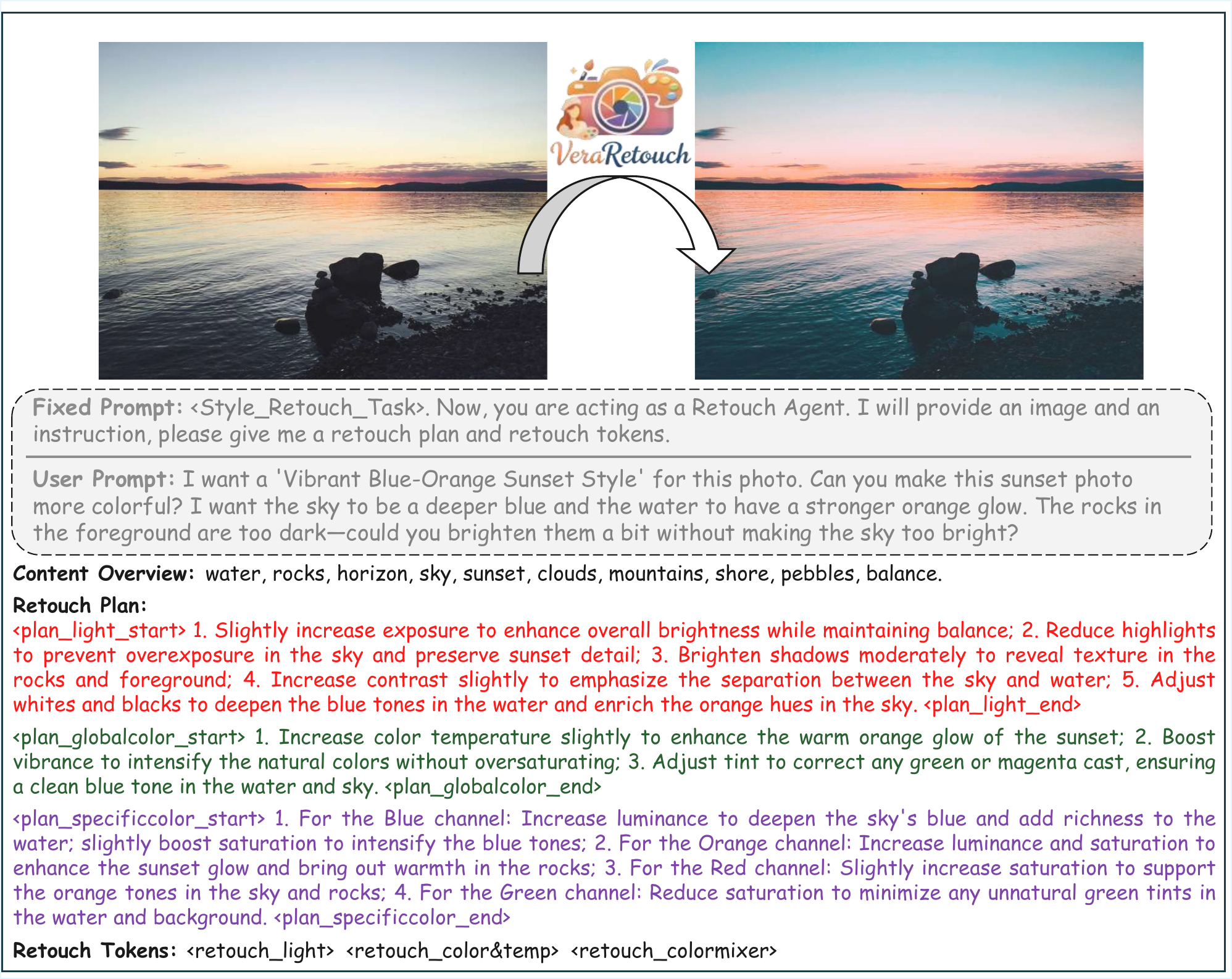}
  \caption{Complete input-output example of VeraRetouch on the \style~task.}

  \label{fig:style_r2}
\end{figure*}

%% file: figs/texs/style_result-s3.tex
\begin{figure*}
    \centering
  \includegraphics[width=\textwidth]{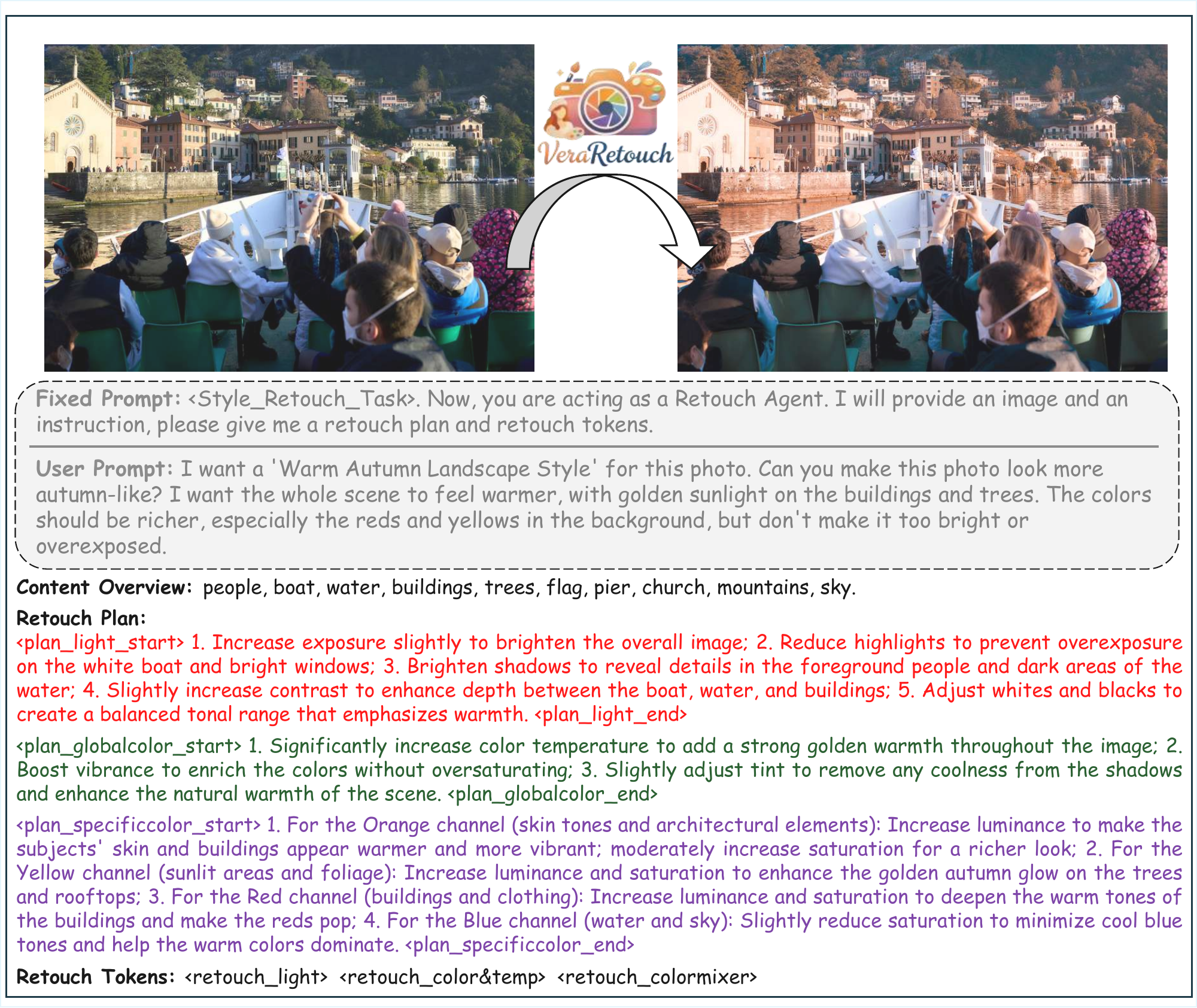}
  \caption{Complete input-output example of VeraRetouch on the \style~task.}

  \label{fig:style_r3}
\end{figure*}

%% file: figs/texs/style_result-s4.tex
\begin{figure*}
    \centering
  \includegraphics[width=\textwidth]{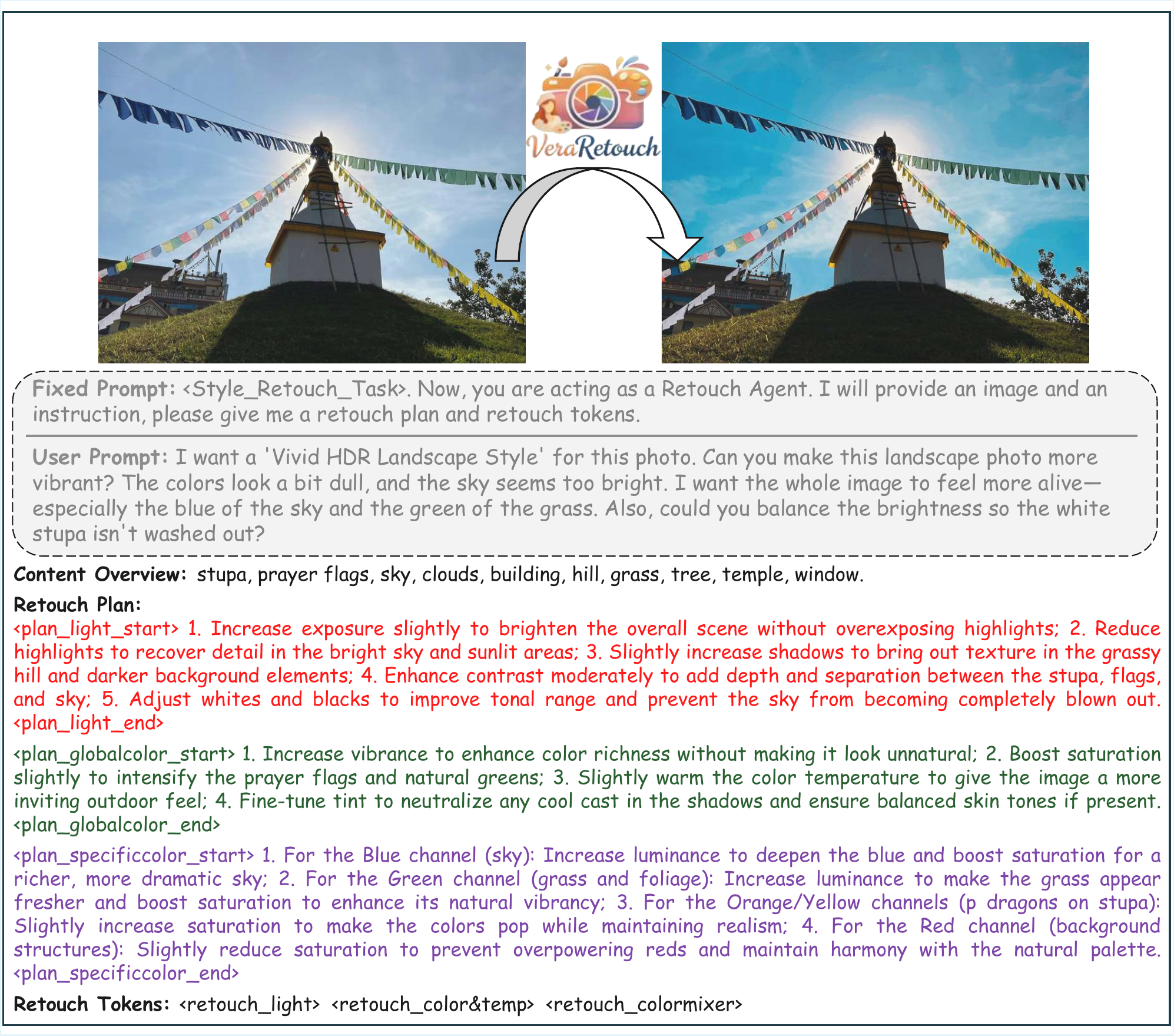}
  \caption{Complete input-output example of VeraRetouch on the \style~task.}

  \label{fig:style_r4}
\end{figure*}

%% file: figs/texs/style_result-s5.tex
\begin{figure*}
    \centering
  \includegraphics[width=\textwidth]{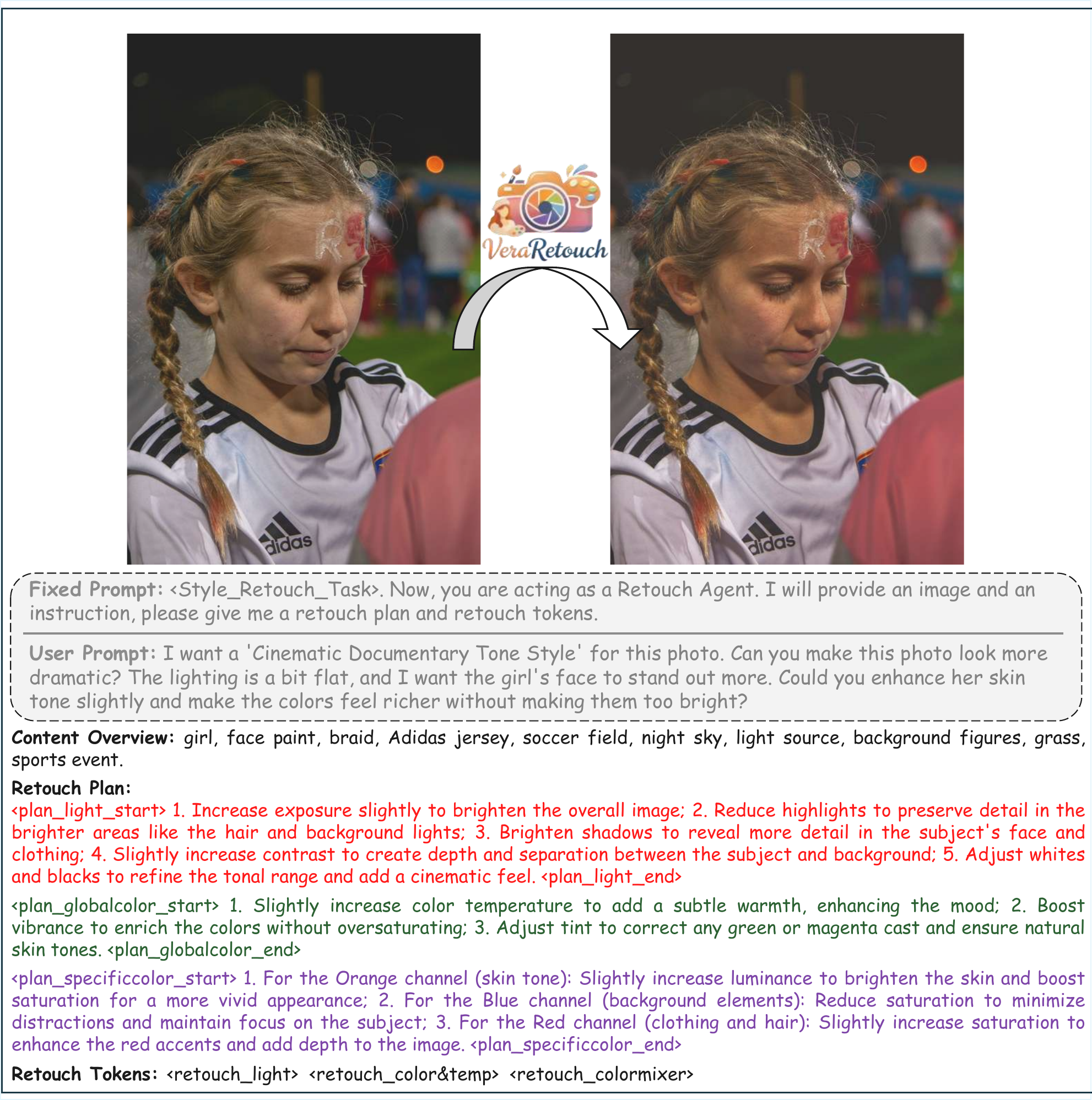}
  \caption{Complete input-output example of VeraRetouch on the \style~task.}

  \label{fig:style_r5}
\end{figure*}

%% file: figs/texs/style_result-s6.tex
\begin{figure*}
    \centering
  \includegraphics[width=\textwidth]{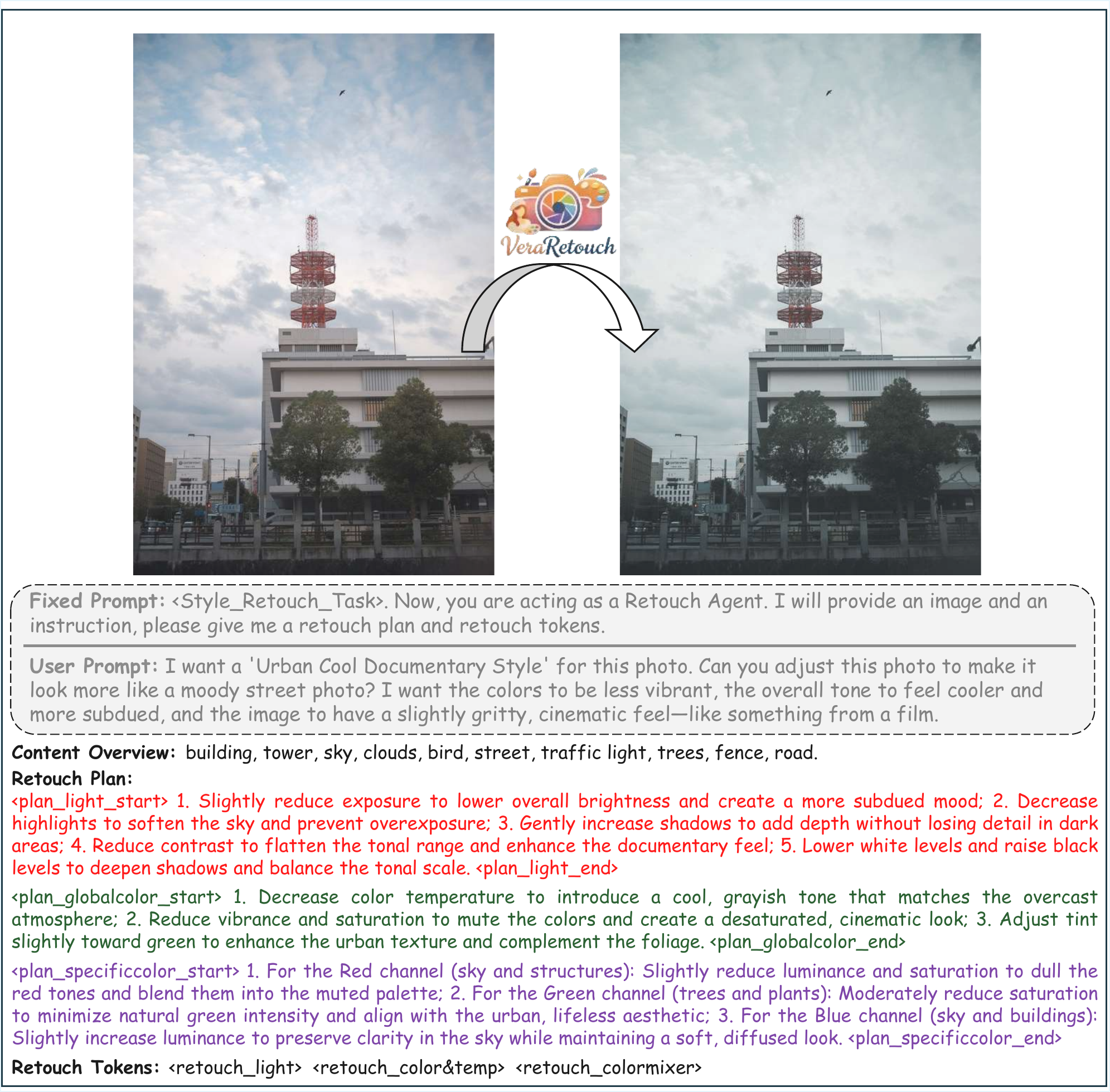}
  \caption{Complete input-output example of VeraRetouch on the \style~task.}

  \label{fig:style_r6}
\end{figure*}

%% file: figs/texs/style_result-s7.tex
\begin{figure*}
    \centering
  \includegraphics[width=\textwidth]{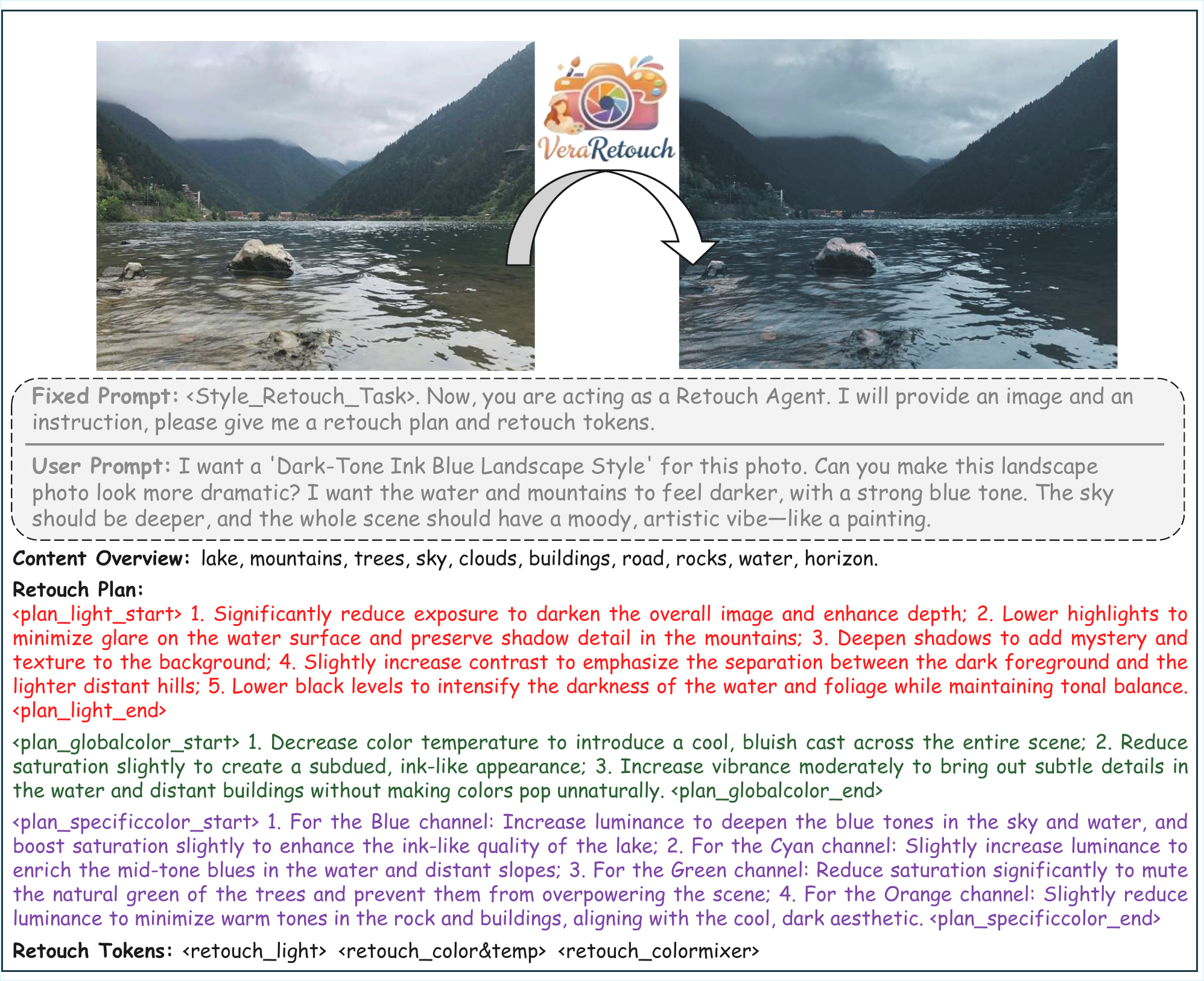}
  \caption{Complete input-output example of VeraRetouch on the \style~task.}

  \label{fig:style_r7}
\end{figure*}

%% file: figs/texs/professional_result-s1.tex
\begin{figure*}
    \centering
  \includegraphics[width=\textwidth]{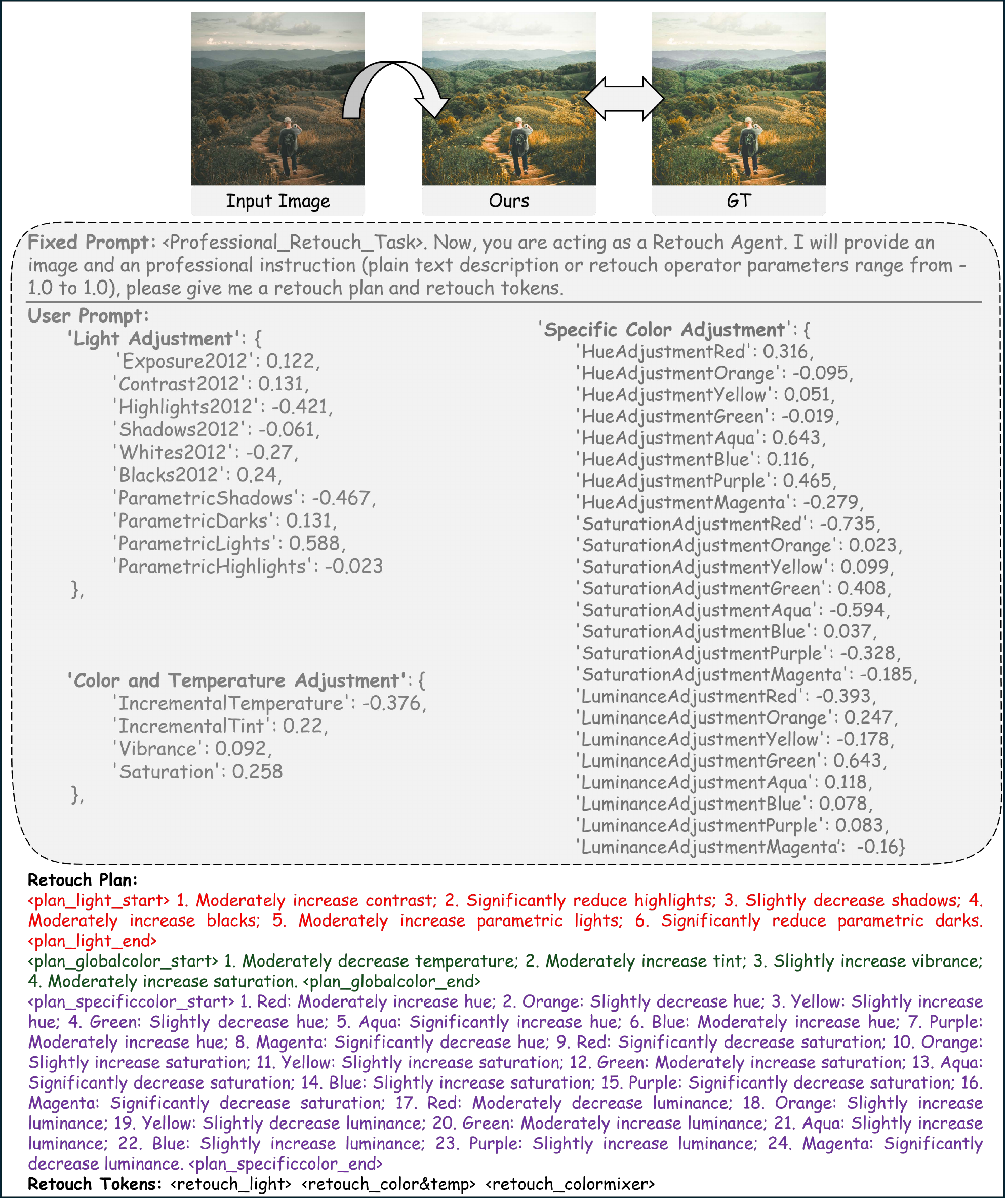}
  \caption{Complete input-output example of VeraRetouch on the \param~task, where all adjustment parameters are rearranged to the $[-1, 1]$.}

  \label{fig:param_r1}
\end{figure*}

%% file: figs/texs/professional_result-s2.tex
\begin{figure*}
    \centering
  \includegraphics[width=\textwidth]{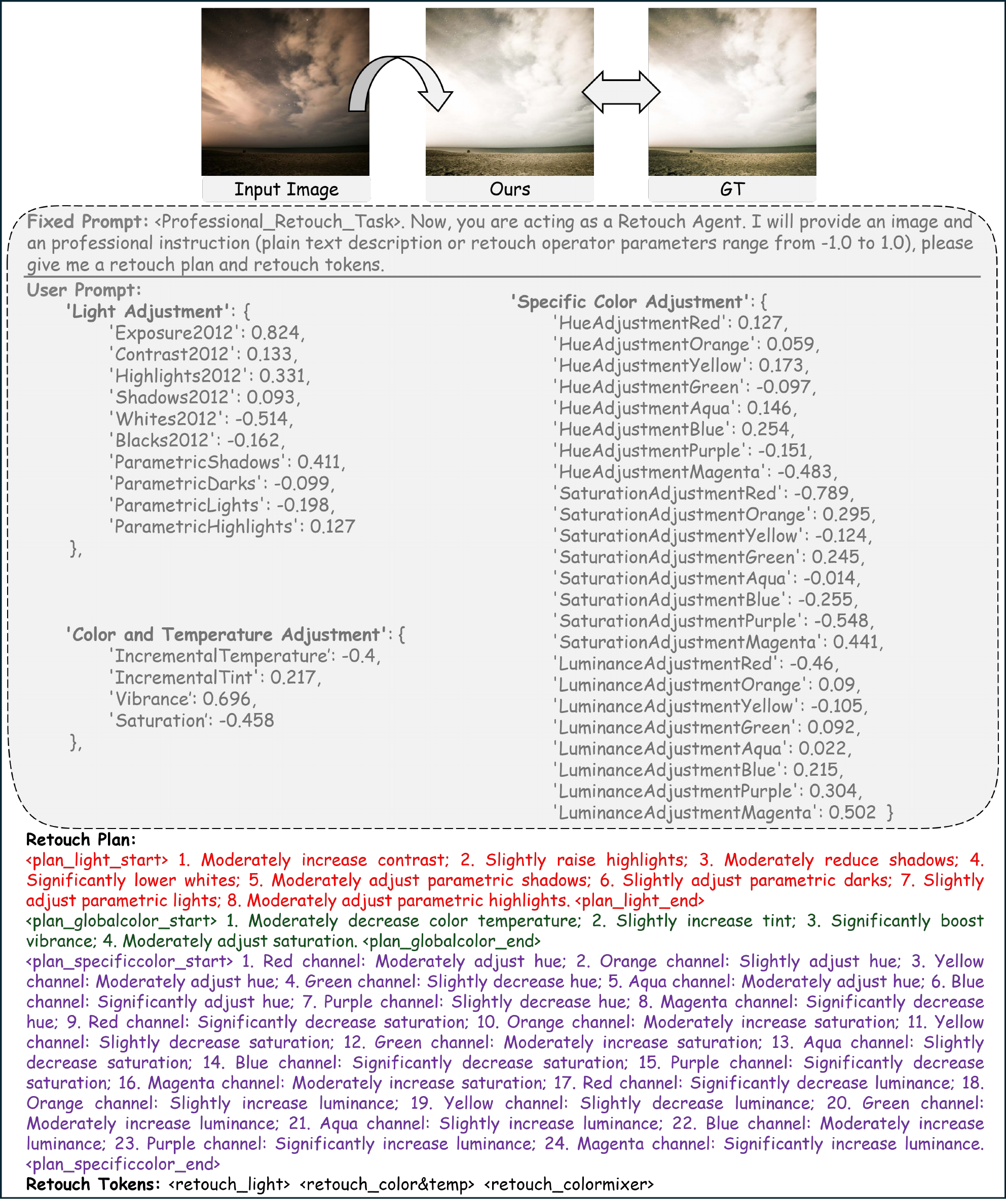}
  \caption{Complete input-output example of VeraRetouch on the \param~task, where all adjustment parameters are rearranged to the $[-1, 1]$.}

  \label{fig:param_r2}
\end{figure*}

%% file: figs/texs/professional_result-s3.tex
\begin{figure*}
    \centering
  \includegraphics[width=\textwidth]{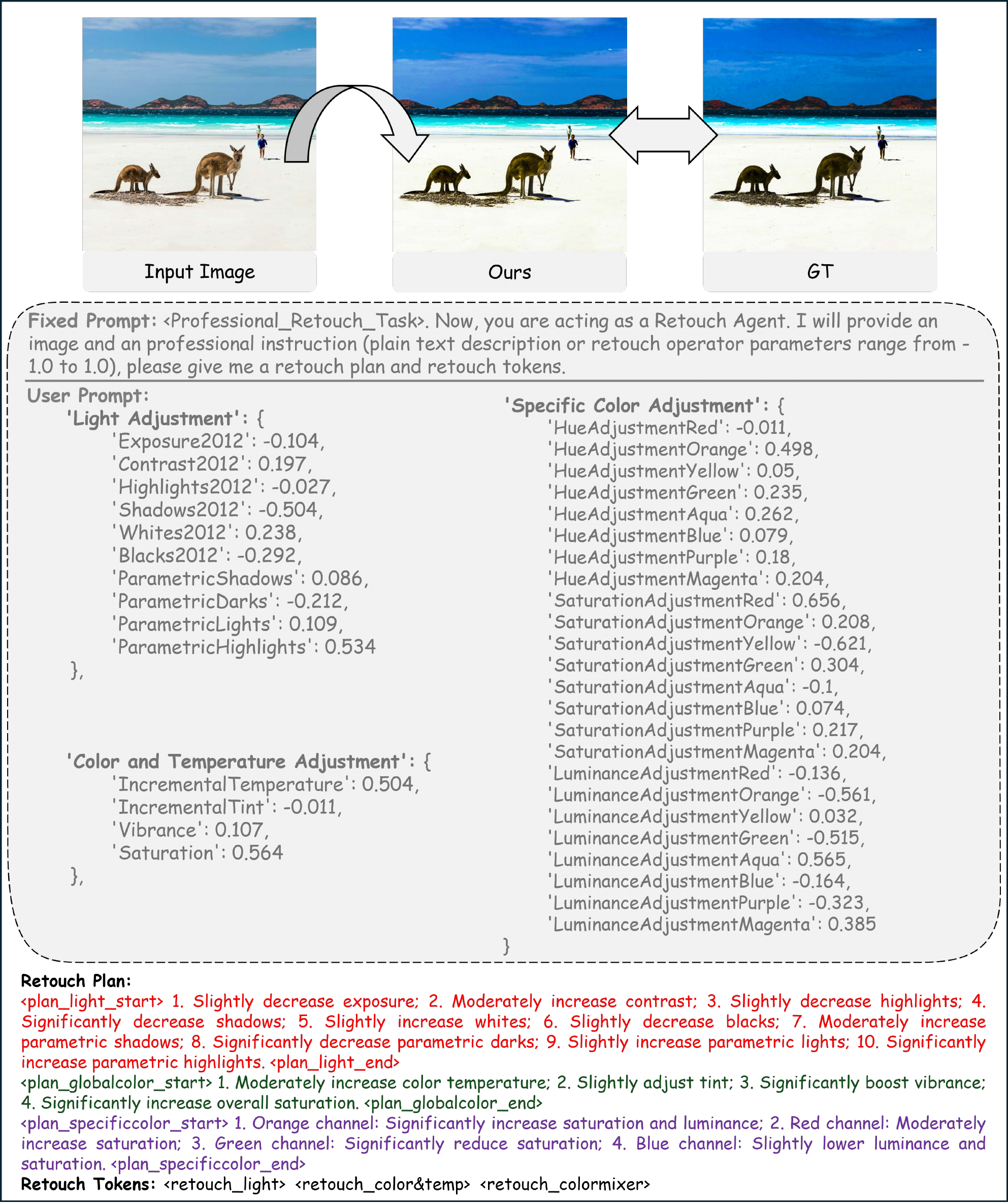}
  \caption{Complete input-output example of VeraRetouch on the \param~task, where all adjustment parameters are rearranged to the $[-1, 1]$.}

  \label{fig:param_r3}
\end{figure*}

%% file: figs/texs/professional_result-s4.tex
\begin{figure*}
    \centering
  \includegraphics[width=\textwidth]{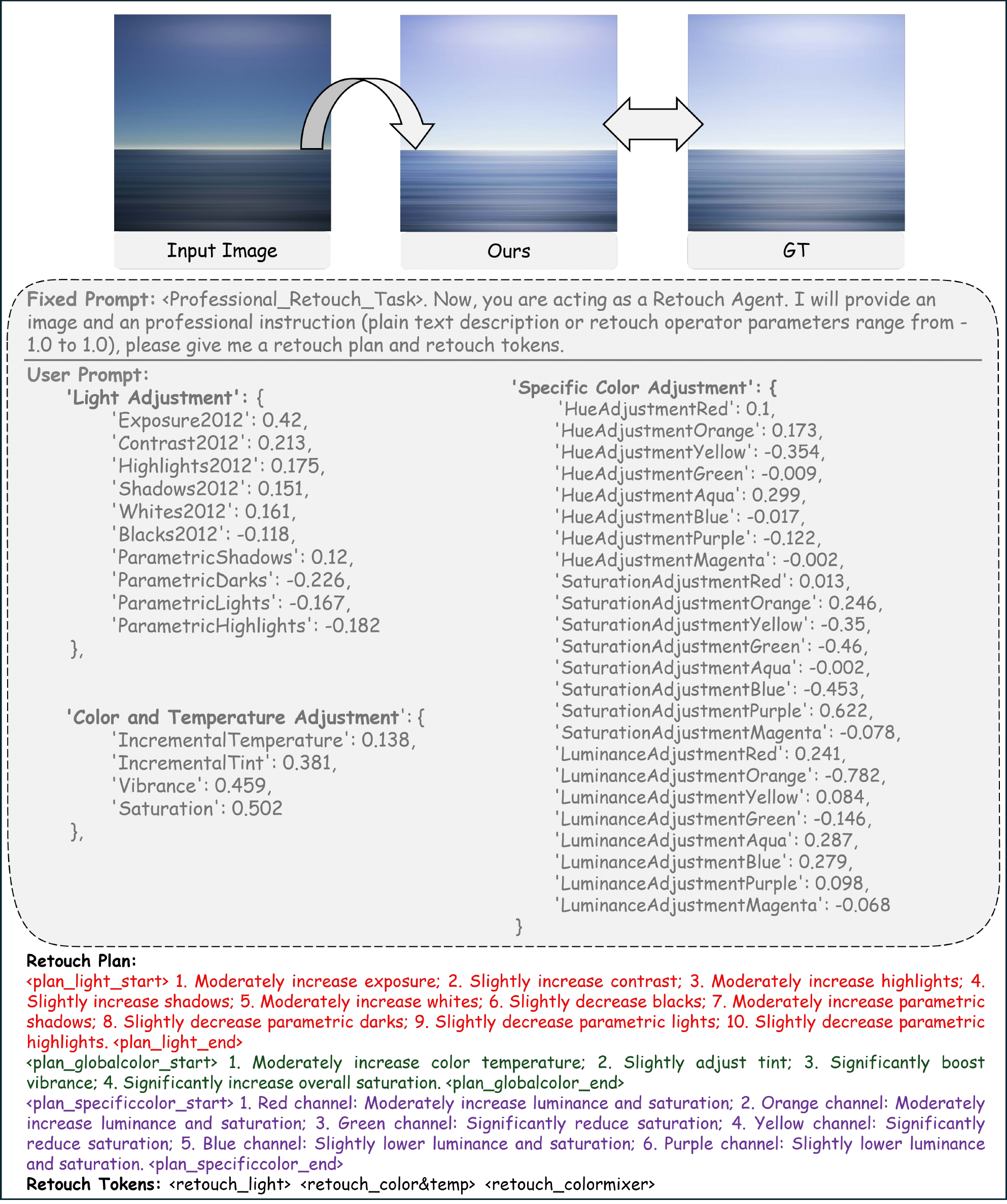}
  \caption{Complete input-output example of VeraRetouch on the \param~task, where all adjustment parameters are rearranged to the $[-1, 1]$.}

  \label{fig:param_r4}
\end{figure*}

%% file: figs/texs/professional_result-s5.tex
\begin{figure*}
    \centering
  \includegraphics[width=\textwidth]{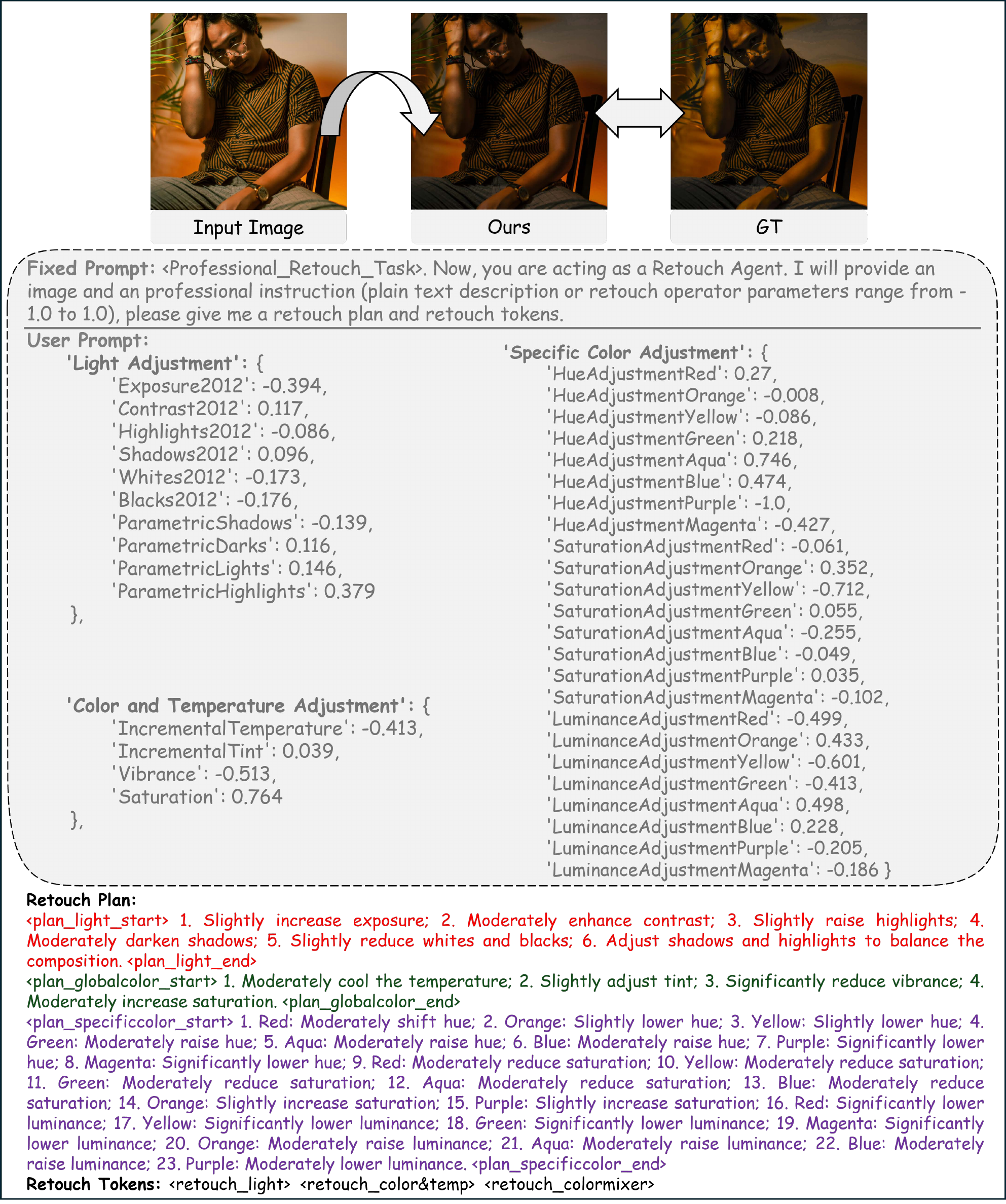}
  \caption{Complete input-output example of VeraRetouch on the \param~task, where all adjustment parameters are rearranged to the $[-1, 1]$.}

  \label{fig:param_r5}
\end{figure*}

%% file: figs/texs/professional_result-s6.tex
\begin{figure*}
    \centering
  \includegraphics[width=\textwidth]{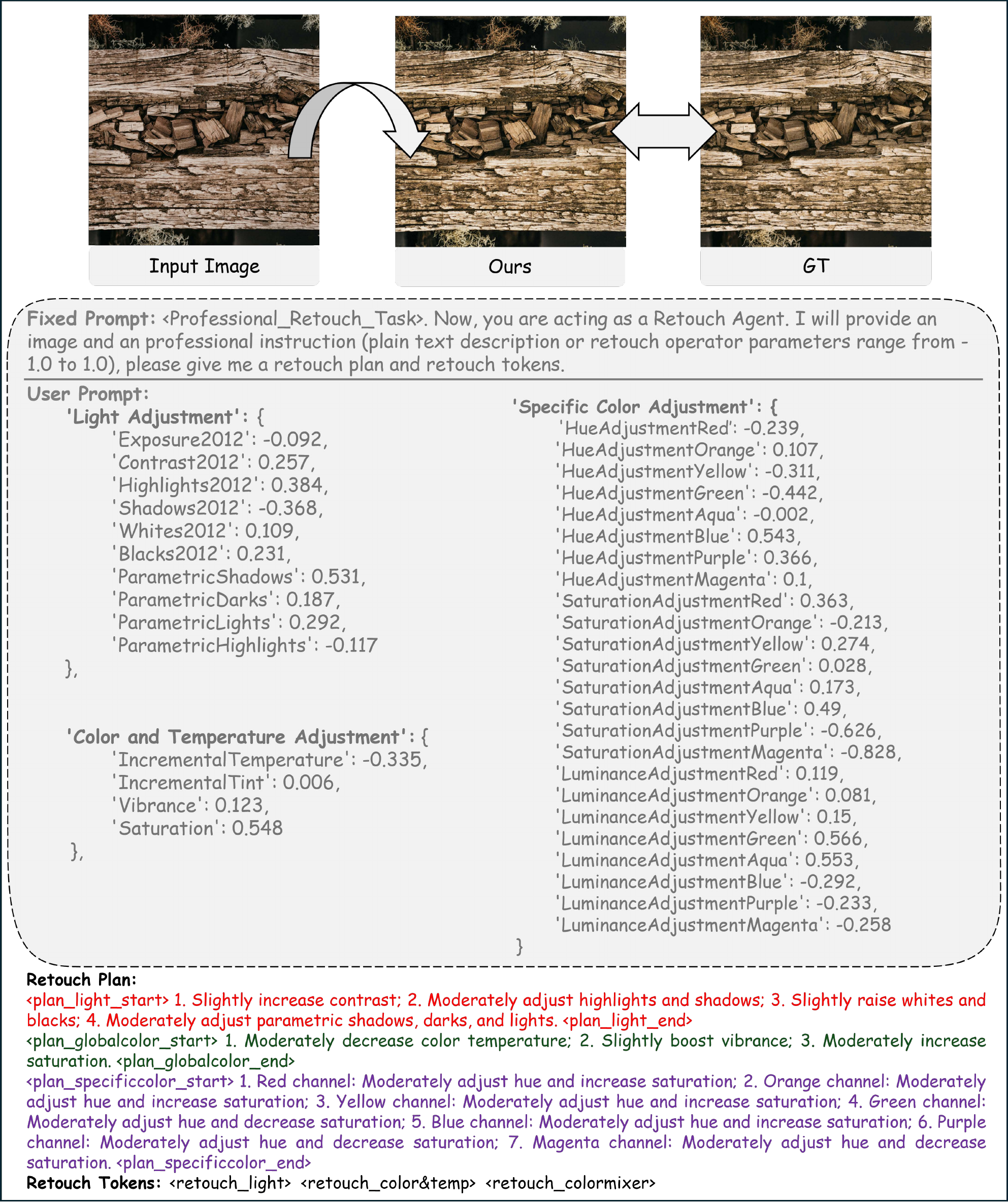}
  \caption{Complete input-output example of VeraRetouch on the \param~task, where all adjustment parameters are rearranged to the $[-1, 1]$.}

  \label{fig:param_r6}
\end{figure*}

%% file: figs/texs/professional_result-s7.tex
\begin{figure*}
    \centering
  \includegraphics[width=\textwidth]{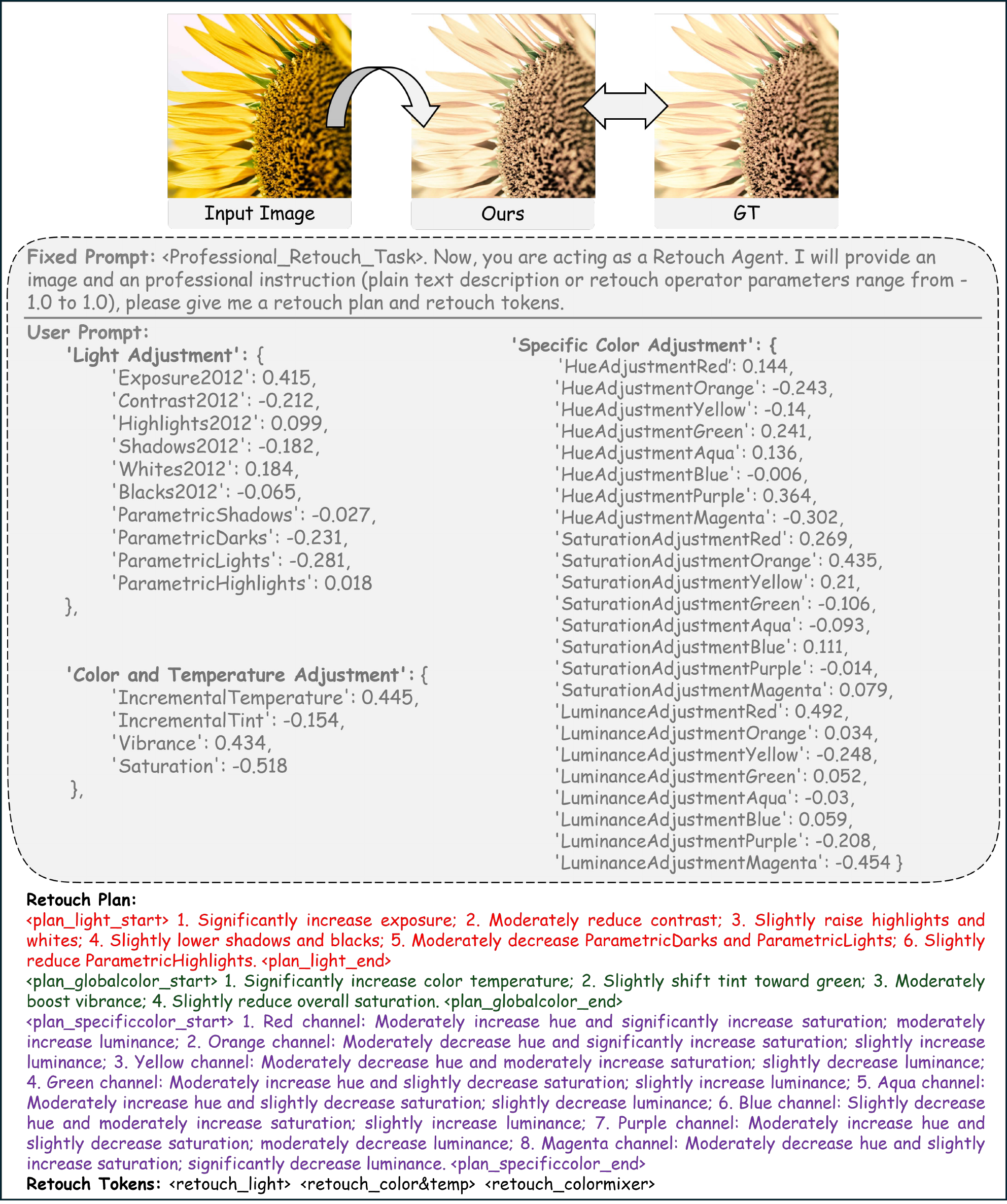}
  \caption{Complete input-output example of VeraRetouch on the \param~task, where all adjustment parameters are rearranged to the $[-1, 1]$.}

  \label{fig:param_r7}
\end{figure*}